\documentclass[a4paper,fleqn]{cas-sc}

\usepackage[authoryear,longnamesfirst]{natbib}
\usepackage{tabularx}
\usepackage{placeins}
\usepackage{subcaption}
\usepackage{caption}
\usepackage{booktabs}

\begin{document}
\let\WriteBookmarks\relax
\renewcommand{\topfraction}{0.95}
\renewcommand{\bottomfraction}{0.90}
\renewcommand{\textfraction}{0.05}
\renewcommand{\floatpagefraction}{0.85}
\setcounter{topnumber}{5}
\setcounter{bottomnumber}{5}
\setcounter{totalnumber}{10}

\shorttitle{Video-to-PDE}    
\shortauthors{Minoli, Sarkar}  

\title [mode = title]{From Video-to-PDE: Data-Driven Discovery of Nonlinear Dye Plume Dynamics}  

\author[1]{Cesar Acosta-Minoli}[orcid=https://orcid.org/0000-0002-7726-160X]
\ead{cminoli@uniquindio.edu.co}
\ead[LinkedIn]{https://www.linkedin.com/in/cesar-acosta-minoli/}
\credit{Conceptualization, Data curation, Investigation, Methodology, Software, Validation, Visualization, Writing -- original draft, Writing -- review and editing}

\affiliation[1]{organization={{GEDES, Universidad del Quindio}},
            addressline={Cra 15 Cl 12-00}, 
            city={Armenia},
            postcode={630004}, 
            state={Quindio},
            country={Colombia}}

\author[2]{Sayantan Sarkar}[orcid=https://orcid.org/0000-0002-9630-9509]
\ead{sayantan@buffalo.edu}
\cormark[1]
\ead[LinkedIn]{www.linkedin.com/in/sayantan-sarkar-s117}
\credit{Conceptualization, Formal analysis, Methodology, Software, Validation, Visualization, Writing -- original draft, Writing -- review and editing}

\affiliation[2]{organization={Department of Mathematics, State University of New York at Buffalo},
            addressline={Mathematics Building}, 
            city={Buffalo},
            postcode={14260}, 
            state={New York},
            country={USA}}

\cortext[1]{Corresponding author}

\begin{abstract}
Inferring continuum models directly from video is hampered by two facts: the recorded field is uncalibrated image intensity rather than a physical state, and direct numerical differentiation of noisy frames is unstable. We develop a video-to-PDE pipeline that converts grayscale recordings of an ink plume into a normalised scalar field $u(x,y,t)$, isolates a bulk drift $\mathbf{v}(t)$ from intrinsic spreading via the intensity-weighted centroid, and identifies an effective transport law by weak-form sparse regression. Conditioning, threshold-sweep and random-centre diagnostics show that overcomplete libraries are strongly collinear; the search is therefore restricted to compact gradient-based libraries. Coefficients are refined by an inverse physics-informed network and recalibrated against forward rollouts, with a chronological block bootstrap quantifying uncertainty. The selected reduced model
$u_t+\mathbf v(t)\!\cdot\!\nabla u = 9.005\,|\nabla u|^{2}+0.666\,\Delta u$
outperforms advection--diffusion baselines on held-out frames, retains a positive Laplacian coefficient, and admits a Cole--Hopf reduction to a linear advection--diffusion equation. The framework demonstrates that uncalibrated visual data can yield compact, predictive and structurally interpretable continuum models when discovery, calibration and uncertainty are treated as distinct stages.
\end{abstract}


\begin{highlights}
\item A video-to-PDE pipeline is developed for dye-plume dynamics.
\item Weak-form regression avoids direct time differentiation of video data.
\item Rollout calibration selects a nonlinear-gradient transport model.
\item Bootstrap and front diagnostics assess coefficient robustness.
\item The selected PDE admits a Cole--Hopf linearization.
\end{highlights}


\maketitle

\section{Introduction}\label{sec:introduction}

Modern experiments routinely record transport processes on video, but the recorded quantity is image intensity rather than a calibrated physical field, and apparent motion is entangled with photometric, geometric, and finite-resolution artefacts. Classical optical-flow methods recover apparent velocities from image sequences~\citep{Horn1981}; particle-image and particle-tracking velocimetry yield flow estimates from visual measurements~\citep{Westerweel2013PIV,Ohmi2009PTV}; and image-processing techniques extract fronts and scalar structures from experimental footage~\citep{DeLeonRuiz2022FlameFrontDetection}. More recent hidden-physics models, physics-informed networks, and physics-guided vision recover latent fields, parameters, and surrogate dynamics directly from images~\citep{Raissi2018,Both2021,Raissi2020HiddenFluidMechanics,Chu2022PINeRF,Yu2023HybridNeuralFluid,Dreisbach2024PINNs4Drops,Jaques2020PhysicsInverseGraphics}, and recent reviews emphasise the growing role of vision in computational fluid inference~\citep{Banerjee2024PhysicsInformedCV,Yu2023,Sharma2023PhysicsInformedML,Vinuesa2023MLExperiments}. None of these approaches, however, returns an explicit and simulable PDE in the observed image-intensity variable -- the gap addressed here.

The sparse identification framework of \citet{brunton2016sindy} represents a governing law as a sparse combination of library terms, and PDE-FIND extended this idea to spatiotemporal differential operators~\citep{Rudy2017PDEFIND}. Subsequent developments include information-theoretic model selection, sparse relaxation, parametric and BVP formulations, and reproducible software~\citep{mangan2017aicsindy,zheng2018sr3,Rudy2019,Shea2021SINDyBVP,Kaptanoglu2022}; symbolic--numeric hybrids such as PDE-Net offer a complementary route~\citep{Long2019PDENet}. Strong-form discovery is, however, fragile when applied to noisy or limited data, since numerical differentiation amplifies pixel-scale fluctuations. Integral, weak, and ensemble formulations transfer derivatives onto smooth test functions and are now well established~\citep{Schaeffer2017IntegralSINDy,Messenger2021WeakSINDy,messenger2021weak,Reinbold2020,Kaheman2020SINDyPI,Fasel2022EnsembleSINDy,Wentz2023DSINDy,zhang2019convergencesindy}, with sparse Galerkin and reduced-order extensions linking discovery to interpretable flow modelling~\citep{Loiseau2018CSGR,Loiseau2018SparseROM,Fukami2021LowDimSINDy,Joshi2022}. Even in weak form, libraries built from image-intensity features tend to be strongly collinear, so identifiability -- not regression fit -- must drive model selection~\citep{antonelli2022}.

Three further issues motivate the pipeline developed here. First, the plume drifts across the field of view; we estimate this drift from the intensity-weighted centroid and impose it as a prescribed advective component, decoupling bulk translation from intrinsic dynamics. Second, regression coefficients minimise a Galerkin residual rather than a forward-rollout error; following the spirit of \citet{antonelli2022}, we therefore separate structure discovery from coefficient calibration, refining intrinsic coefficients with an inverse physics-informed network~\citep{Raissi2019PINNs,Karniadakis2021PINNs,Cuomo2022PINNReview} and then recalibrating them by a derivative-free Nelder--Mead routine~\citep{NelderMead1965,GaoHan2012,Larson2019DFO} against forward rollouts. Third, pixel-wise error rarely tells the full story: a model may track image intensity faithfully while misplacing the centroid or front. Coefficient stability is assessed by chronological block bootstrap~\citep{Efron1979,Kunsch1989,LiuSingh1992,PolitisWhite2004}, with centre-of-mass and equivalent-front-radius diagnostics complementing pixel-wise RMSE.

Applied to a top-view ink-plume video, the pipeline selects the reduced nonlinear-gradient transport law
\[
u_t + \mathbf{v}(t)\!\cdot\!\nabla u = a|\nabla u|^{2}+\beta\,\Delta u,\qquad a,\beta>0,
\]
over advection--diffusion baselines and over larger collinear libraries. The contributions are threefold. (i)~A complete pipeline that converts uncalibrated dye-plume video into an interpretable, simulable PDE, integrating measured-drift correction, weak-form discovery, iPINN refinement, bootstrap calibration, and front-aware diagnostics. (ii)~A model-selection protocol tied to forward-rollout performance and geometric admissibility rather than to regression residuals. (iii)~The selected equation is a viscous Hamilton--Jacobi equation that admits a Cole--Hopf reduction to a linear advection--diffusion equation, providing analytical interpretability of the discovered nonlinear gradient term.

\FloatBarrier
\section{Data Collection and Preprocessing}\label{sec:data_preprocessing}

We construct a normalised, consistently indexed, and mildly smoothed image-derived observable on which all subsequent regression and rollout diagnostics are computed.

\subsection{Experimental Setup}\label{subsec:experimental_setup}

\begin{figure}
    \centering
    \includegraphics[width=0.45\linewidth]{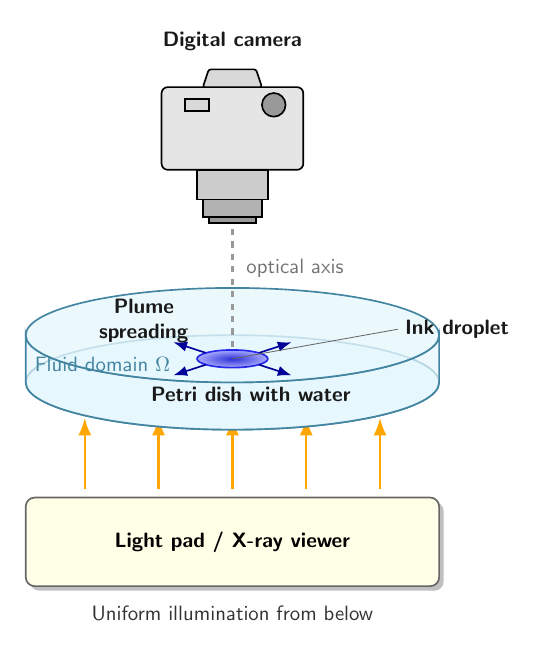}
    \caption{Experimental setup. A top-view camera records the spreading of an ink droplet over the spatial domain \(\Omega\); grayscale frames are converted into the image-derived scalar field \(u(x,y,t)\).}
    \label{fig:experimental_setup}
\end{figure}

A transparent dish of water sits on a uniformly back-illuminated source, an ink droplet is introduced at a fixed location, and the resulting quasi-two-dimensional plume is recorded by a fixed overhead camera (Fig.~\ref{fig:experimental_setup}). The fixed geometry suppresses frame-to-frame jitter and the top-view configuration permits a two-dimensional representation of the evolving plume. The setup is deliberately simple: the goal is not to control transport precisely but to obtain a controlled, repeatable, and non-ideal observational record from which an effective evolution law can be recovered.

\subsection{Construction of the Image-Derived Field}\label{subsec:video_to_field}

Let \(I_{\mathrm{gray}}(x,y,t)\in[0,255]\) denote the grayscale intensity of a recorded frame. Since ink appears darker than the back-lit background, we define the inverted, normalised observable
\begin{equation}
    u(x,y,t) = 1 - I_{\mathrm{gray}}(x,y,t)/255,\qquad 0\le u\le 1,
    \label{eq:image_observable}
\end{equation}
so that background regions are small and ink-rich regions are large. The field \(u\) is not a calibrated dye concentration; the discovered PDE should therefore be read as an effective evolution law for the image-derived observable, with coefficients in image-coordinate units.

\begin{figure}
    \centering
    \includegraphics[width=0.5\textwidth]{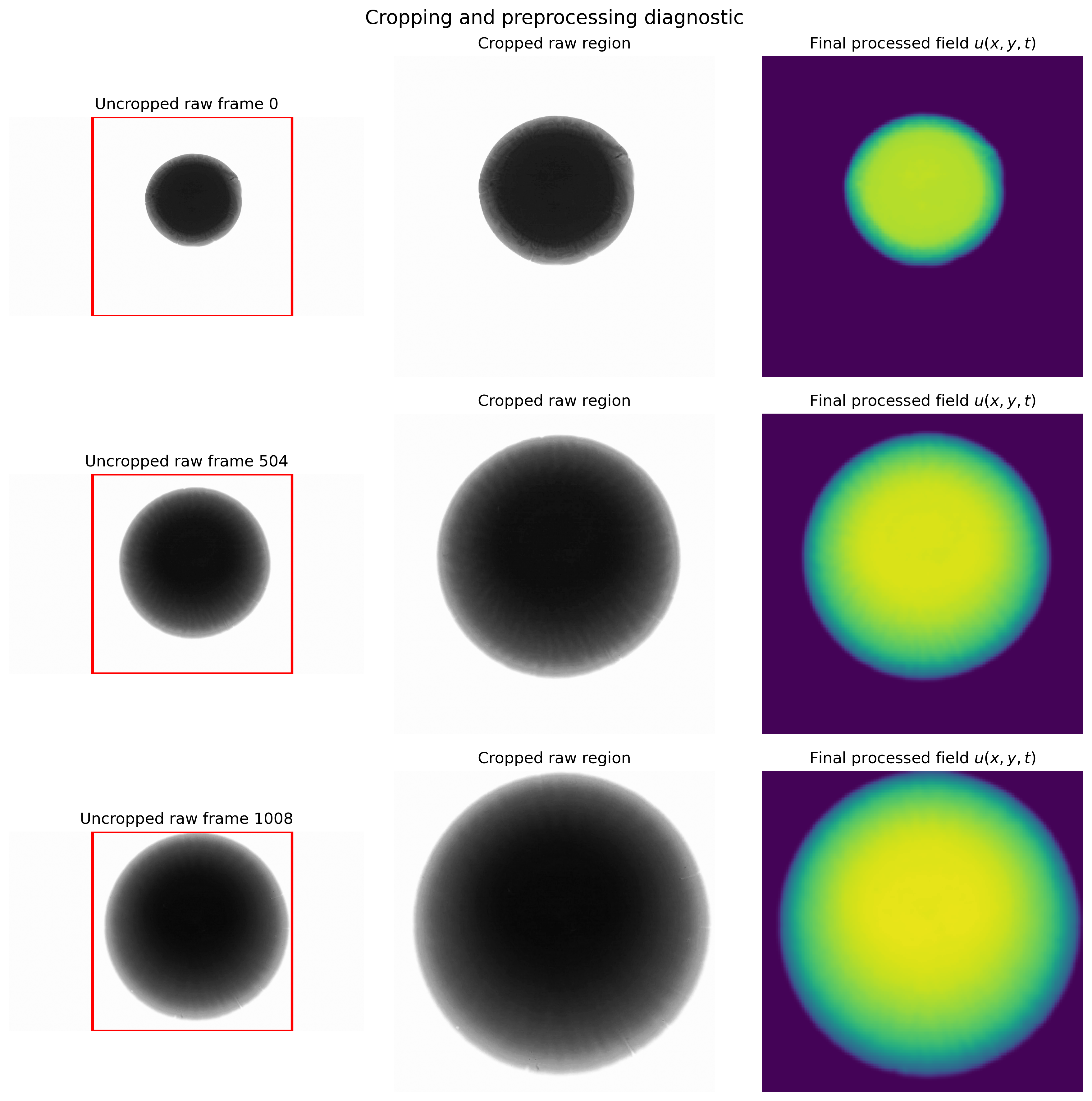}
    \caption{Preprocessing diagnostic. Each row shows an uncropped grayscale frame with the retained crop region indicated, the corresponding cropped region, and the final field \(u(x,y,t)\). Colour represents the normalised intensity, not physical dye colour.}
    \label{fig:preprocessing_diagnostic}
\end{figure}

The processed data are stored as a tensor \(U=\{u_{j,i}^{k}\}\in\mathbb{R}^{n_y\times n_x\times n_t}\) with \(U[j,i,k]\approx u(x_i,y_j,t_k)\), on a uniform grid with spacings \(\Delta x=L_x/(n_x-1)\), \(\Delta y=L_y/(n_y-1)\), \(\Delta t=T/(n_t-1)\). The reported data use \(n_x=n_y=200\) on \(L_x=L_y=200\) image-coordinate units, with \(T=34\)\,s. Spatial coordinates are image units, not calibrated lengths. Figure~\ref{fig:preprocessing_diagnostic} shows representative raw frames, the retained crop, and the resulting field; the pseudocolour palette is for visualisation only.

\subsection{Preprocessing Pipeline}\label{subsec:preprocessing_pipeline}

Each raw frame is converted to grayscale, cropped to the rectangular region \((x_0,y_0,w,h)=(450,0,1080,1080)\) in pixel units, the eight-pixel boundary is removed, intensity is inverted and normalised via Eq.~\eqref{eq:image_observable}, the frame is resized to a \(200\times 200\) grid, and a Gaussian filter with \(\sigma=1.0\) pixel is applied. The smoothing parameter was chosen from a sensitivity sweep over \(\sigma\in\{0,0.5,1.0,1.5,2.0,3.0\}\) (Appendix~\ref{app:smoothing_diagnostic}); \(\sigma=1.0\) suppresses pixel-scale noise while preserving the centre-line profile and the large-scale plume geometry.

Table~\ref{tab:preprocessing} summarises the stages. The mean intensity is preserved across resizing, indicating that spatial reduction does not materially distort the global content; a crop-retention check (background-corrected signal inside vs.\ outside the crop) confirms that the retained region carries the dominant plume signal.

\begin{table}[htbp]
\centering
\caption{Preprocessing stages and bulk statistics. The raw stack uses acquisition order \((n_t,H,W)\); processed arrays use storage order \((n_y,n_x,n_t)\). The final field is \(u\in[0,1]\).}
\label{tab:preprocessing}
\begin{tabularx}{\linewidth}{@{}>{\raggedright\arraybackslash}Xcccc@{}}
\toprule
\textbf{Stage} & \textbf{Shape} & \textbf{Min} & \textbf{Mean} & \textbf{Max} \\
\midrule
Cropped raw frame stack 
& \((1009,1080,1080)\) 
& -- 
& -- 
& -- \\
After border removal and normalisation 
& \((1064,1064,1009)\) 
& \(0.0000\) 
& \(0.3498\) 
& \(0.9843\) \\
Final resized and smoothed field 
& \((200,200,1009)\) 
& \(0.0050\) 
& \(0.3498\) 
& \(0.9720\) \\
\bottomrule
\end{tabularx}
\end{table}

\subsection{Preliminary Plume Diagnostics}\label{subsec:preliminary_diagnostics}

\begin{figure}
    \centering
    \includegraphics[width=0.6\textwidth]{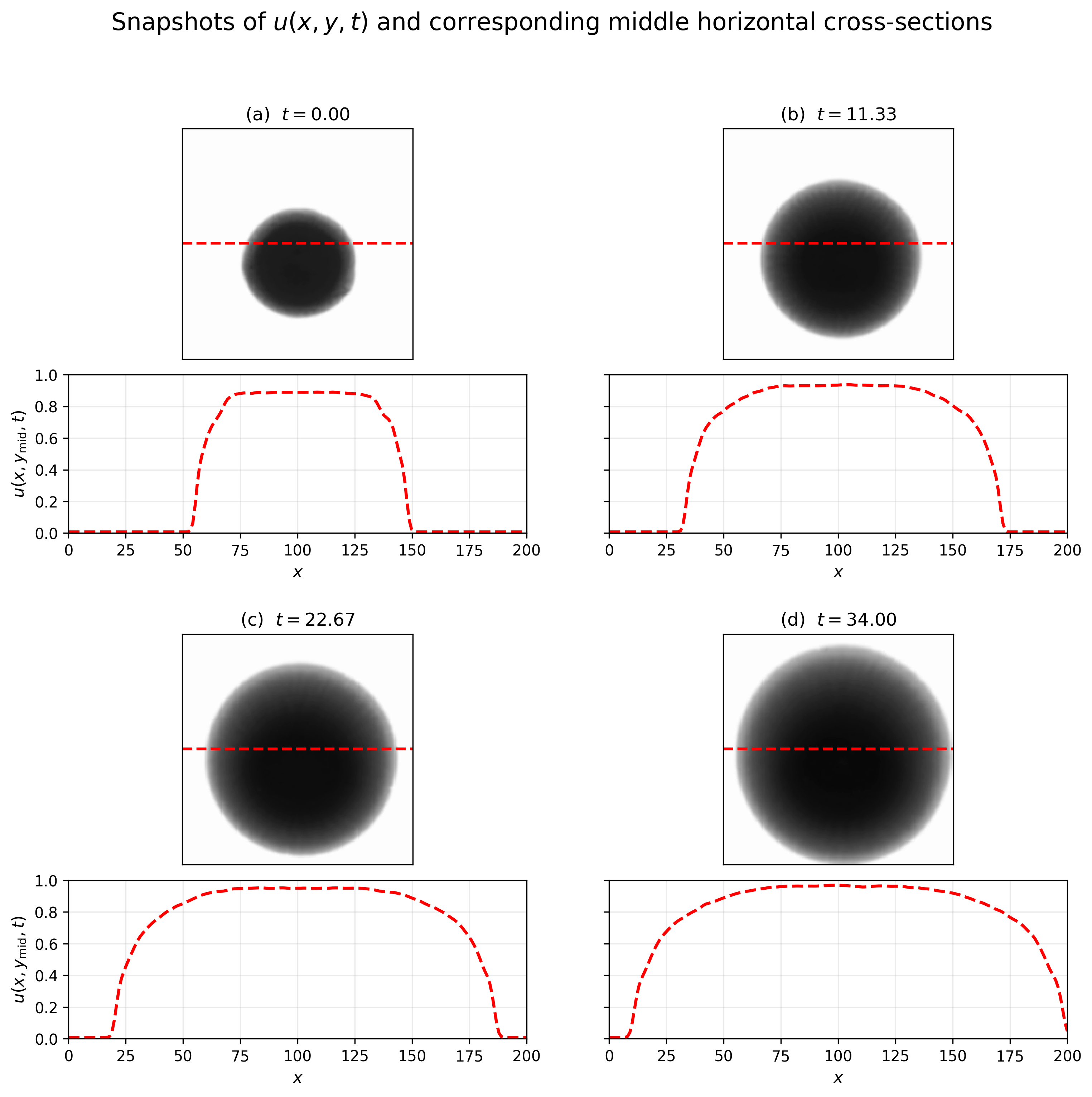}
    \caption{Snapshots of \(u(x,y,t)\) (top of each panel) and the middle horizontal cross-section (bottom). The plume undergoes spreading, deformation, and weak drift over the observation window.}
    \label{fig:plume_snapshots_cross_sections}
\end{figure}

Figure~\ref{fig:plume_snapshots_cross_sections} confirms that the processed field captures a coherent evolving plume: the field is bounded, smooth, and spatially coherent, with the plume undergoing spreading, deformation, and weak drift over the observation window. These features motivate the candidate libraries considered next, which include advection, diffusion, and nonlinear gradient-dependent terms.

\FloatBarrier
\section{Methodology}\label{sec:methodology}

\subsection{Centre-of-Mass Drift Estimation}\label{subsec:drift_estimation}

The processed plume exhibits a weak bulk drift in addition to its intrinsic spreading. To decouple the two, we estimate a time-dependent drift velocity from the image-derived field. With \(u(x,y,t)\) the processed observable, the total signal and intensity-weighted centroid are
\begin{equation}
    M(t) = \int_{\Omega} u\,dA, \qquad
    x_c(t) = \frac{\int_{\Omega} x\,u\,dA}{M(t)}, \quad
    y_c(t) = \frac{\int_{\Omega} y\,u\,dA}{M(t)}.
    \label{eq:com_continuous}
\end{equation}
\(M(t)\) is a grayscale-derived signal, not physical mass; the centroid \((x_c,y_c)\) is a threshold-free descriptor of the bulk plume location that coincides, on the discrete grid, with the standard first image moment~\citep{Hu1962}. Replacing the integrals by Riemann sums on the uniform grid yields the discrete versions, in which the area weight \(\Delta x\Delta y\) cancels in the ratios.

\begin{figure}
    \centering
    \includegraphics[width=0.3\textwidth]{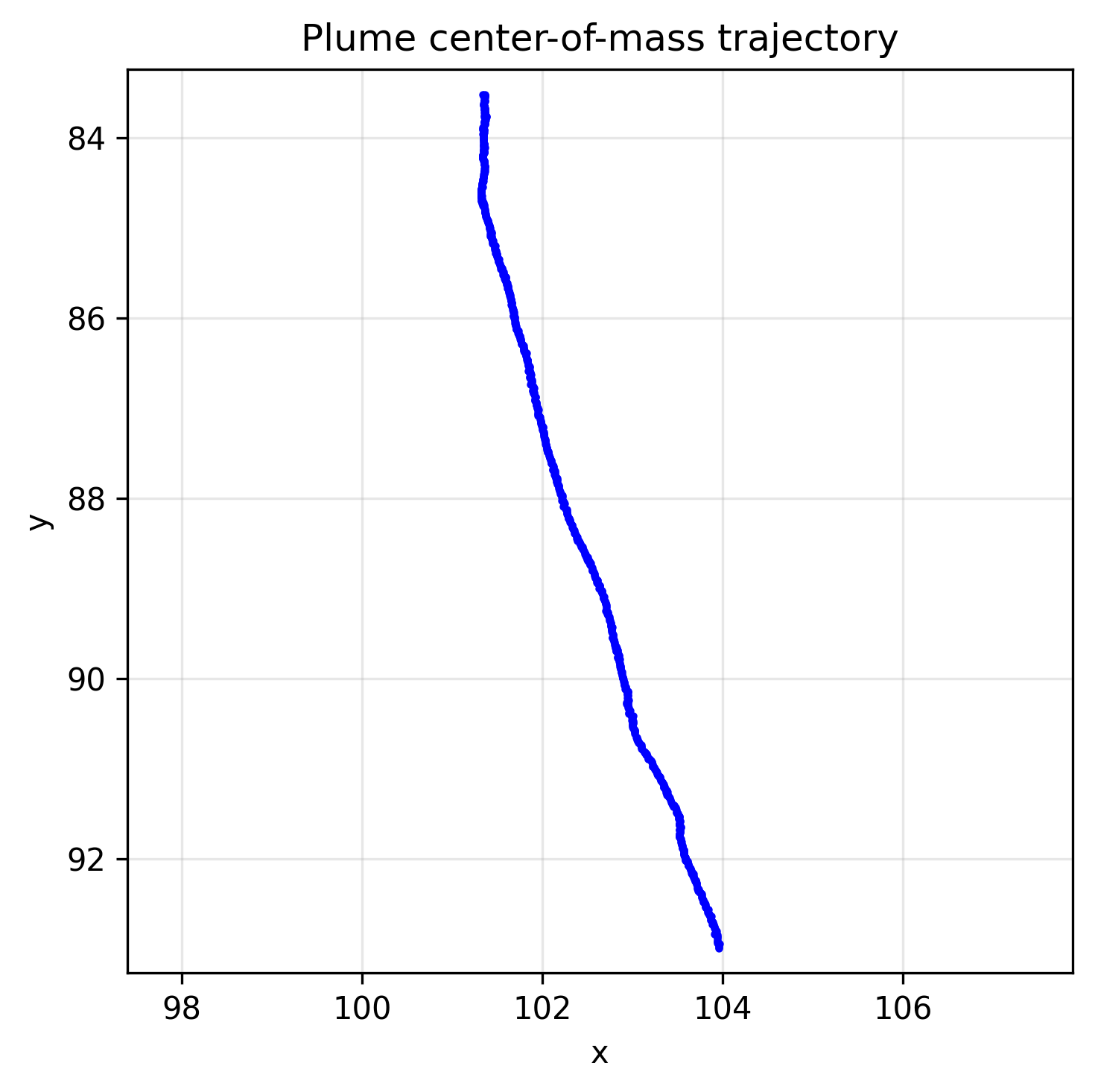}
    \caption{Intensity-weighted centroid trajectory of the processed plume; the vertical image axis follows the image-coordinate convention.}
    \label{fig:com_trajectory}
\end{figure}

Direct differentiation of the centroid trajectory amplifies frame-to-frame noise. We therefore smooth \(x_c(t)\) and \(y_c(t)\) with a Savitzky--Golay filter~\citep{SavitzkyGolay1964} of polynomial order \(p=3\) and temporal window equal to about \(8\%\) of the number of frames, then differentiate in time:
\begin{equation}
    v_x(t) = \tfrac{d}{dt}\,\mathcal{S}_{w,p}[x_c](t), \qquad
    v_y(t) = \tfrac{d}{dt}\,\mathcal{S}_{w,p}[y_c](t).
    \label{eq:sg_velocity}
\end{equation}
For the present dataset, the mean drift velocities are \(\overline{v}_x = 0.0768\) and \(\overline{v}_y = 0.2784\) image units per second, and the trajectory in Fig.~\ref{fig:com_trajectory} is essentially linear. The time-dependent fields \(v_x(t),\,v_y(t)\) are used as prescribed drift inputs in the structure-discovery, calibration, and evaluation stages that follow.

\subsection{Chronological Data Split}\label{subsec:chronological_split}

The processed sequence is split chronologically into training, validation, and test windows of \(60\%\), \(20\%\), and \(20\%\) of the total time series. The training window is used for weak-SINDy structure discovery, weak-form coefficient estimation, iPINN refinement, and bootstrap calibration; the validation window for model comparison (library choice, advection mode, post-bootstrap geometric assessment); and the test window is held out for the final rollout of the selected model. Rollouts on validation or test windows are initialised from the first frame of that window and advanced forward by the discovered PDE.

\subsection{Weak-Form Sparse Regression}\label{subsec:weak_sindy}

We seek an effective evolution law of the form
\begin{equation}
    u_t = \sum_{\ell=1}^{K}\xi_\ell\,\mathcal{F}_\ell[u;\mathbf v],
    \label{eq:pde_library_general}
\end{equation}
where \(\mathcal{F}_\ell\) are candidate features and \(\xi_\ell\) are sparse coefficients. Following \citet{Schaeffer2017IntegralSINDy} and \citet{Messenger2021WeakSINDy}, we test Eq.~\eqref{eq:pde_library_general} against localised smooth functions and assemble a weak system using integration by parts, transferring all derivatives onto the test functions and avoiding direct numerical differentiation of \(u\).

Let \(Q_{\mathrm{tr}}=\Omega\times[0,0.60T]\) denote the training space-time domain, and define the inner product
\begin{equation}
    \langle f,g\rangle_{Q_{\mathrm{tr}}} = \int_{0}^{0.60T}\!\!\int_\Omega f g\,dA\,dt,
    \label{eq:space_time_inner_product}
\end{equation}
discretised on the grid by Riemann sums with the same notation. For each space-time centre \(c_m=(x_m,y_m,t_m)\in Q_{\mathrm{tr}}\), we use a separable Gaussian test function
\begin{equation}
    \varphi_m(x,y,t) = G_{\sigma_x}(x-x_m)\,G_{\sigma_y}(y-y_m)\,G_{\sigma_t}(t-t_m),
    \qquad G_\sigma(s) = \exp\!\bigl(-s^{2}/(2\sigma^{2})\bigr).
    \label{eq:gaussian_test_function}
\end{equation}
Multiplying Eq.~\eqref{eq:pde_library_general} by \(\varphi_m\) and integrating by parts in time gives a weak regression system
\begin{equation}
    \Theta\xi \approx b, \qquad b_m = -\langle \partial_t\varphi_m, u\rangle_{Q_{\mathrm{tr}}},
    \label{eq:weak_system}
\end{equation}
whose columns are listed in Table~\ref{tab:weak_feature_construction}. We adopt the transport sign convention \(u_t + \mathbf{v}(t)\!\cdot\!\nabla u = \mathcal{N}[u]\), so the prescribed drift contributes \(-v_x(t)u_x - v_y(t)u_y\) to the right-hand side. Integration by parts is valid since each Gaussian kernel is truncated to support radius \(k_\sigma\sigma\) and centres are sampled at distance at least \(k_\sigma\sigma\) from the boundary of \(Q_{\mathrm{tr}}\). We use \(k_\sigma = 4\) and grid-relative widths \(\sigma_x = \sigma_y = 0.06\,n_{x,y}\), \(\sigma_t = 0.025\,n_t\); these values define local averaging neighbourhoods that suppress pixel-scale noise while retaining plume-scale structure.

\begin{table}[h!]
\centering
\caption{Strong-form features and their weak counterparts. \(\varphi_m\) is the Gaussian test function in Eq.~\eqref{eq:gaussian_test_function}; angle brackets denote the inner product in Eq.~\eqref{eq:space_time_inner_product}.}
\label{tab:weak_feature_construction}
\small
\begin{tabular}{lll}
\toprule
\textbf{Feature} & \textbf{Strong form} & \textbf{Weak column} \\
\midrule
Constant & \(1\) & \(\langle \varphi_m,1\rangle\) \\
Linear & \(u\) & \(\langle \varphi_m,u\rangle\) \\
Quadratic & \(u^{2}\) & \(\langle \varphi_m,u^{2}\rangle\) \\
Gradient energy & \(|\nabla u|^{2}\) & \(\langle \varphi_m,u_x^{2}+u_y^{2}\rangle\) \\
Weighted gradient energy & \(u|\nabla u|^{2}\) & \(\langle \varphi_m,u(u_x^{2}+u_y^{2})\rangle\) \\
Laplacian & \(\Delta u\) & \(-\langle \partial_x\varphi_m,u_x\rangle -\langle \partial_y\varphi_m,u_y\rangle\) \\
\(x\)-advection & \(-v_x(t)u_x\) & \(\langle \partial_x\varphi_m,v_x(t)\,u\rangle\) \\
\(y\)-advection & \(-v_y(t)u_y\) & \(\langle \partial_y\varphi_m,v_y(t)\,u\rangle\) \\
\bottomrule
\end{tabular}
\end{table}

A useful computational consequence of Gaussian test functions is that each weak inner product reduces to a separable convolution, evaluated once per feature with derivative-of-Gaussian kernels and then sampled at the chosen centres. Sparse coefficients are obtained by sequentially thresholded least squares: at each iteration we solve a ridge-regularised system on the active set,
\begin{equation}
    \xi^{(r)} = \arg\min_{\xi}\bigl\{\|\Theta_{\mathcal{A}^{(r)}}\xi - b\|_2^{2} + \alpha\|\xi\|_2^{2}\bigr\},
    \label{eq:ridge_step}
\end{equation}
remove coefficients of magnitude below \(\lambda\), and iterate until the active set stabilises. Numerical settings are listed in Table~\ref{tab:weak_sindy_settings}; the candidate libraries and identifiability diagnostics follow.

\begin{table}[h!]
\centering
\caption{Weak-form regression settings on the training window.}
\label{tab:weak_sindy_settings}
\begin{tabular}{ll}
\toprule
\textbf{Quantity} & \textbf{Value} \\
\midrule
Number of test functions & \(M=2000\) \\
Interior support parameter & \(k_\sigma=4.0\) \\
Spatial Gaussian widths & \(\sigma_x=\sigma_y=0.06\,n_{x,y}\) \\
Temporal Gaussian width & \(\sigma_t=0.025\,n_t\) \\
STLSQ threshold & \(\lambda=10^{-3}\) \\
Ridge regularisation & \(\alpha=10^{-6}\) \\
Maximum STLSQ iterations & \(100\) \\
Advection treatment & measured \(v_x(t),v_y(t)\) \\
\bottomrule
\end{tabular}
\end{table}

\subsection{Candidate Libraries and Identifiability Diagnostics}\label{subsec:candidate_libraries}

Because the effective equation for the image-derived plume is not known a priori, we test several nested libraries (Table~\ref{tab:candidate_libraries}). All include the prescribed drift terms; the non-advection content varies from a pure advection--diffusion kernel (A), through a linear reaction term (B), to nonlinear gradient-dependent corrections (C, C-alt, C-both), with an overcomplete reference library at the top (Full).

\begin{table}[h!]
\centering
\caption{Candidate libraries. The drift terms \(-v_x u_x\) and \(-v_y u_y\) are included in every library when drift velocities are supplied.}
\label{tab:candidate_libraries}
\small
\begin{tabular*}{\linewidth}{@{\extracolsep{\fill}}ll@{}}
\toprule
\textbf{Library} & \textbf{Non-advection candidate terms} \\
\midrule
A: advection--diffusion & \(\Delta u\) \\
B: advection--diffusion \(+\,u\) & \(u,\ \Delta u\) \\
C: advection--diffusion \(+\,|\nabla u|^{2}\) & \(|\nabla u|^{2},\ \Delta u\) \\
C-alt: advection--diffusion \(+\,u|\nabla u|^{2}\) & \(u|\nabla u|^{2},\ \Delta u\) \\
C-both: advection--diffusion \(+\) both gradient terms & \(|\nabla u|^{2},\ u|\nabla u|^{2},\ \Delta u\) \\
Full library & \(1,\ u,\ u^{2},\ |\nabla u|^{2},\ u|\nabla u|^{2},\ \Delta u\) \\
\bottomrule
\end{tabular*}
\end{table}

For each library we compute three diagnostics on the training window. The condition number \(\kappa(\Theta)=\sigma_{\max}/\sigma_{\min}\) flags ill-conditioning. The normalised column-correlation matrix
\(R_{ij}=\Theta_i^{\mathsf T}\Theta_j/(\|\Theta_i\|_2\|\Theta_j\|_2)\)
identifies near-collinear features. A logarithmic threshold sweep \(\lambda_m=10^{-5+(m-1)/4}\) for \(m=1,\dots,21\) traces the support of \(\xi(\lambda)\); terms persisting over a wide range of \(\lambda\) are interpreted as structurally robust.

Because the weak system is built from random test-function centres, we further test selection stability over \(N_{\mathrm{runs}}=100\) independent centre samples (\(M_{\mathrm{stab}}=1000\) test functions each) and record the selection frequency
\begin{equation}
    p_\ell = \frac{1}{N_{\mathrm{runs}}}\sum_{r=1}^{N_{\mathrm{runs}}}\mathbf{1}\bigl\{\mathcal{F}_\ell\text{ selected in run }r\bigr\}
    \label{eq:selection_frequency}
\end{equation}
together with the mean and standard deviation of the selected coefficients. Settings used for the threshold sweep and the stability study are listed in Table~\ref{tab:diagnostic_settings}.

\begin{table}[h!]
\centering
\caption{Settings for the threshold sweep and the random-centre stability study.}
\label{tab:diagnostic_settings}
\begin{tabular}{ll}
\toprule
\textbf{Quantity} & \textbf{Value} \\
\midrule
Threshold sweep values & \(\lambda_m = 10^{-5+(m-1)/4},\ m=1,\dots,21\) \\
Sweep ridge parameter & \(\alpha=10^{-6}\) \\
Sweep maximum STLSQ iterations & \(100\) \\
Active-term tolerance & \(\varepsilon_{\mathrm{act}}=10^{-12}\) \\
Test functions per stability run & \(M_{\mathrm{stab}}=1000\) \\
Number of stability runs & \(N_{\mathrm{runs}}=100\) \\
Stability STLSQ threshold & \(\lambda=10^{-3}\) \\
Stability ridge parameter & \(\alpha=10^{-6}\) \\
Advection in stability study & measured \(v_x(t),v_y(t)\) \\
\bottomrule
\end{tabular}
\end{table}

\subsection{Rollout Evaluation: Learned versus Measured Advection}\label{subsec:validation_modes}

The centroid trajectory in Fig.~\ref{fig:com_trajectory} provides a direct estimate of the bulk drift, which we can either impose with unit coefficient or let the regression rescale. Two advection modes are therefore compared in validation rollouts:
\begin{align}
    \text{Mode B (measured):}&\quad u_t = -v_x u_x - v_y u_y + \mathcal{N}[u], \label{eq:measured_advection_validation}\\
    \text{Mode A (learned):}&\quad u_t = -c_x v_x u_x - c_y v_y u_y + \mathcal{N}[u], \label{eq:learned_advection_validation}
\end{align}
where \(\mathcal{N}[u]\) collects the non-advection terms selected by the discovered model and \((c_x,c_y)\) are coefficients estimated by weak-SINDy. Comparing the two modes tests whether the regression preserves, rescales, or compensates for the prescribed drift.

For both modes we report a full rollout (initialised once from the first frame of the evaluation window and advanced over its full length) and a one-step rollout (each observed frame predicts the next). The full rollout measures accumulated error; the one-step rollout isolates local model error from long-time accumulation. Predictions are integrated with explicit substepping (safety factor \(0.25\), at most \(2000\) substeps per frame interval), a small numerical diffusion \(\varepsilon_{\mathrm{visc}}=0.01\), and clipping to \([0,1]\) when the data are normalised.

Prediction accuracy on an evaluation window \(\mathcal{W}\) is the relative root-mean-square error
\begin{equation}
    \mathrm{rRMSE}_{\mathcal{W}}
    = \left(\frac{\sum_{k,j,i}(U_{\mathrm{pred},j,i}^{k}-U_{\mathcal{W},j,i}^{k})^{2}}
    {\sum_{k,j,i}(U_{\mathcal{W},j,i}^{k})^{2}}\right)^{1/2}.
    \label{eq:relative_rmse_validation}
\end{equation}
We complement Eq.~\eqref{eq:relative_rmse_validation} with the centroid error and the equivalent-front-radius error at five thresholds \(\gamma\in\{0.05,0.10,0.15,0.20,0.25\}\), since pixel-wise accuracy alone need not preserve the bulk trajectory or the plume front. Validation errors drive model selection; test-window errors are reported only after the final model has been frozen. The settings are summarised in Table~\ref{tab:validation_settings}.

\begin{table}[h!]
\centering
\caption{Rollout numerical settings used for validation and test evaluation.}
\label{tab:validation_settings}
\begin{tabular}{ll}
\toprule
\textbf{Quantity} & \textbf{Value} \\
\midrule
Training/validation/test split & \(60\%/20\%/20\%\) \\
Measured-advection coefficient (Mode B) & \(1.0\) \\
Numerical diffusion & \(\varepsilon_{\mathrm{visc}}=0.01\) \\
Safety factor & \(0.25\) \\
Maximum substeps per frame interval & \(2000\) \\
Active coefficient tolerance & \(10^{-12}\) \\
Automatic clipping & \([0,1]\) \\
Front-radius levels & \(0.05,\,0.10,\,0.15,\,0.20,\,0.25\) \\
Selection metric & validation rollout rRMSE \\
Reporting metric & test rollout rRMSE, plus geometric diagnostics \\
\bottomrule
\end{tabular}
\end{table}

\subsection{iPINN Coefficient Refinement}\label{subsec:ipinn_refinement}

Once weak-SINDy has identified candidate structures, we refine the intrinsic coefficients of the two leading nonlinear-gradient libraries with an inverse physics-informed neural network~\citep{Raissi2019PINNs,Karniadakis2021PINNs,Cuomo2022PINNReview}. The network represents \(u\) by \(\widehat{u}_\theta(x,y,t)\), a fully connected \(\tanh\) MLP with sigmoid output (so the prediction stays in the normalised range), with derivatives taken by automatic differentiation. The PDE structure is fixed; only the coefficients \(a\) and \(\beta\) are trainable, and the prescribed drift is held with unit coefficient. The residuals for the two libraries are
\begin{align}
\mathcal{R}_C[\widehat{u}_\theta;a,\beta] &= \partial_t\widehat{u}_\theta + v_x\partial_x\widehat{u}_\theta + v_y\partial_y\widehat{u}_\theta - a|\nabla\widehat{u}_\theta|^{2} - \beta\Delta\widehat{u}_\theta,\label{eq:ipinn_residual_C}\\
\mathcal{R}_{C\text{-alt}}[\widehat{u}_\theta;a,\beta] &= \partial_t\widehat{u}_\theta + v_x\partial_x\widehat{u}_\theta + v_y\partial_y\widehat{u}_\theta - a\,\widehat{u}_\theta|\nabla\widehat{u}_\theta|^{2} - \beta\Delta\widehat{u}_\theta.\label{eq:ipinn_residual_Calt}
\end{align}

Training proceeds in two phases. In pretraining the network minimises the data loss
\(\mathcal{L}_{\mathrm{data}}(\theta) = N_d^{-1}\sum_{q}|\widehat{u}_\theta(x_q,y_q,t_q) - U_q|^{2}\)
on training samples; in joint training the network and coefficients are updated against
\begin{equation}
    \mathcal{L}_{\mathrm{iPINN}} = w_d\,\mathcal{L}_{\mathrm{data}} + w_p\,\mathcal{L}_{\mathrm{PDE}},
    \qquad
    \mathcal{L}_{\mathrm{PDE}} = N_p^{-1}\sum_{r}|\mathcal{R}[\widehat{u}_\theta;a,\beta](x_r,y_r,t_r)|^{2},
    \label{eq:ipinn_total_loss}
\end{equation}
initialised from the weak-SINDy estimates \(a^{(0)}=\xi^{\mathrm{WS}}\), \(\beta^{(0)}=\xi^{\mathrm{WS}}_{\Delta u}\). The weak-SINDy values serve as initialisation only and are not penalised in the loss; this matters because PINN training is known to be sensitive to loss-balance and to suffer characteristic failure modes when residual penalties dominate~\citep{Krishnapriyan2021PINNFailureModes}, and a soft prior tied to the weak coefficient would inherit those biases. The chronological split of \S\ref{subsec:chronological_split} is respected throughout. Network and optimisation settings are listed in Table~\ref{tab:ipinn_settings}; the iPINN outputs \(\{a^{\mathrm{iPINN}},\beta^{\mathrm{iPINN}}\}\) are then carried into the bootstrap stage as one of two initialisation sources.

\begin{table}[h!]
\centering
\caption{iPINN settings for libraries C and C-alt.}
\label{tab:ipinn_settings}
\begin{tabular}{ll}
\toprule
\textbf{Quantity} & \textbf{Value} \\
\midrule
Refined libraries / coefficients & C, C-alt; \(a,\beta\) \\
Advection treatment & fixed measured drift, coefficient \(1.0\) \\
Network architecture & \(\tanh\) MLP, sigmoid output \\
Hidden width / depth & \(128\) / \(6\) \\
Pretraining steps / joint training steps & \(5000\) / \(12000\) \\
Data / PDE samples per step & \(8192\) / \(4096\) \\
Pretraining / joint learning rate & \(8\times 10^{-4}\) / \(2\times 10^{-4}\) \\
Data / PDE / prior weight & \(1.0\) / \(10^{-3}\) / \(0\) \\
Random seed & \(2026\) \\
\bottomrule
\end{tabular}
\end{table}

\subsection{Bootstrap Rollout Calibration}\label{subsec:bootstrap_calibration}

The iPINN refinement improves coefficients against a continuous PDE residual; the final stage calibrates them against the rollout used for prediction. Both leading models are recalibrated with the structure fixed and the prescribed drift held at unit coefficient:
\begin{align}
\text{C:}\quad u_t + \mathbf v(t)\!\cdot\!\nabla u &= a|\nabla u|^{2} + \beta\Delta u, \label{eq:bootstrap_model_C}\\
\text{C-alt:}\quad u_t + \mathbf v(t)\!\cdot\!\nabla u &= a\,u|\nabla u|^{2} + \beta\Delta u. \label{eq:bootstrap_model_Calt}
\end{align}
\(\beta\) is left unconstrained at this stage and may take either sign. For each model we compare two initialisations -- the weak-SINDy/STLSQ estimates and the iPINN-refined values -- to test whether the rollout-calibrated coefficients are sensitive to the starting point.

Because the frames are temporally ordered, we use a chronological block bootstrap rather than independent frame-level resampling~\citep{Kunsch1989,LiuSingh1992,PolitisWhite2004}. Contiguous blocks of length \(L_b=\lfloor\sqrt{n_{\mathrm{tr}}}\rceil\) are sampled with replacement from the training interval and concatenated to length \(n_{\mathrm{tr}}\); the resampled data and drift values are
\(U^{(r)} = U_{\mathrm{tr}}(:,:,\mathcal{I}^{(r)}),\;v^{(r)} = v_{\mathrm{tr}}(\mathcal{I}^{(r)})\),
with the bootstrap index set \(\mathcal{I}^{(r)}\). For each replicate, the coefficients \(\theta=(a,\beta)\) are refit by minimising the rollout MSE
\begin{equation}
    J^{(r)}(\theta) = \frac{1}{|\mathcal{Q}^{(r)}|}\sum_{(j,i,k)\in\mathcal{Q}^{(r)}}\bigl(U^{(r),k}_{\theta,j,i}-U^{(r),k}_{j,i}\bigr)^{2},
    \label{eq:bootstrap_rollout_objective}
\end{equation}
on a subsampled set \(\mathcal{Q}^{(r)}\) of at most \(10^{5}\) space-time points. Because each evaluation requires a forward solve with substepping and clipping, we use the derivative-free Nelder--Mead method~\citep{NelderMead1965,GaoHan2012,Larson2019DFO}. After \(B=50\) replicates we summarise each coefficient by its empirical median and \([q_{0.025},q_{0.975}]\) interval, and assess the median model on the validation window via Eq.~\eqref{eq:relative_rmse_validation} before any test-window evaluation. Settings are collected in Table~\ref{tab:bootstrap_settings}.

\begin{table}[h!]
\centering
\caption{Block-bootstrap rollout-calibration settings for libraries C and C-alt.}
\label{tab:bootstrap_settings}
\begin{tabular}{ll}
\toprule
\textbf{Quantity} & \textbf{Value} \\
\midrule
Calibrated libraries / coefficients & C, C-alt; \(a,\beta\) \\
Initialisation sources & weak-SINDy/STLSQ; iPINN \\
Bootstrap replicates & \(B=50\) \\
Block length & \(L_b\approx\sqrt{n_{\mathrm{tr}}}\) \\
Rollout substeps for fitting & \(100\) \\
Maximum objective points & \(10^{5}\) \\
Maximum Nelder--Mead iterations & \(300\) \\
Numerical diffusion / clipping & \(\varepsilon_{\mathrm{visc}}=0.01\) / \([0,1]\) \\
Random seed & \(42\) \\
\bottomrule
\end{tabular}
\end{table}

\subsection{Geometric Diagnostics and Front-Aware Recalibration}\label{subsec:post_bootstrap_front_aware}

Pixel-wise rRMSE is supplemented by two geometric diagnostics. The centroid error
\begin{equation}
    e_{\mathrm{COM}}(t_k) = \bigl[(x_c^{\mathrm{pred}}-x_c^{\mathrm{true}})^{2} + (y_c^{\mathrm{pred}}-y_c^{\mathrm{true}})^{2}\bigr]^{1/2}
    \label{eq:com_error_post_boot}
\end{equation}
measures whether the predicted plume follows the observed bulk trajectory. The equivalent-front-radius error is computed from the superlevel sets \(\Omega_\gamma(t)=\{u\ge\gamma\}\) via \(r_\gamma(t)=\sqrt{|\Omega_\gamma(t)|/\pi}\); lower thresholds track the faint outer plume, higher thresholds the concentrated core.

To test whether front behaviour can drive the calibration, we also recalibrate the C-family models against a front-aware objective
\begin{equation}
    J_{\mathrm{front}}(\theta) = J_{\mathrm{pix}}(\theta) + w_f J_{\mathrm{radius}}(\theta) + w_g J_{\mathrm{growth}}(\theta),
    \qquad w_f=5.0,\ w_g=0.05,
    \label{eq:front_aware_objective}
\end{equation}
which adds penalties for front-radius mismatch and front-growth mismatch. Because a physically diffusive contribution requires \(\beta>0\), we additionally consider a constrained variant in which \(\beta\) is optimised in log-space; an unconstrained fit that lowers front-radius error by selecting \(\beta<0\) is treated as a diagnostic rather than as a physically admissible model. The post-bootstrap comparison therefore uses four criteria -- validation rRMSE, front-radius error, centroid error, and the sign of the Laplacian coefficient -- and the selected model is frozen before any test-window evaluation.

\FloatBarrier
\section{Results}\label{sec:results}

\subsection{Library Conditioning and Threshold Sweeps}\label{subsec:library_conditioning_thresholds}

\begin{figure}
\centering
\begin{subfigure}{0.32\textwidth}\centering
\includegraphics[width=\linewidth]{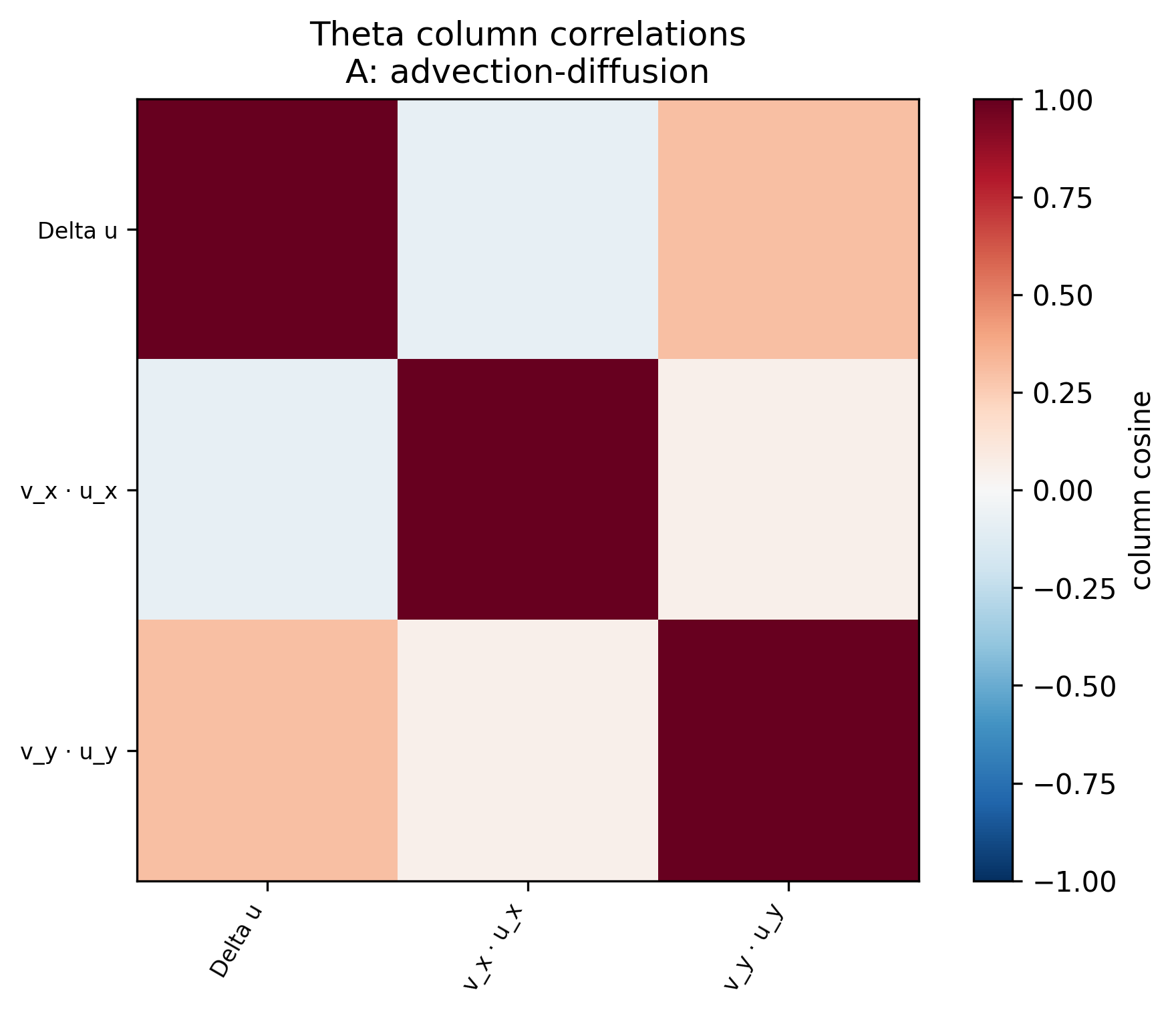}\caption{A}\end{subfigure}\hfill
\begin{subfigure}{0.32\textwidth}\centering
\includegraphics[width=\linewidth]{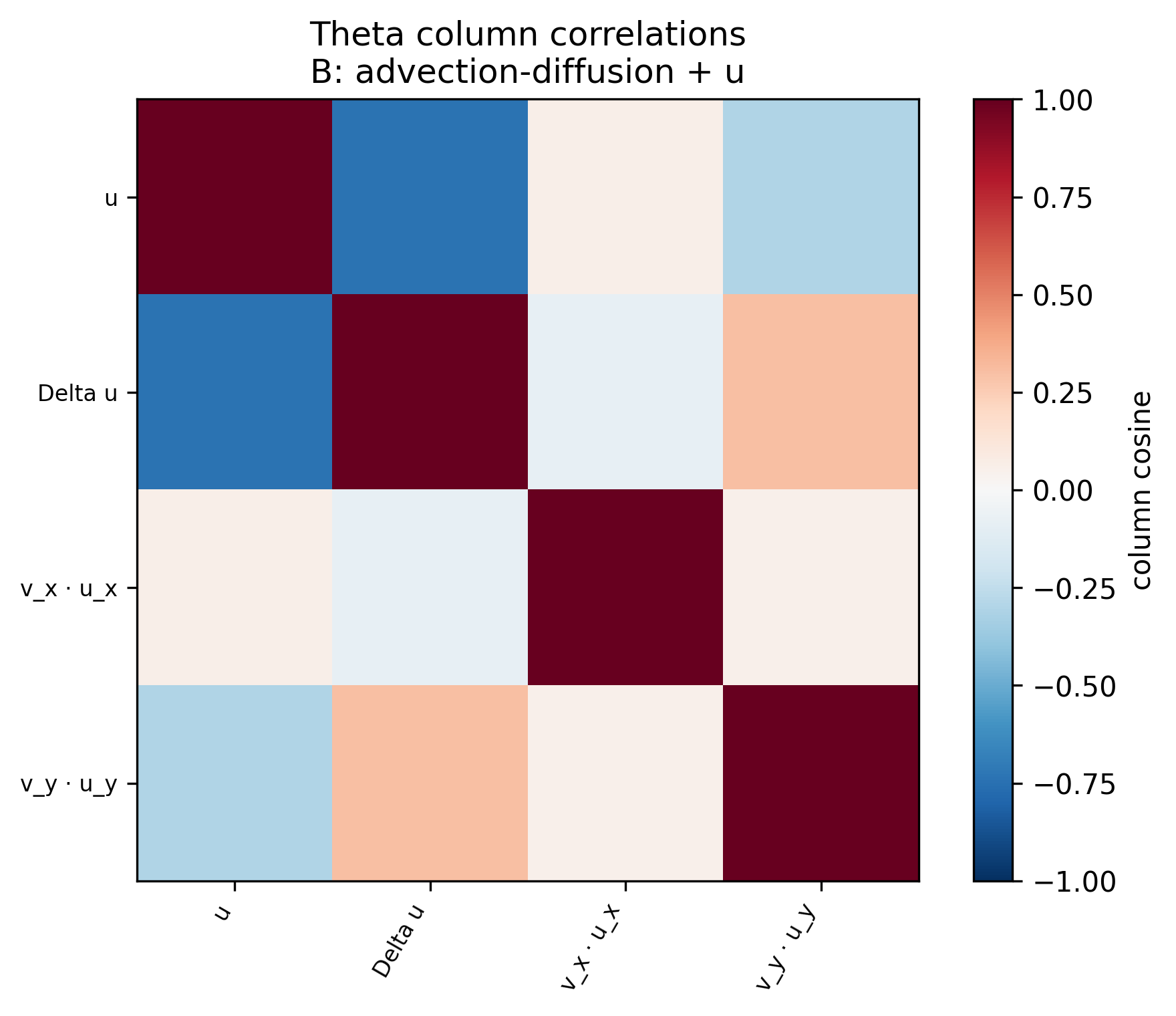}\caption{B}\end{subfigure}\hfill
\begin{subfigure}{0.32\textwidth}\centering
\includegraphics[width=\linewidth]{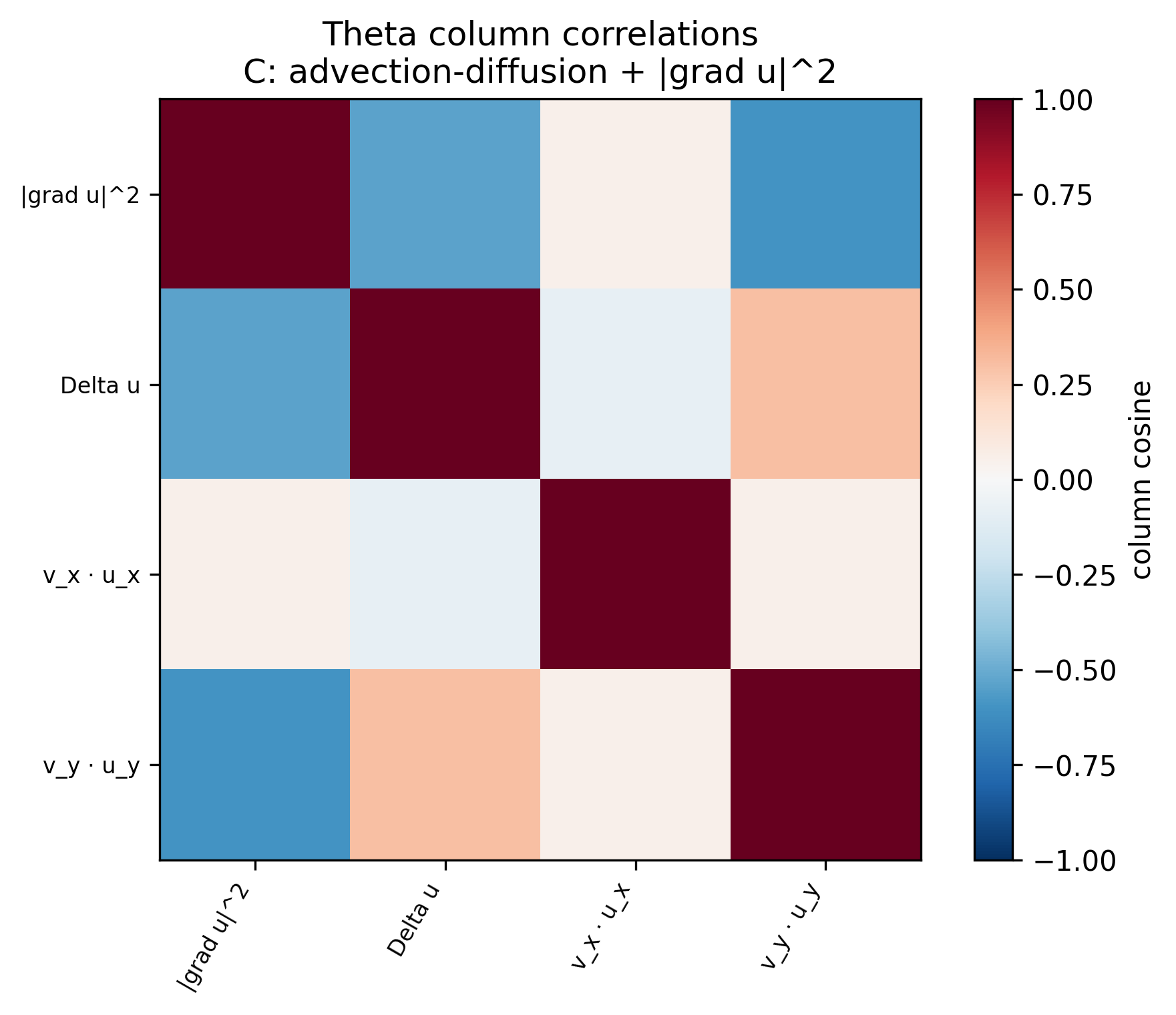}\caption{C}\end{subfigure}

\vspace{0.5em}

\begin{subfigure}{0.32\textwidth}\centering
\includegraphics[width=\linewidth]{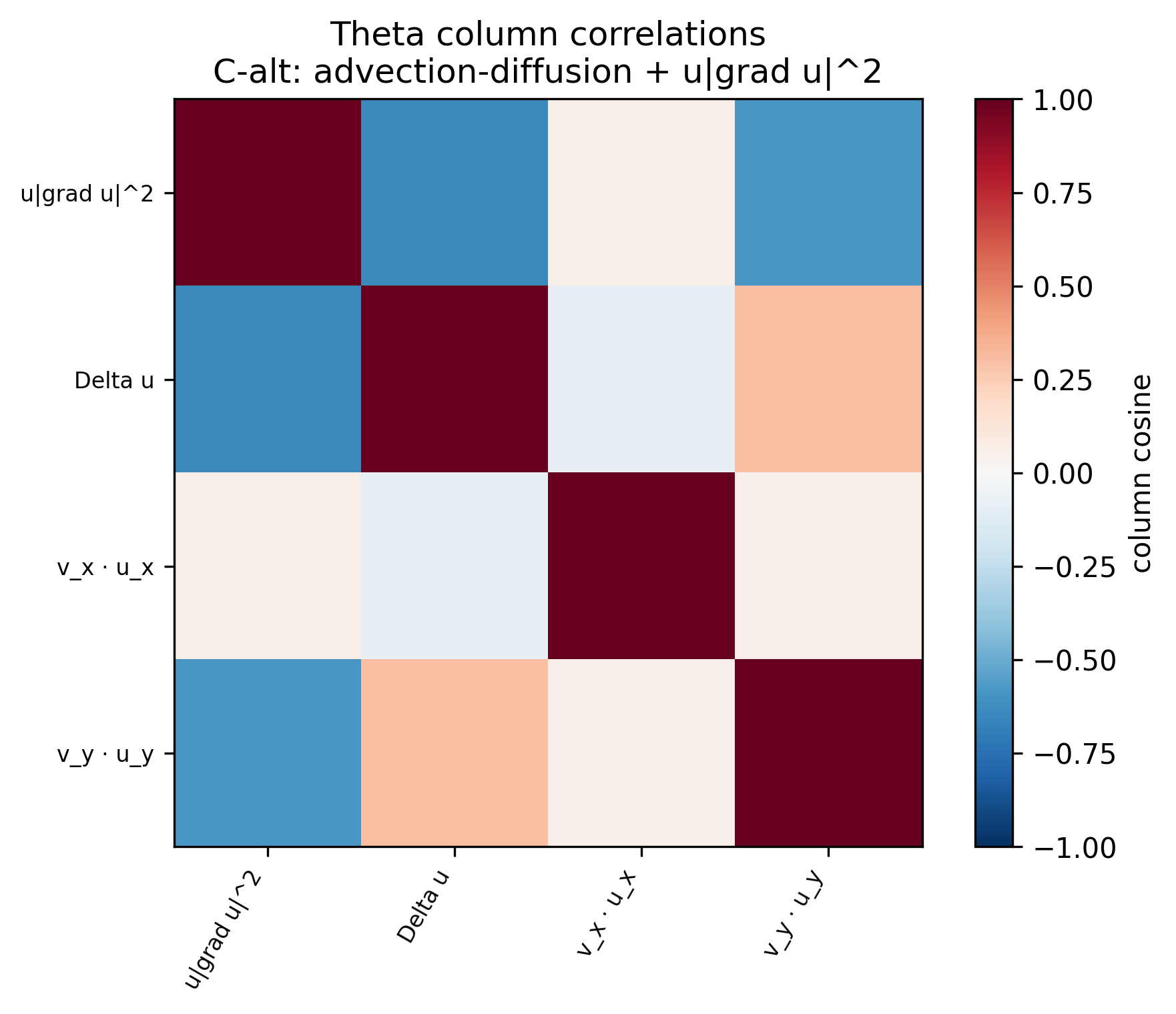}\caption{C-alt}\end{subfigure}\hfill
\begin{subfigure}{0.32\textwidth}\centering
\includegraphics[width=\linewidth]{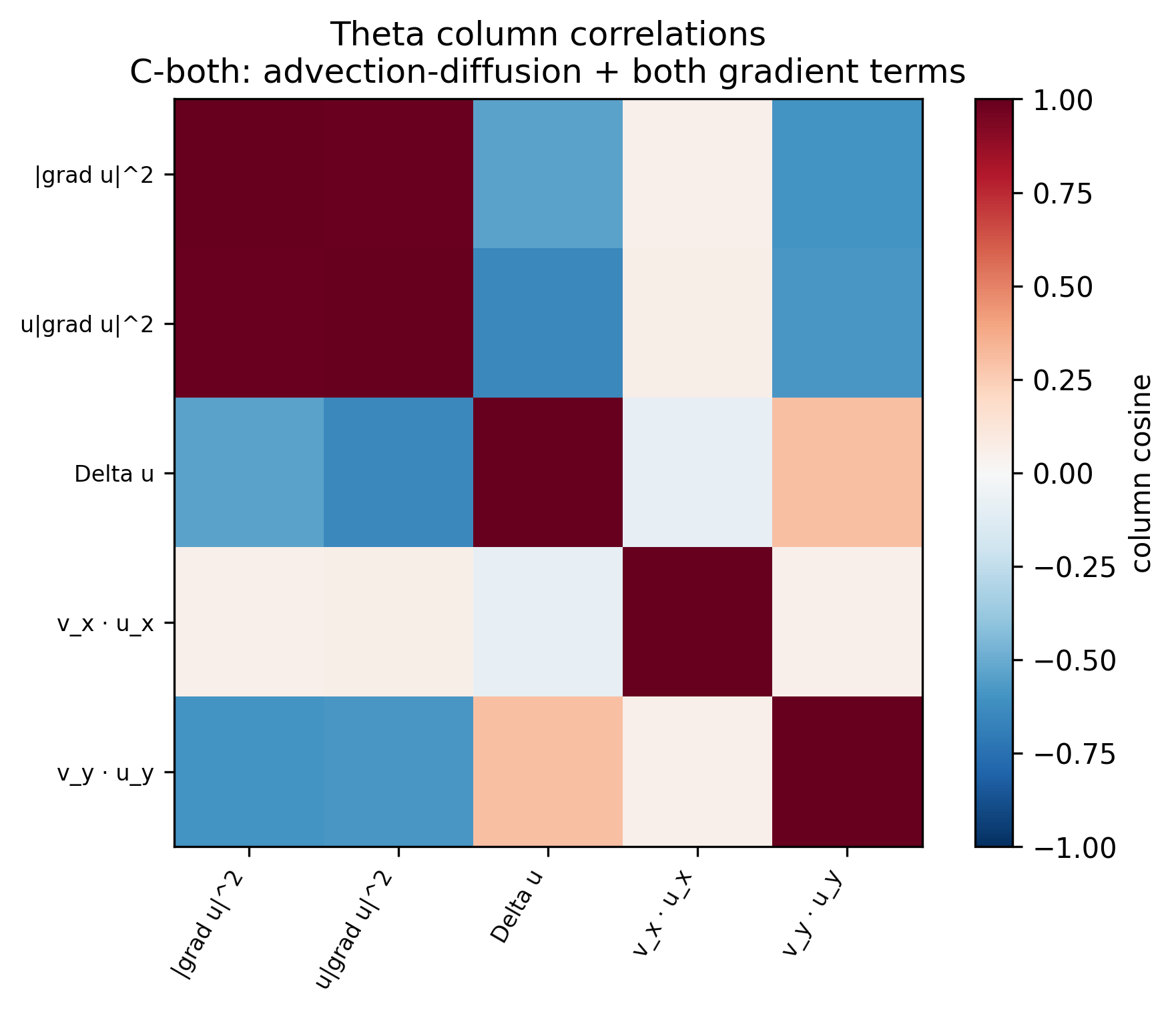}\caption{C-both}\end{subfigure}\hfill
\begin{subfigure}{0.32\textwidth}\centering
\includegraphics[width=\linewidth]{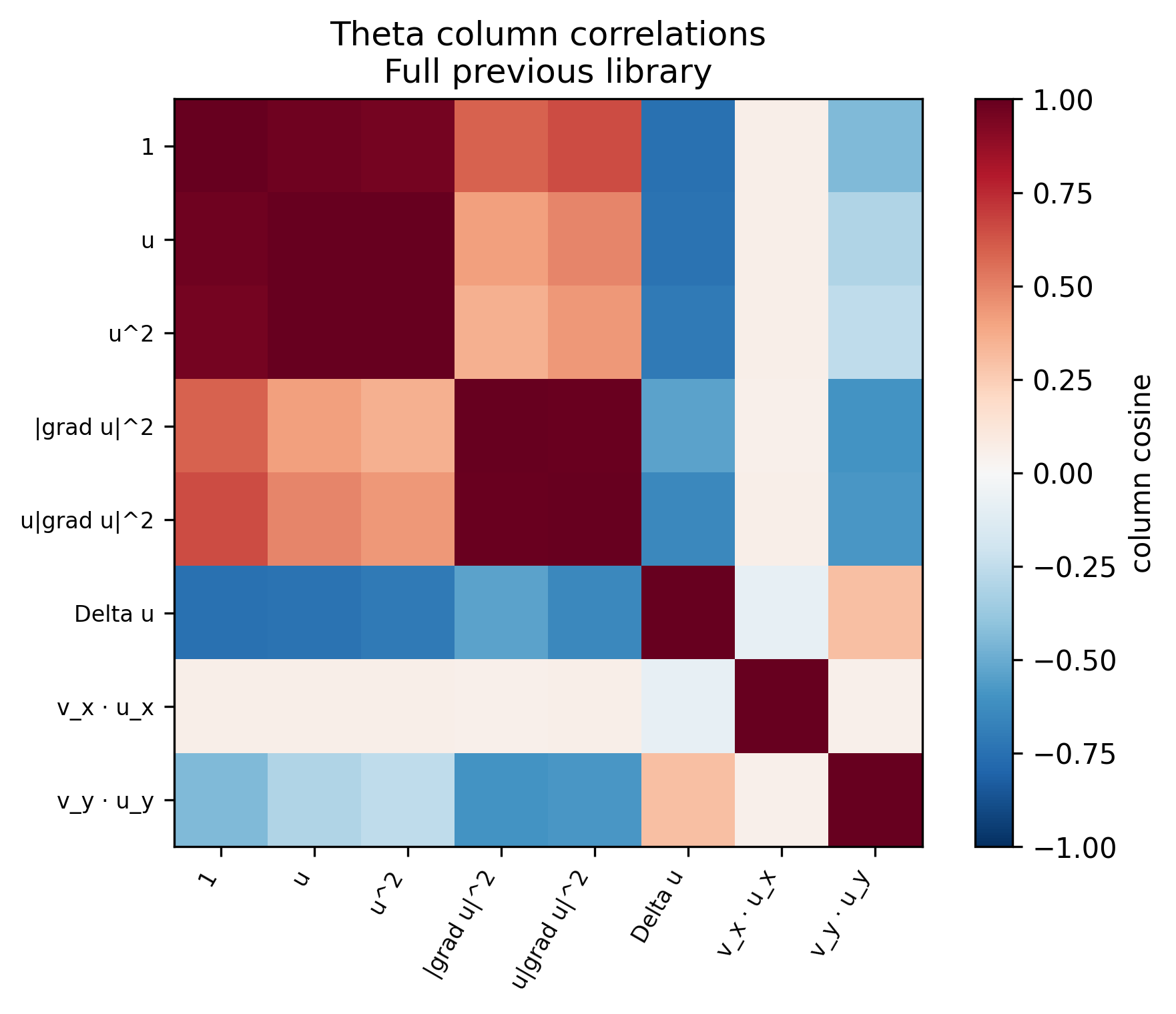}\caption{Full}\end{subfigure}
\caption{Column-correlation heatmaps of the weak feature matrix \(\Theta\). The full library exhibits strong correlations among constant, \(u\), \(u^{2}\), and nonlinear gradient terms, explaining its large condition number; the reduced libraries show clearer feature separation.}
\label{fig:theta_correlation_heatmaps}
\end{figure}

The reduced libraries are far better conditioned than the full library: \(\kappa(\Theta)\le 20\) for A, C, and C-alt, but \(\kappa(\Theta)\sim 10^{5}\) for the full library (Table~\ref{tab:library_conditioning_summary}). The full library also retains several low-order intensity terms together with nonlinear gradient terms in its initial sparse fit, signalling that the fit is dominated by feature collinearity rather than dynamical content. The column-correlation heatmaps (Fig.~\ref{fig:theta_correlation_heatmaps}) tell the same story: B becomes ill-conditioned once the linear \(u\) term is added, and C-both is less stable than C or C-alt because \(|\nabla u|^{2}\) and \(u|\nabla u|^{2}\) are strongly correlated.

\begin{table}[h!]
\centering
\caption{Conditioning and active terms of the initial weak-SINDy fits.}
\label{tab:library_conditioning_summary}
\scriptsize
\begin{tabular*}{\linewidth}{@{\extracolsep{\fill}}lcl@{}}
\toprule
\textbf{Library} & \(\boldsymbol{\kappa(\Theta)}\) & \textbf{Active terms} \\
\midrule
Full library & \(3.52\times 10^{5}\) & \(1,\ u,\ u^{2},\ |\nabla u|^{2},\ u|\nabla u|^{2},\ v_xu_x,\ v_yu_y\) \\
A: advection--diffusion & \(4.22\) & \(\Delta u,\ v_yu_y\) \\
B: advection--diffusion \(+u\) & \(1.77\times 10^{3}\) & \(\Delta u,\ v_yu_y\) \\
C: advection--diffusion \(+|\nabla u|^{2}\) & \(6.36\) & \(|\nabla u|^{2},\ \Delta u,\ v_xu_x,\ v_yu_y\) \\
C-alt: advection--diffusion \(+u|\nabla u|^{2}\) & \(1.91\times 10^{1}\) & \(u|\nabla u|^{2},\ v_xu_x,\ v_yu_y\) \\
C-both: advection--diffusion \(+\) both gradient terms & \(5.12\times 10^{2}\) & \(|\nabla u|^{2},\ \Delta u,\ v_xu_x,\ v_yu_y\) \\
\bottomrule
\end{tabular*}
\end{table}

\begin{figure}
\centering
\begin{subfigure}{0.32\textwidth}\centering
\includegraphics[width=\linewidth]{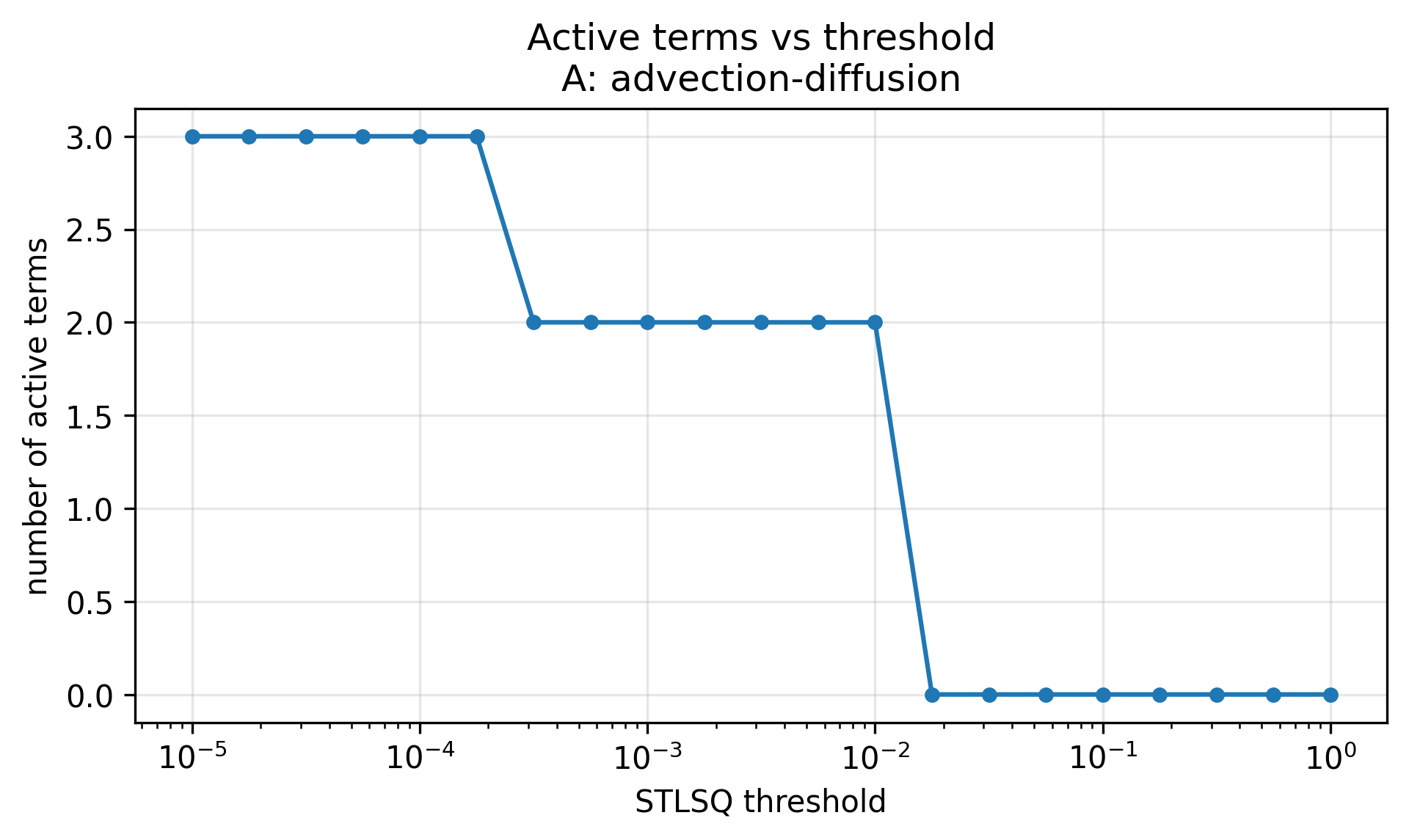}\caption{A}\end{subfigure}\hfill
\begin{subfigure}{0.32\textwidth}\centering
\includegraphics[width=\linewidth]{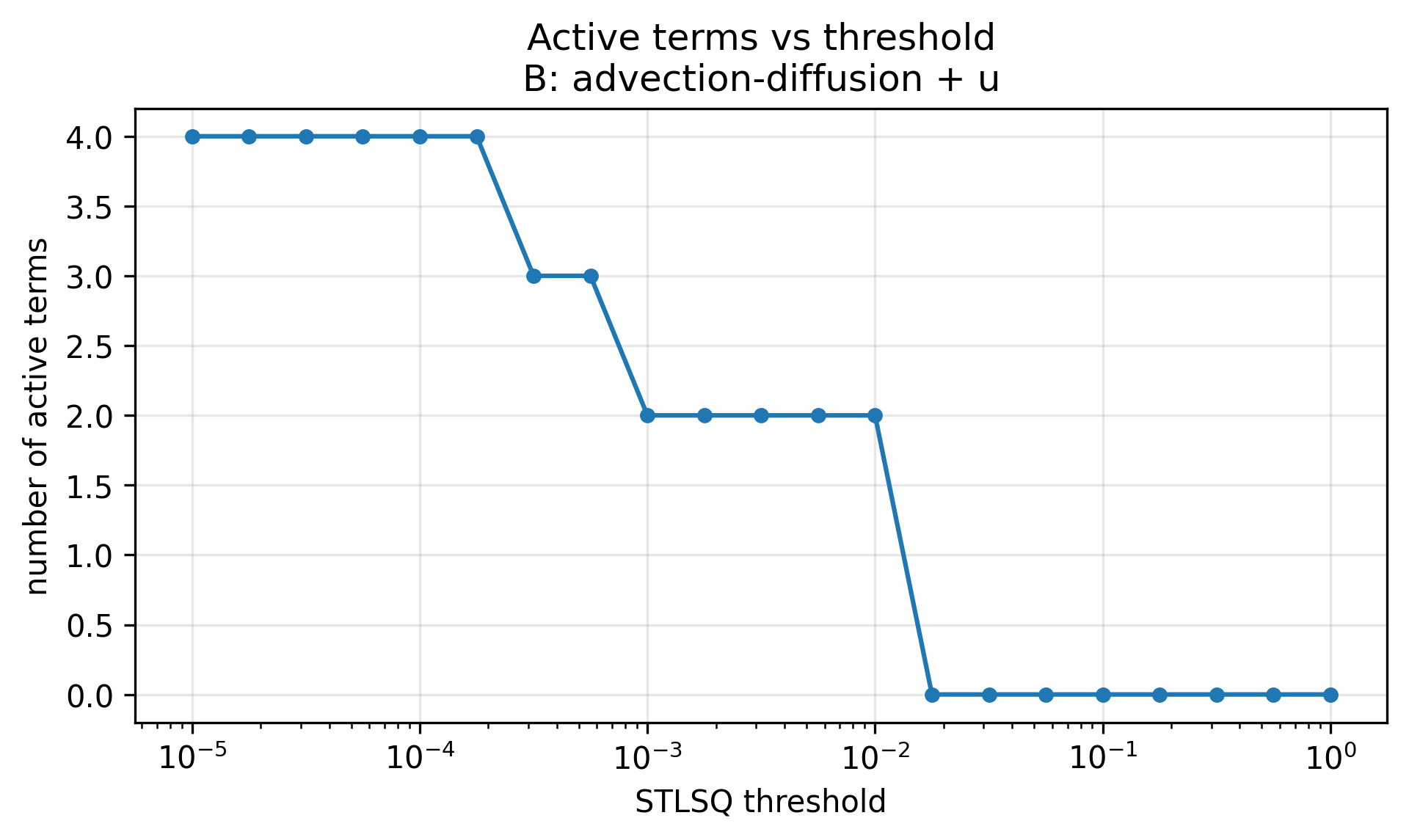}\caption{B}\end{subfigure}\hfill
\begin{subfigure}{0.32\textwidth}\centering
\includegraphics[width=\linewidth]{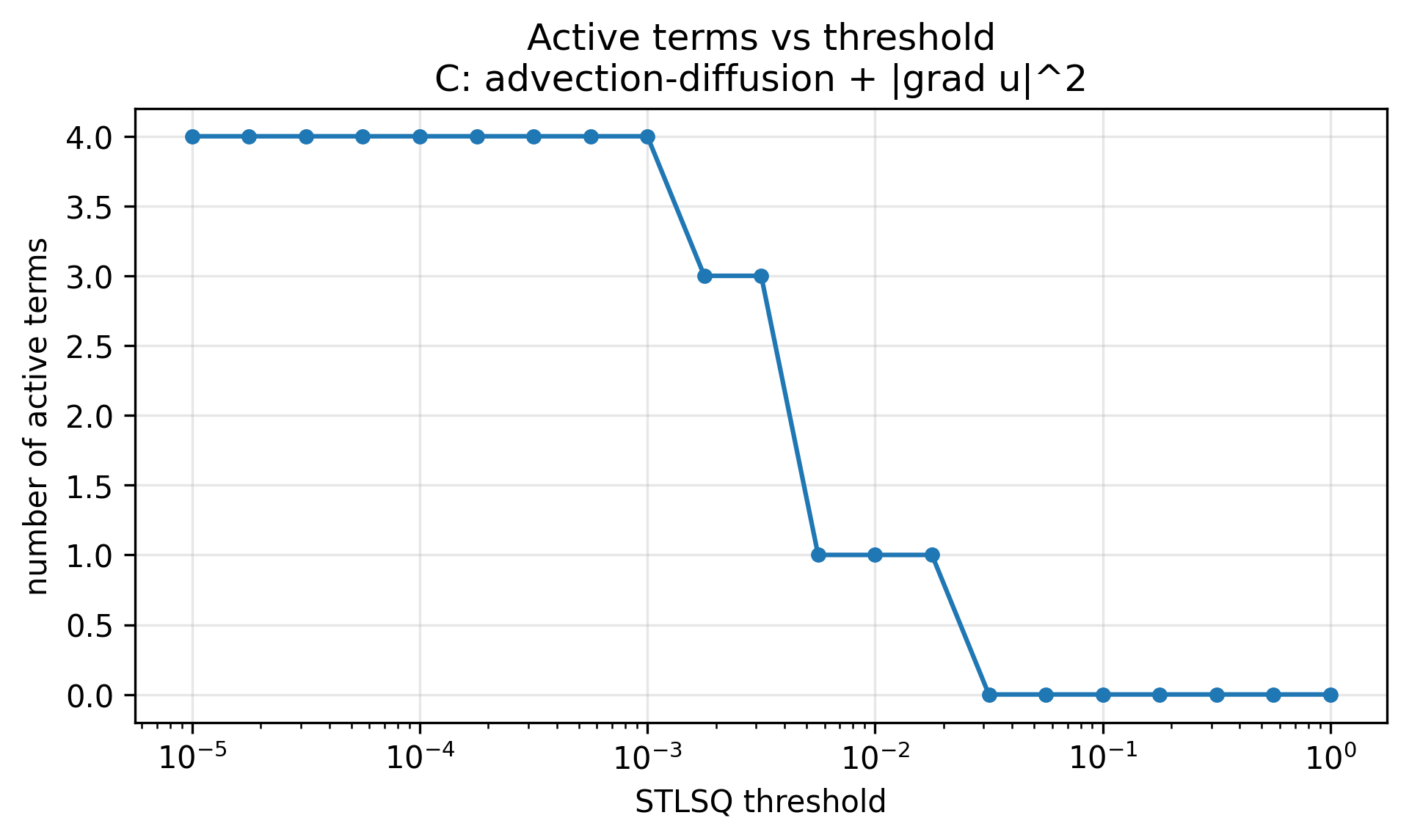}\caption{C}\end{subfigure}

\vspace{0.5em}

\begin{subfigure}{0.32\textwidth}\centering
\includegraphics[width=\linewidth]{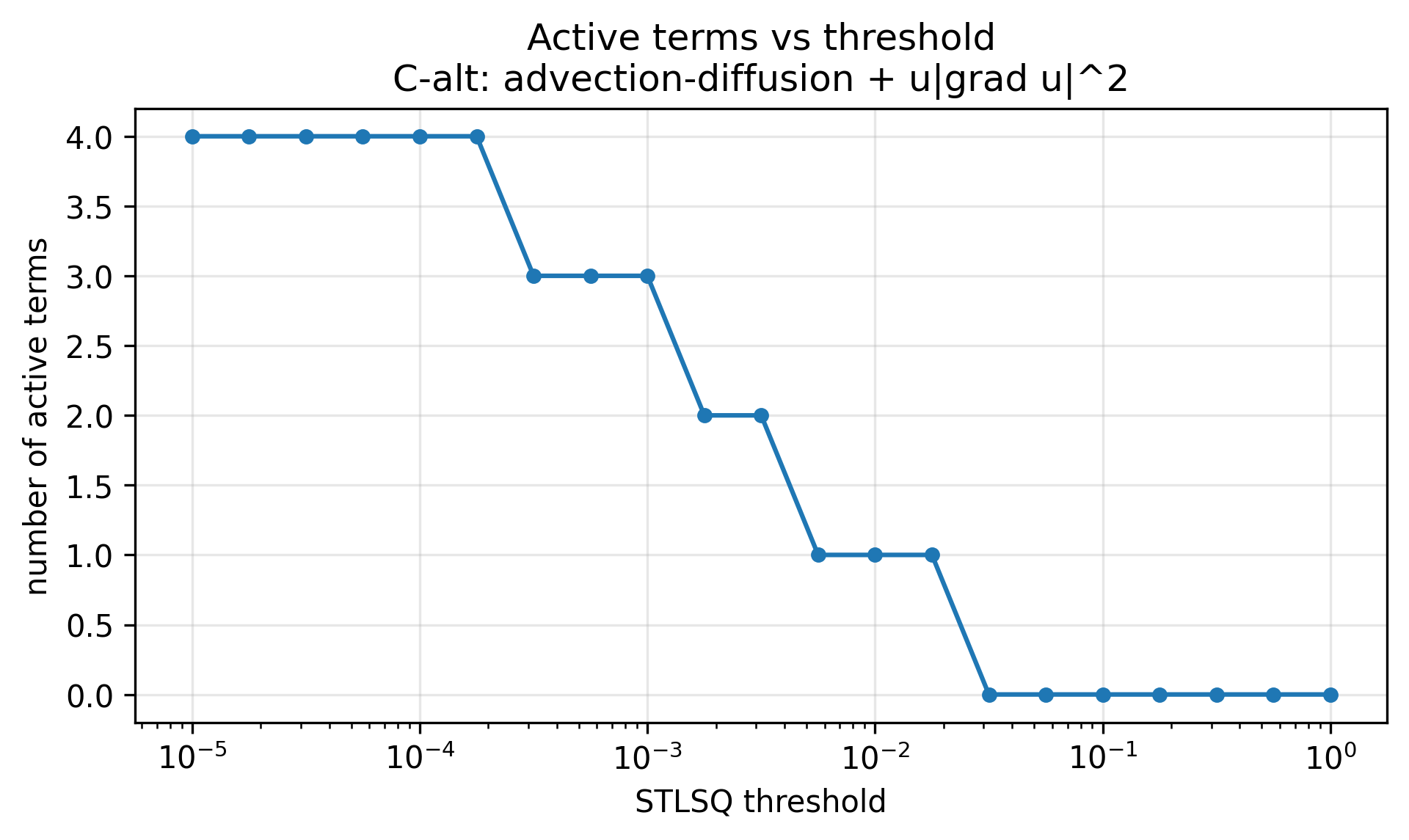}\caption{C-alt}\end{subfigure}\hfill
\begin{subfigure}{0.32\textwidth}\centering
\includegraphics[width=\linewidth]{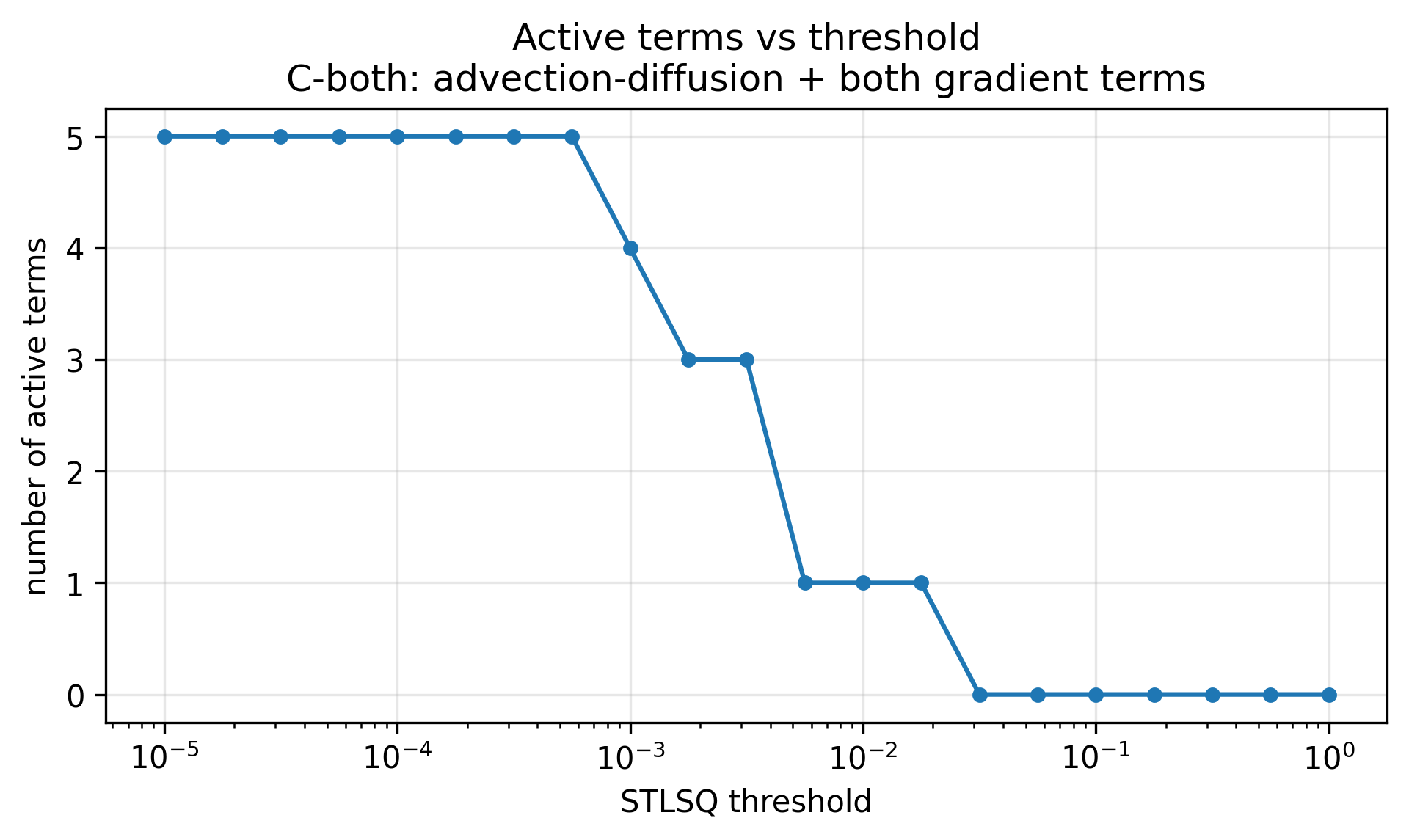}\caption{C-both}\end{subfigure}\hfill
\begin{subfigure}{0.32\textwidth}\centering
\includegraphics[width=\linewidth]{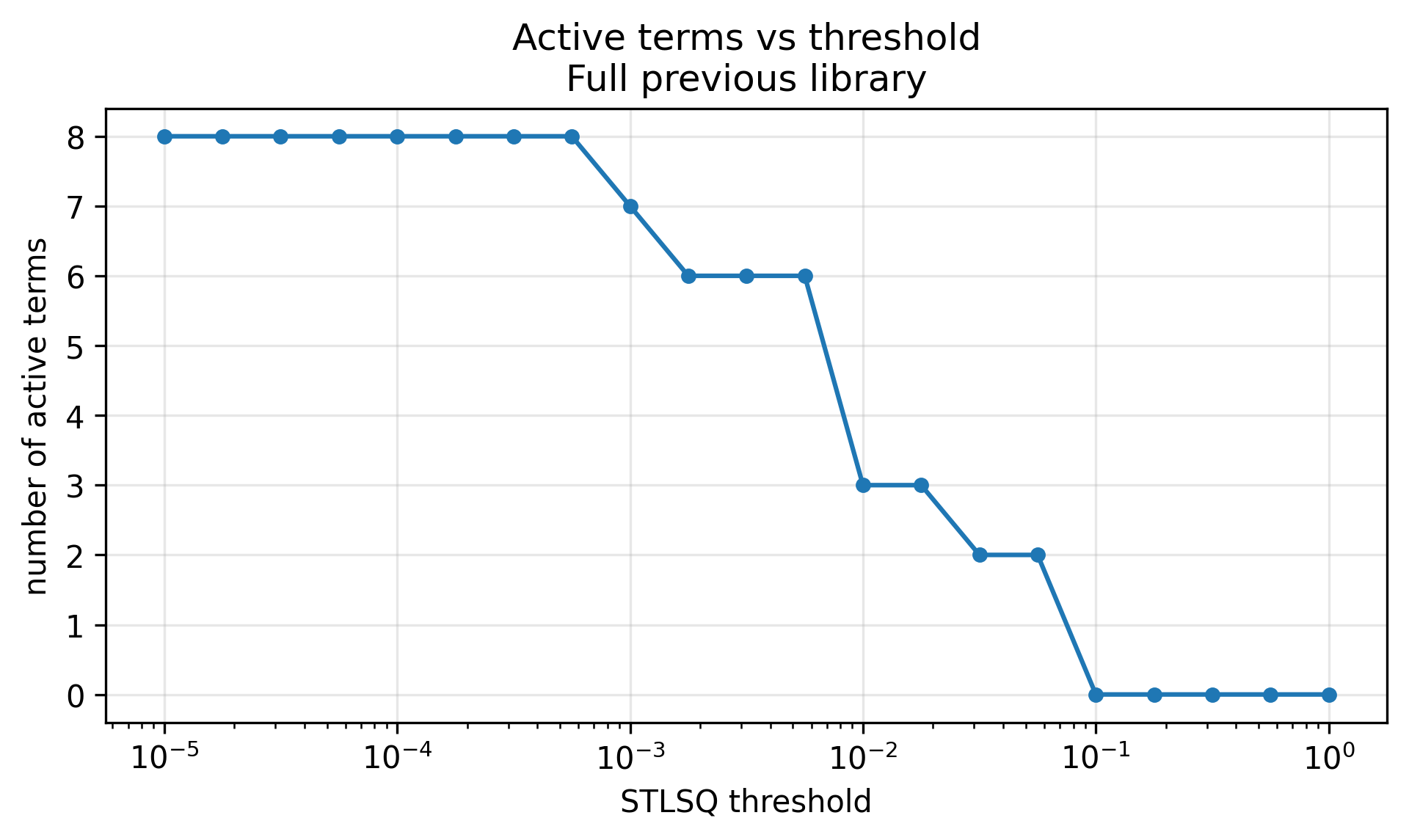}\caption{Full}\end{subfigure}
\caption{Active-term counts under STLSQ as the threshold varies. C and C-alt retain meaningful sparse structures over a broad threshold range; the full library has a more complex support path.}
\label{fig:threshold_active_counts}
\end{figure}

\begin{figure}
\centering
\begin{subfigure}{0.32\textwidth}\centering
\includegraphics[width=\linewidth]{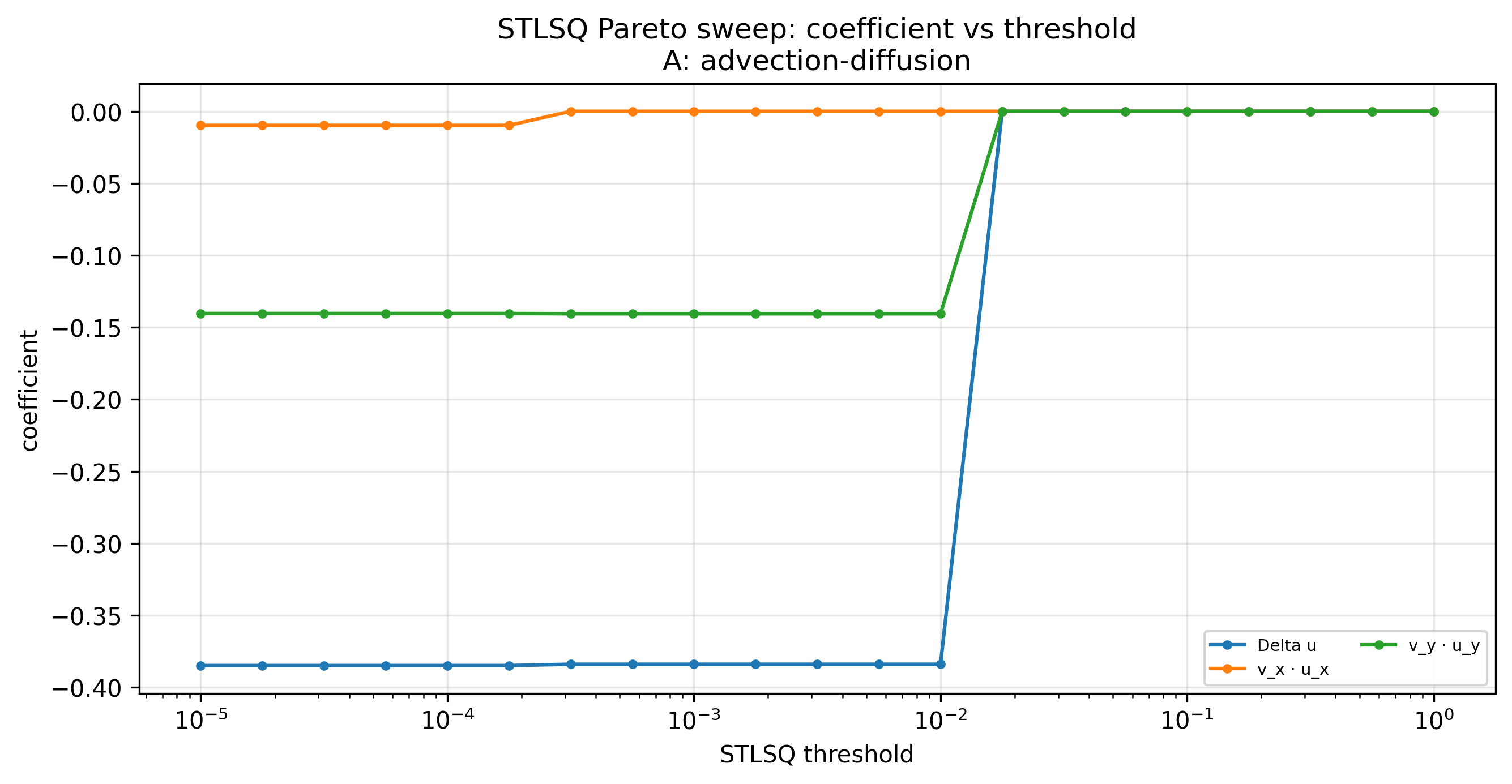}\caption{A}\end{subfigure}\hfill
\begin{subfigure}{0.32\textwidth}\centering
\includegraphics[width=\linewidth]{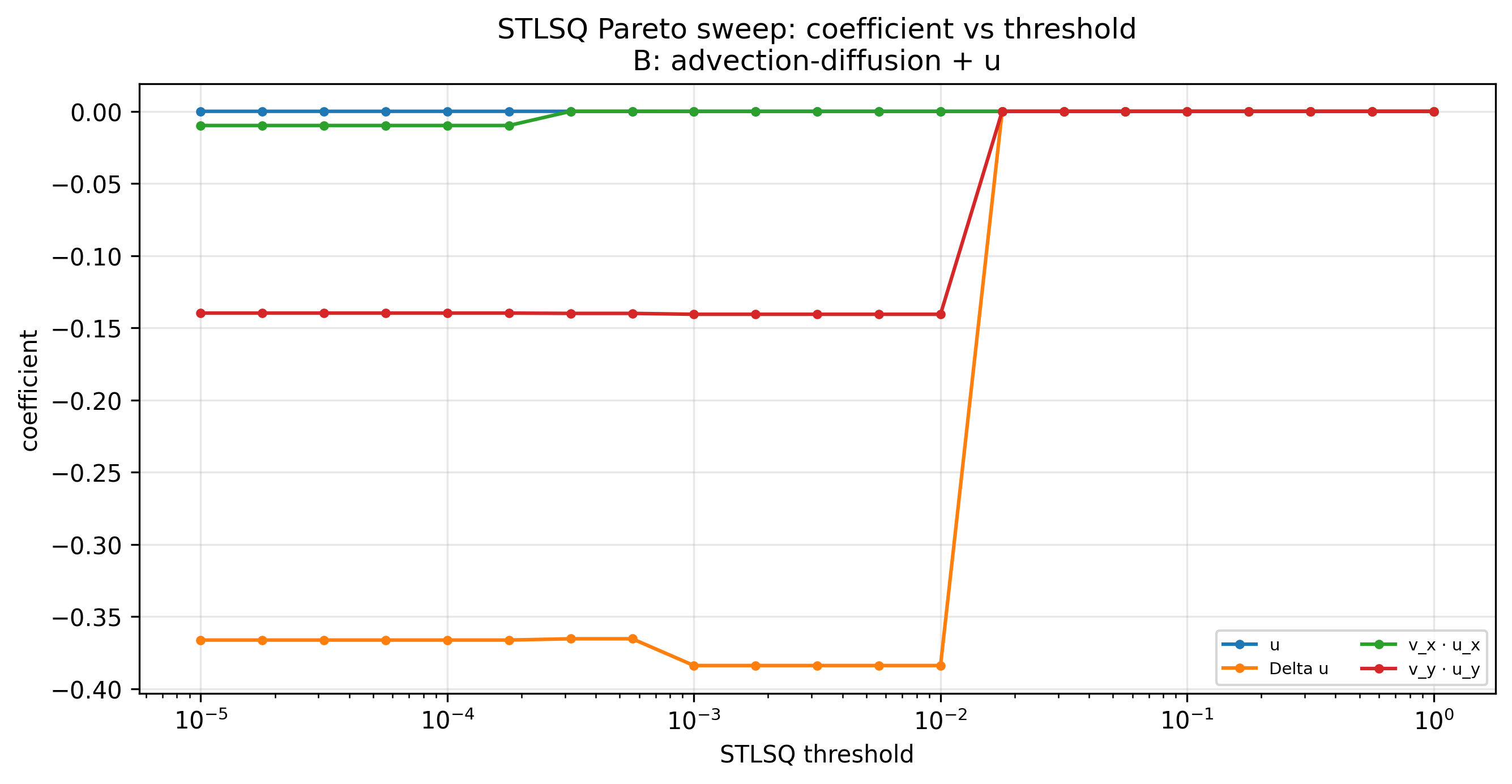}\caption{B}\end{subfigure}\hfill
\begin{subfigure}{0.32\textwidth}\centering
\includegraphics[width=\linewidth]{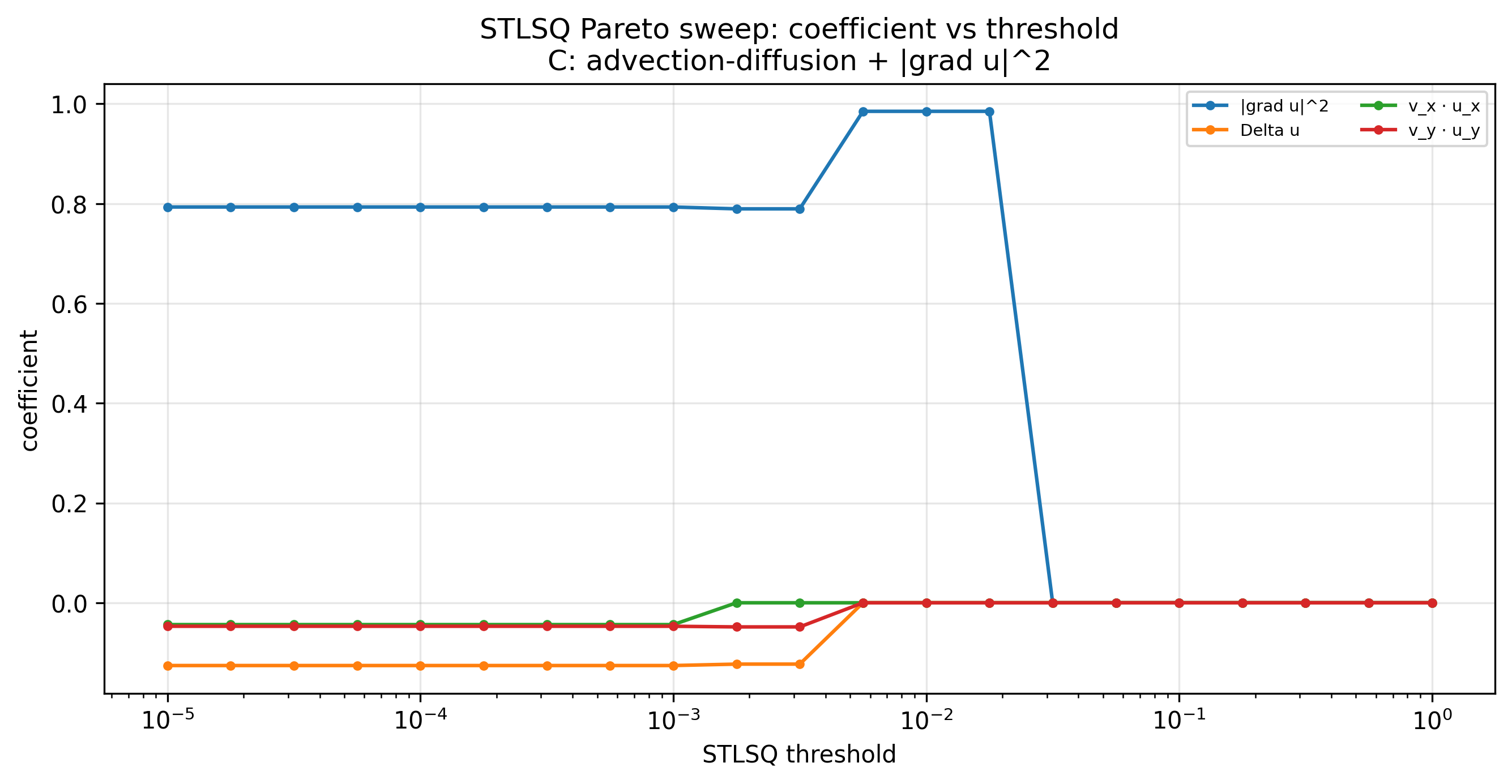}\caption{C}\end{subfigure}

\vspace{0.5em}

\begin{subfigure}{0.32\textwidth}\centering
\includegraphics[width=\linewidth]{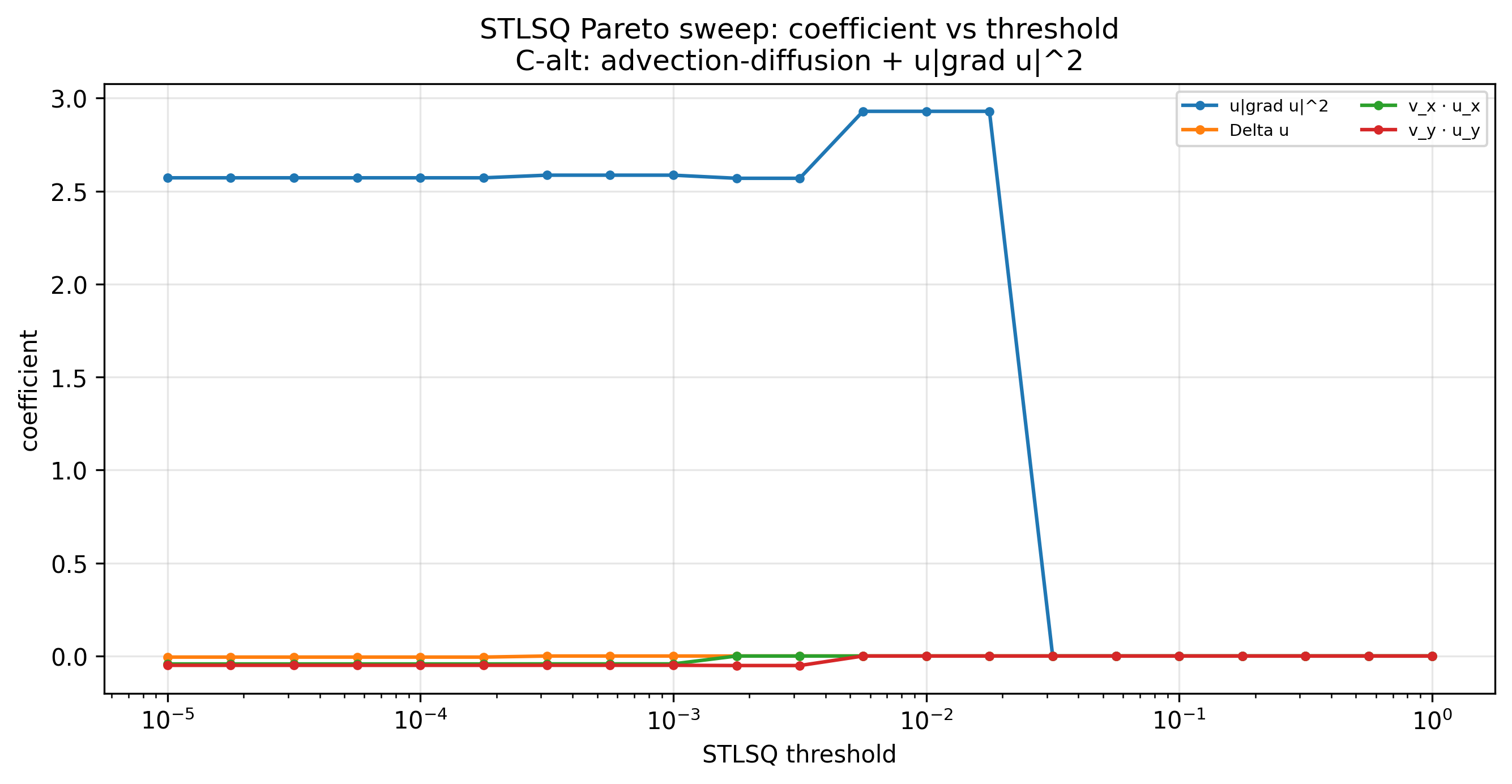}\caption{C-alt}\end{subfigure}\hfill
\begin{subfigure}{0.32\textwidth}\centering
\includegraphics[width=\linewidth]{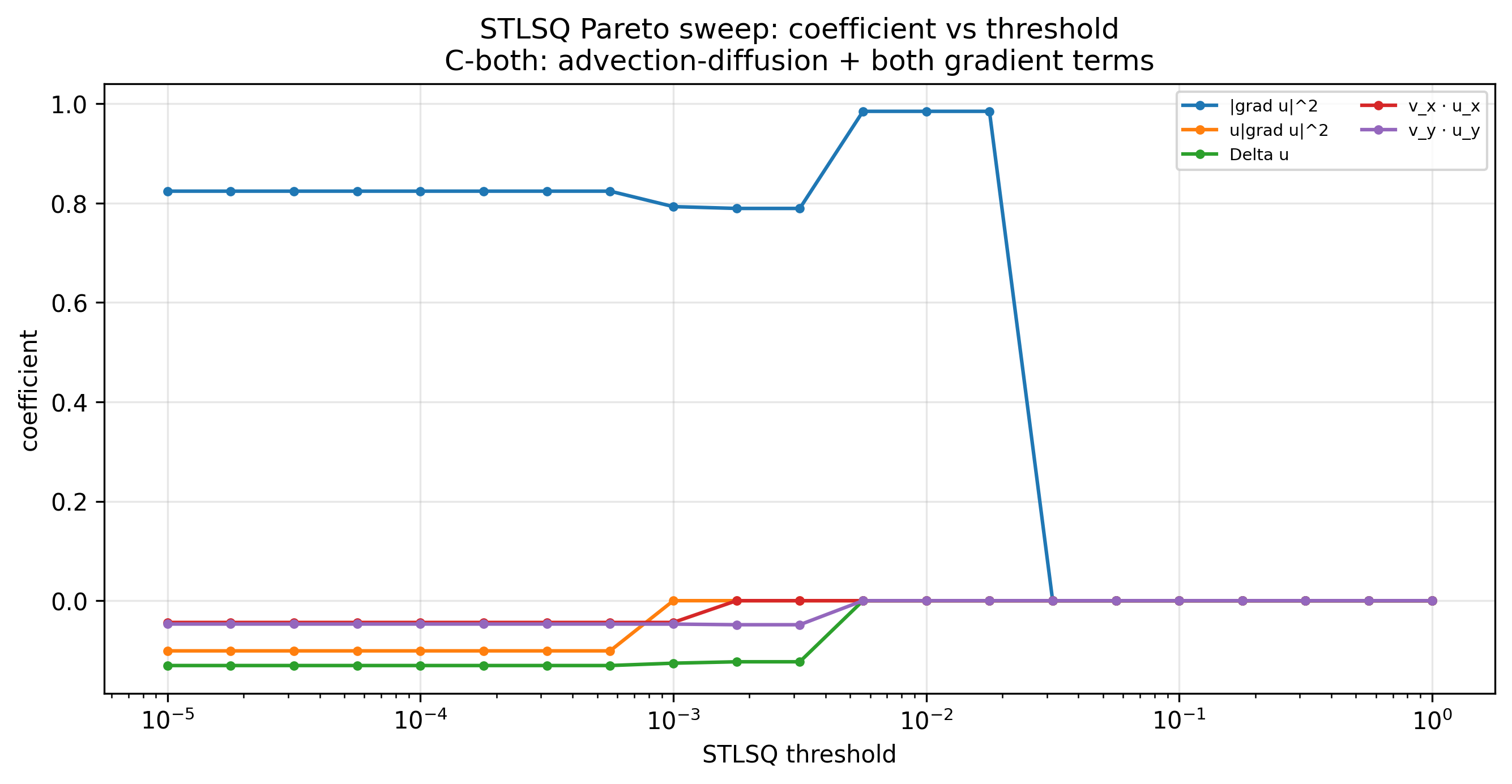}\caption{C-both}\end{subfigure}\hfill
\begin{subfigure}{0.32\textwidth}\centering
\includegraphics[width=\linewidth]{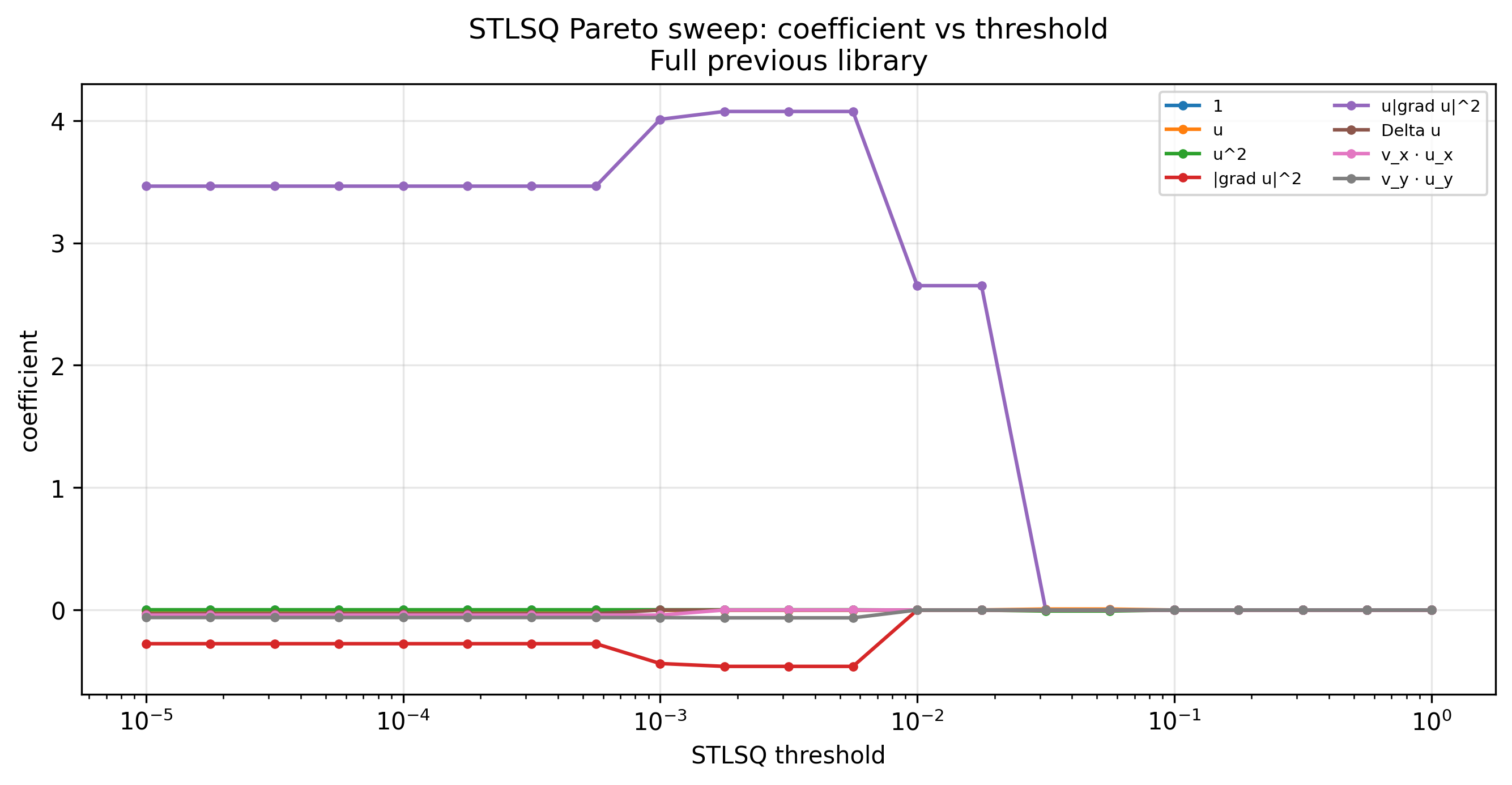}\caption{Full}\end{subfigure}
\caption{Coefficient paths under the STLSQ threshold sweep. Reduced libraries show a single dominant nonlinear-gradient coefficient; the full library shows competing correlated features.}
\label{fig:threshold_coefficients}
\end{figure}

The threshold sweep (Fig.~\ref{fig:threshold_active_counts}) confirms the conditioning picture. A and B collapse rapidly to simple advection--diffusion structures; C and C-alt retain their nonlinear gradient-dependent terms over a broad threshold range; C-both and the full library exhibit complex, less interpretable support paths. The coefficient paths in Fig.~\ref{fig:threshold_coefficients} corroborate this: \(|\nabla u|^{2}\) is the dominant positive term throughout the C path, \(u|\nabla u|^{2}\) plays the analogous role in C-alt, and the full library shows competing intensity and gradient terms with larger variability.

The conditioning, correlation, and threshold-sweep evidence already disqualifies the full and C-both libraries on identifiability grounds; libraries A and B are stable but too restrictive to capture the observed nonlinear deformation. The remaining contenders are C and C-alt.

\subsection{Random-Centre Stability}\label{subsec:stability_results}

\begin{figure}
\centering
\begin{subfigure}{0.32\textwidth}\centering
\includegraphics[width=\linewidth]{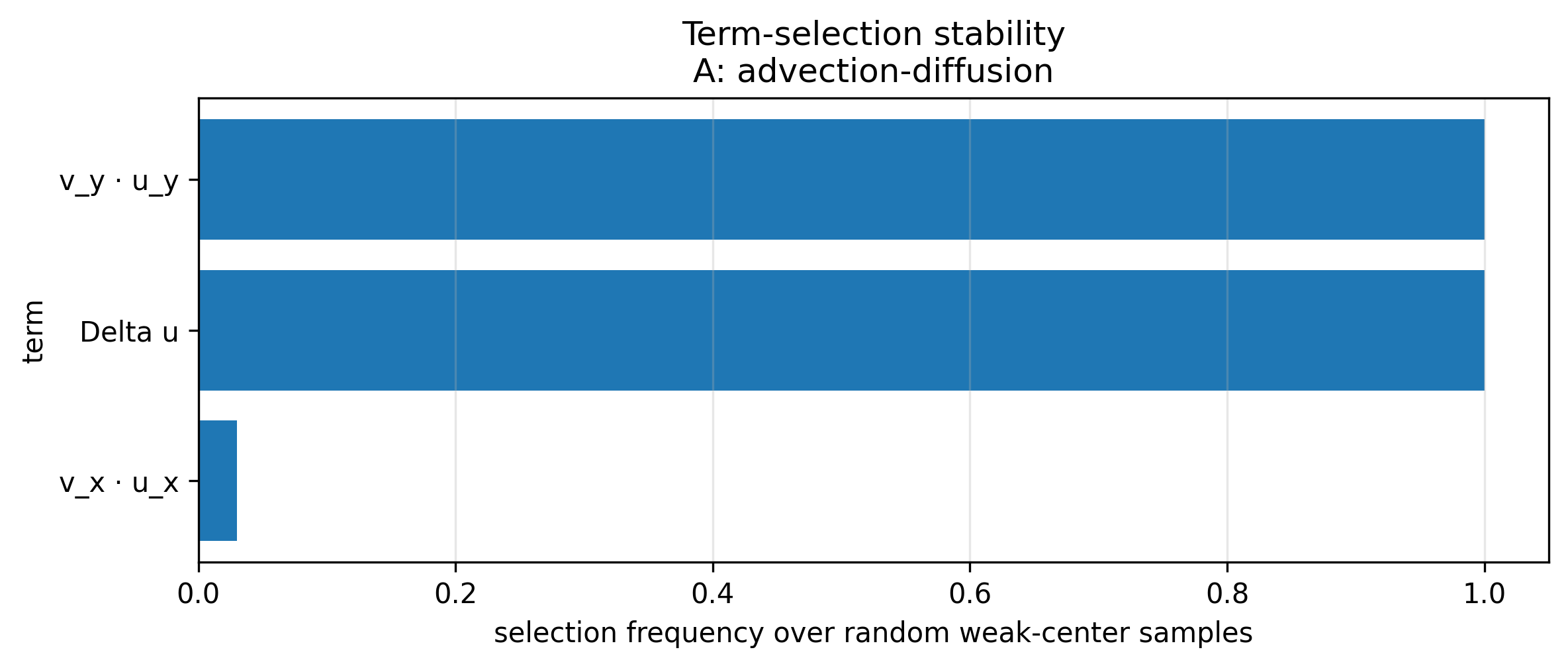}\caption{A}\end{subfigure}\hfill
\begin{subfigure}{0.32\textwidth}\centering
\includegraphics[width=\linewidth]{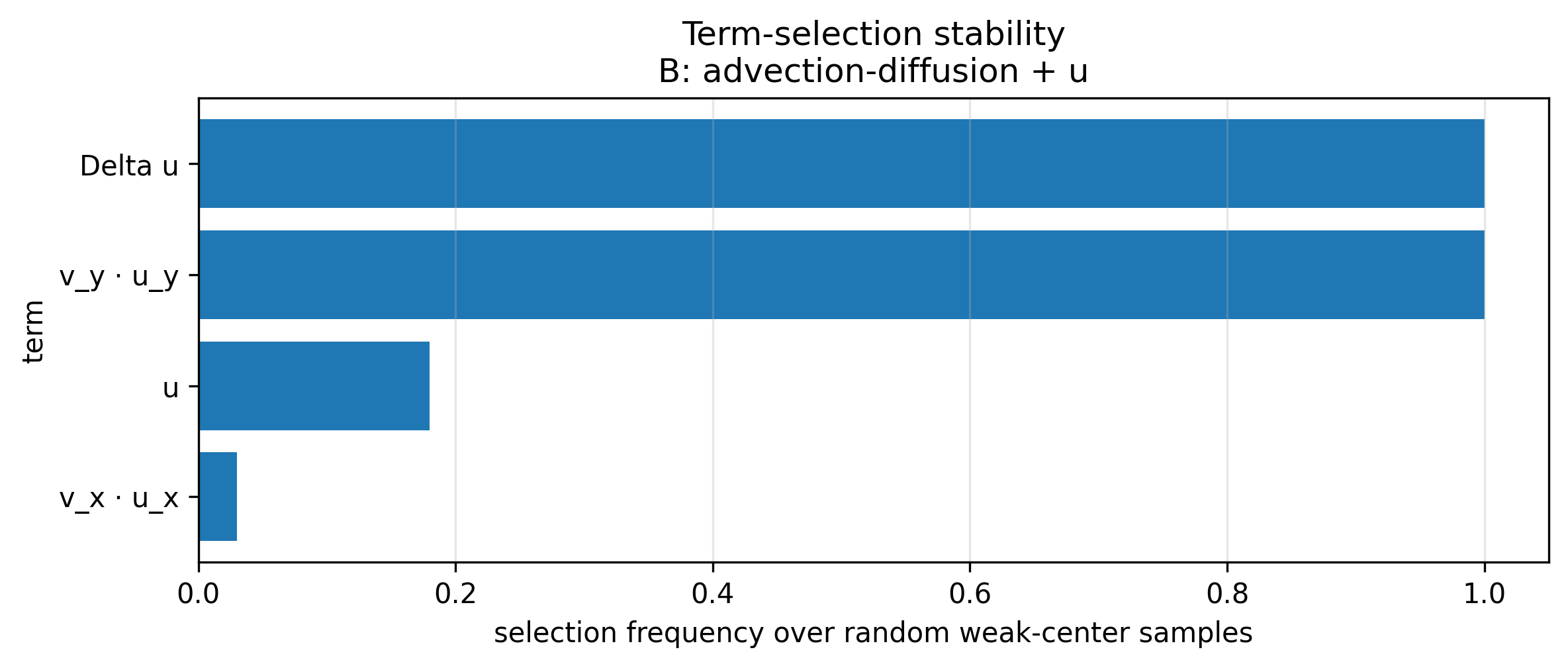}\caption{B}\end{subfigure}\hfill
\begin{subfigure}{0.32\textwidth}\centering
\includegraphics[width=\linewidth]{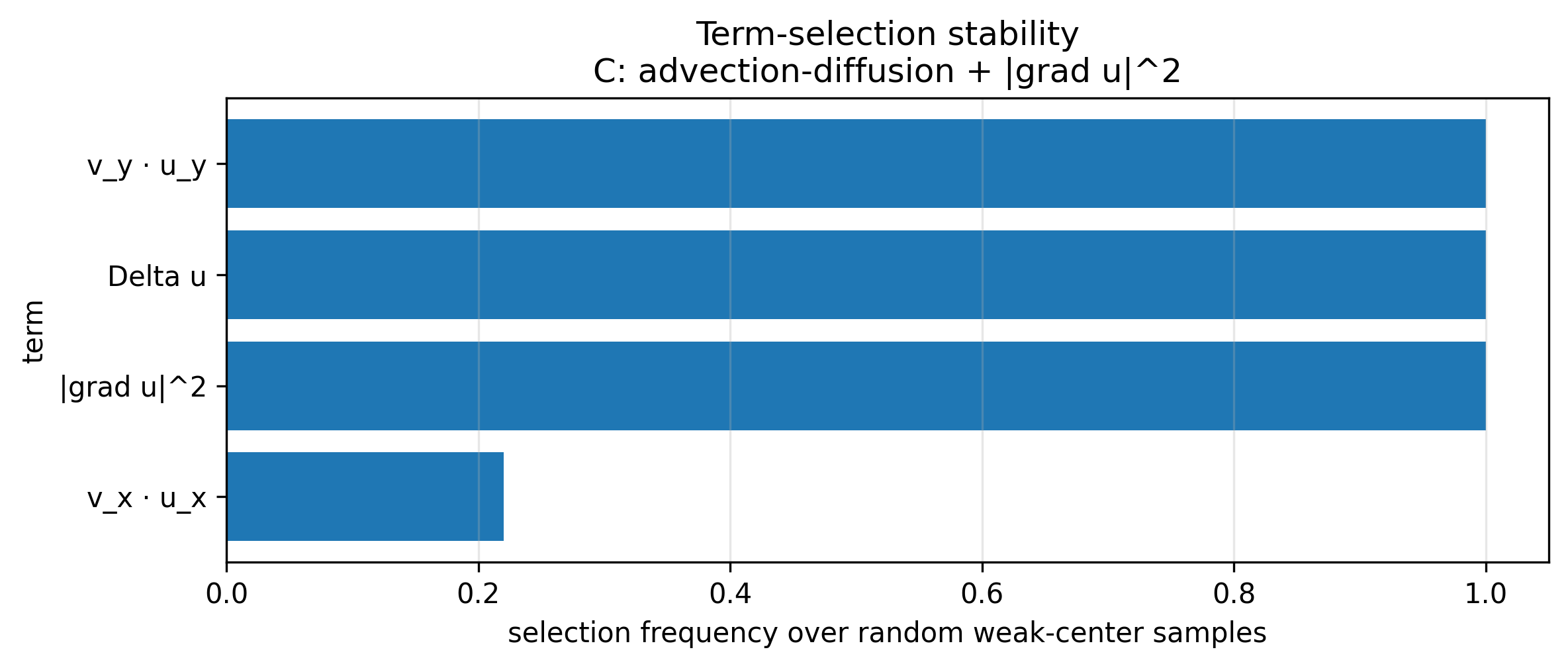}\caption{C}\end{subfigure}

\vspace{0.5em}

\begin{subfigure}{0.32\textwidth}\centering
\includegraphics[width=\linewidth]{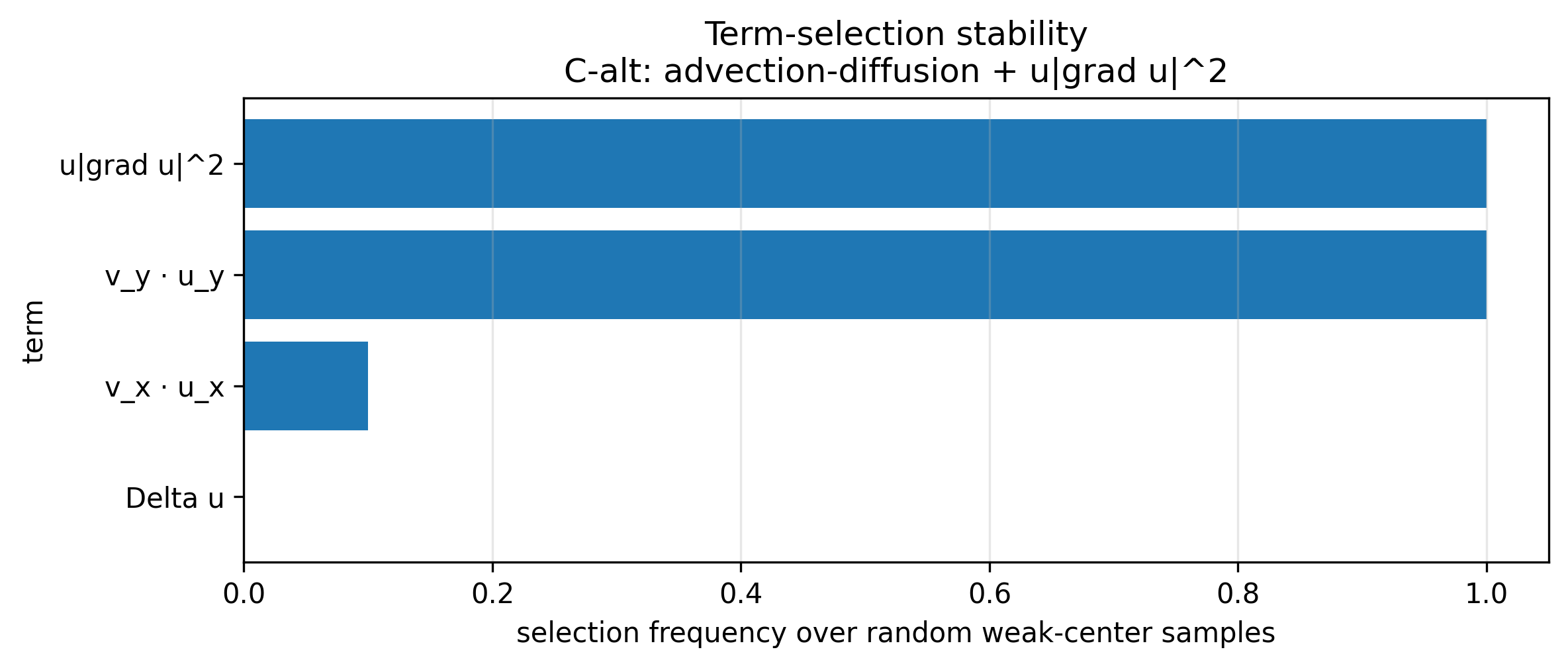}\caption{C-alt}\end{subfigure}\hfill
\begin{subfigure}{0.32\textwidth}\centering
\includegraphics[width=\linewidth]{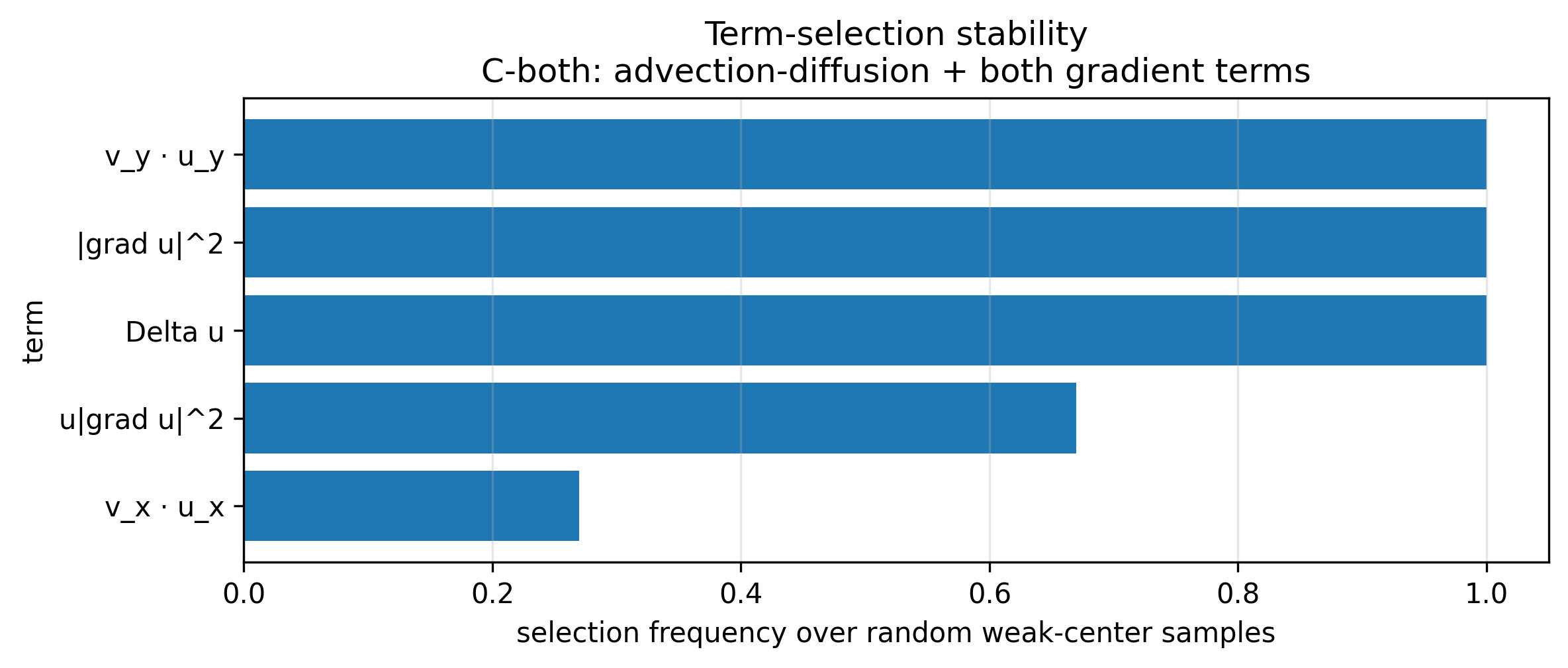}\caption{C-both}\end{subfigure}\hfill
\begin{subfigure}{0.32\textwidth}\centering
\includegraphics[width=\linewidth]{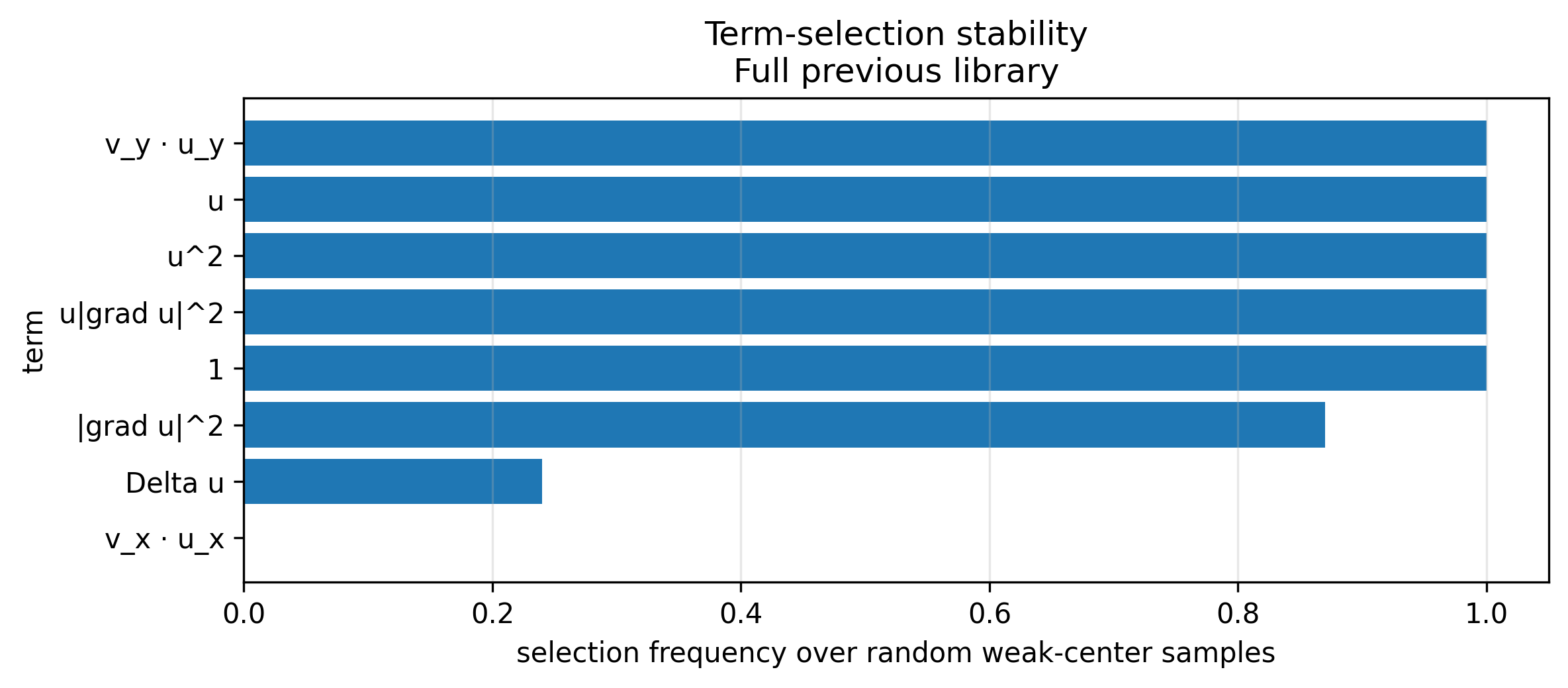}\caption{Full}\end{subfigure}
\caption{Selection frequencies over \(100\) random weak-centre samples.}
\label{fig:stability_selection_frequency}
\end{figure}

Selection frequencies over \(100\) random centre samples (Fig.~\ref{fig:stability_selection_frequency}) show that the dominant nonlinear gradient term and the \(y\)-advection term are selected in every run for both C and C-alt. In B, the additional linear term \(u\) is selected in only \(18\%\) of runs and is therefore not robust. The full library and C-both select many terms with high frequency, but this reflects an overcomplete feature set rather than stronger evidence for any individual term.

\begin{figure}
\centering
\begin{subfigure}{0.32\textwidth}\centering
\includegraphics[width=\linewidth]{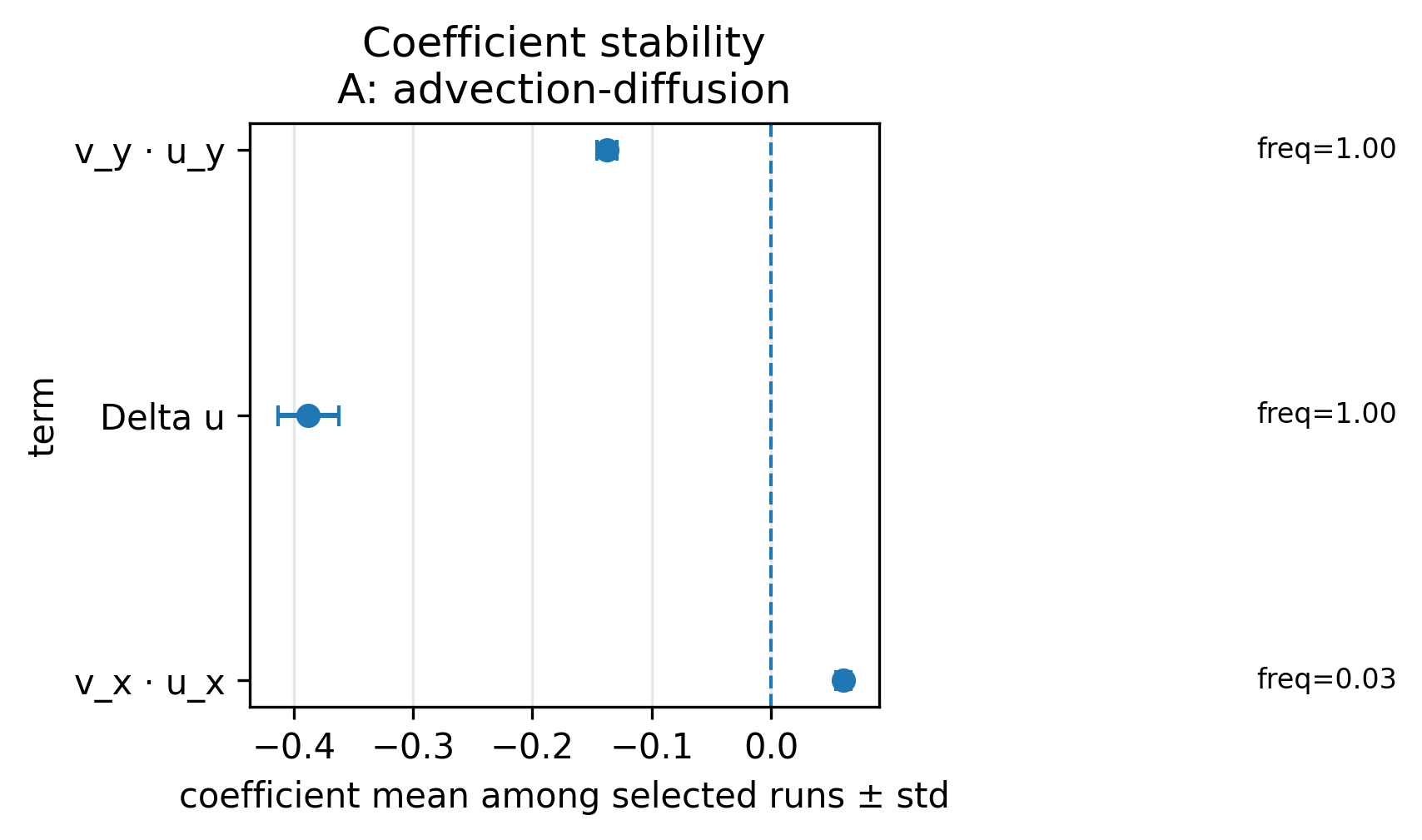}\caption{A}\end{subfigure}\hfill
\begin{subfigure}{0.32\textwidth}\centering
\includegraphics[width=\linewidth]{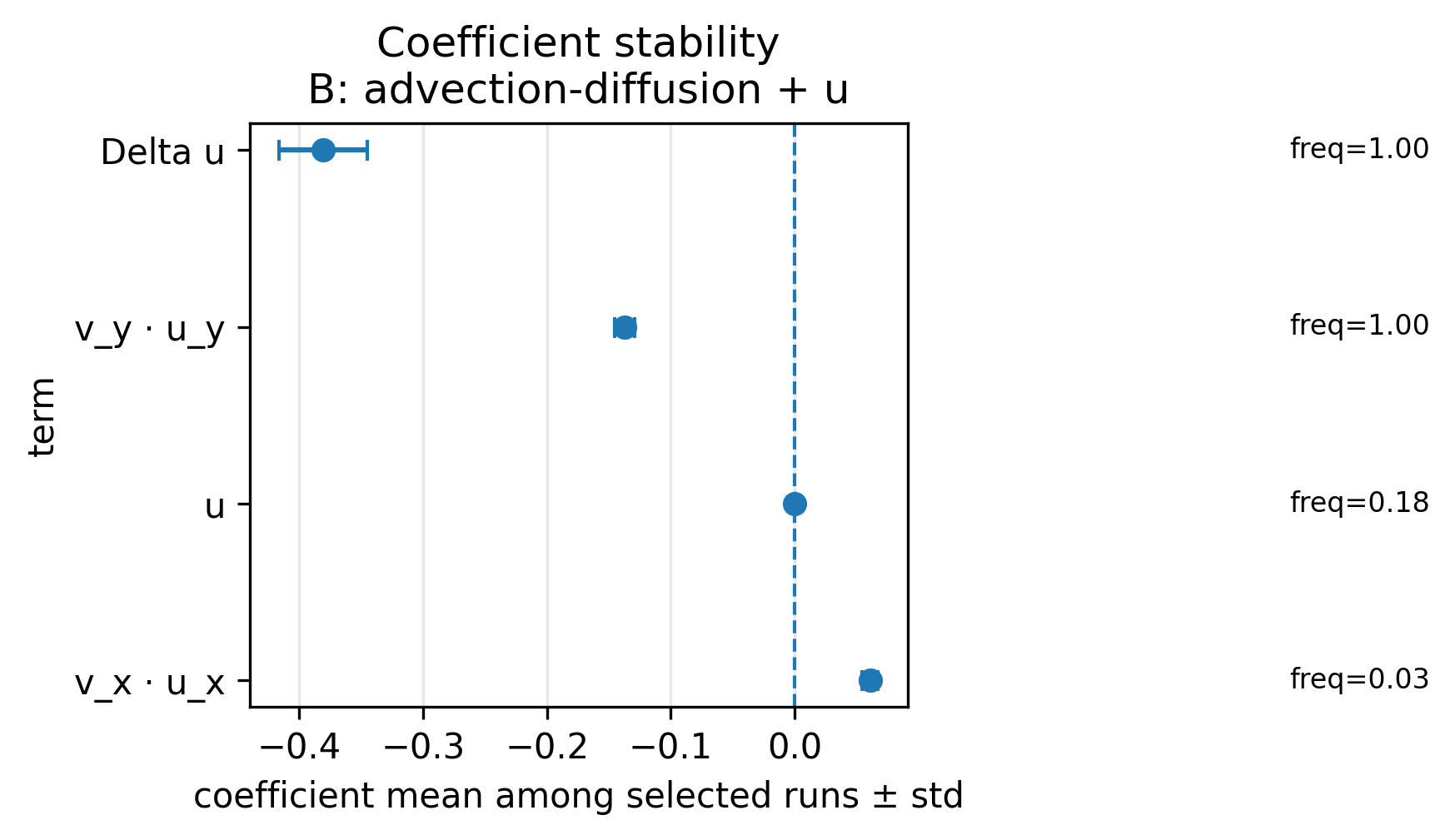}\caption{B}\end{subfigure}\hfill
\begin{subfigure}{0.32\textwidth}\centering
\includegraphics[width=\linewidth]{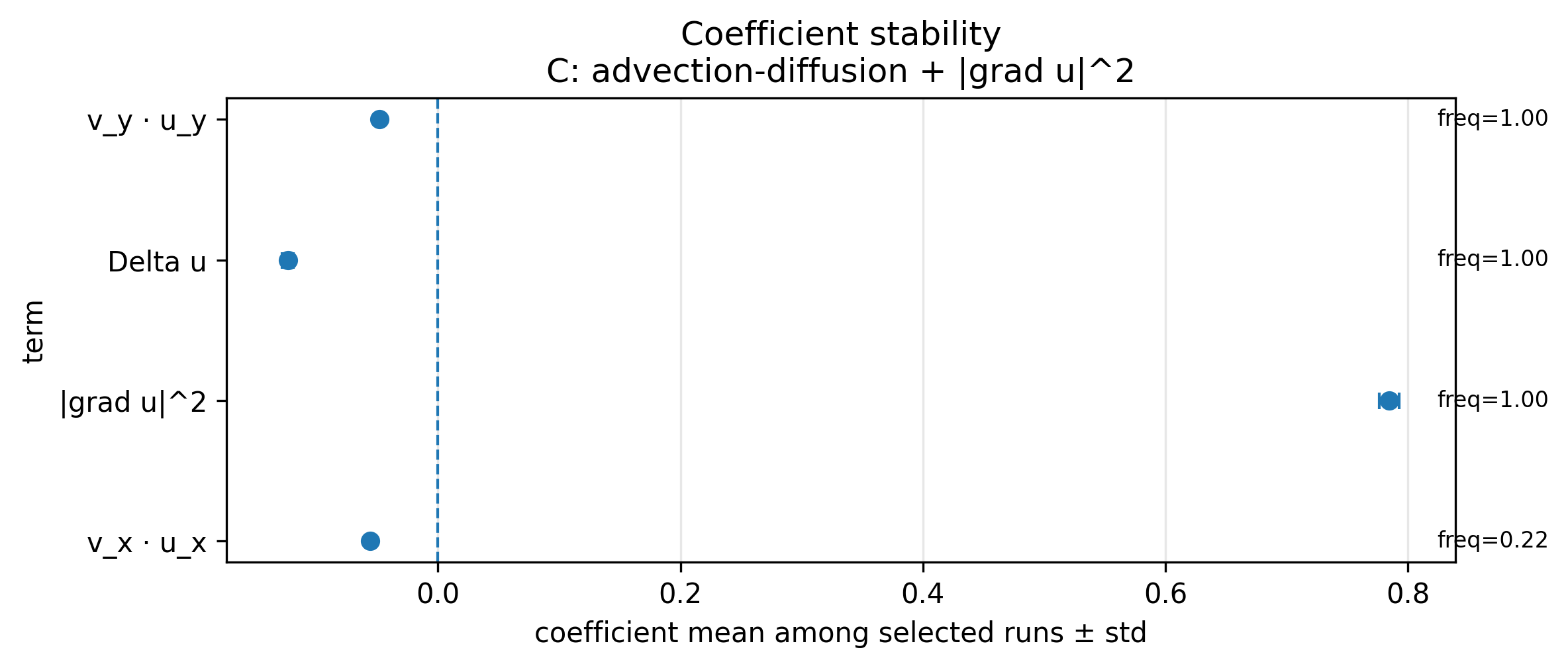}\caption{C}\end{subfigure}

\vspace{0.5em}

\begin{subfigure}{0.32\textwidth}\centering
\includegraphics[width=\linewidth]{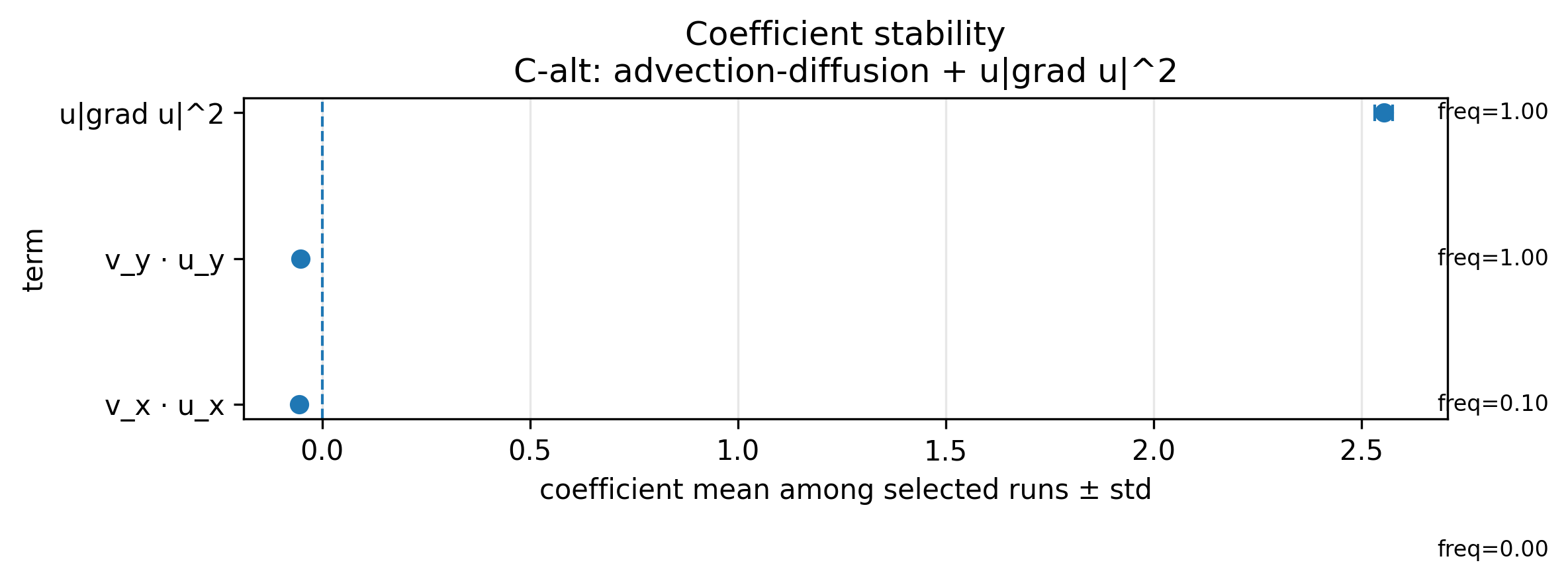}\caption{C-alt}\end{subfigure}\hfill
\begin{subfigure}{0.32\textwidth}\centering
\includegraphics[width=\linewidth]{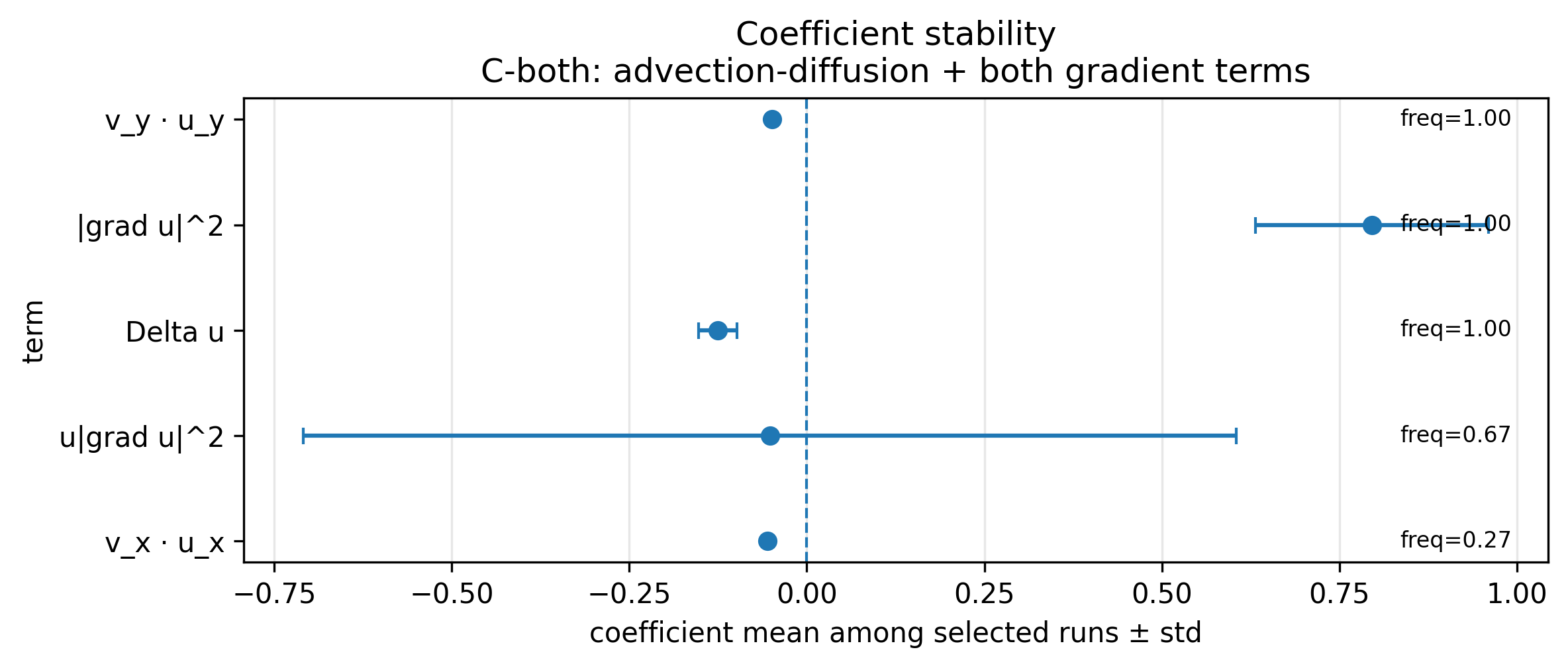}\caption{C-both}\end{subfigure}\hfill
\begin{subfigure}{0.32\textwidth}\centering
\includegraphics[width=\linewidth]{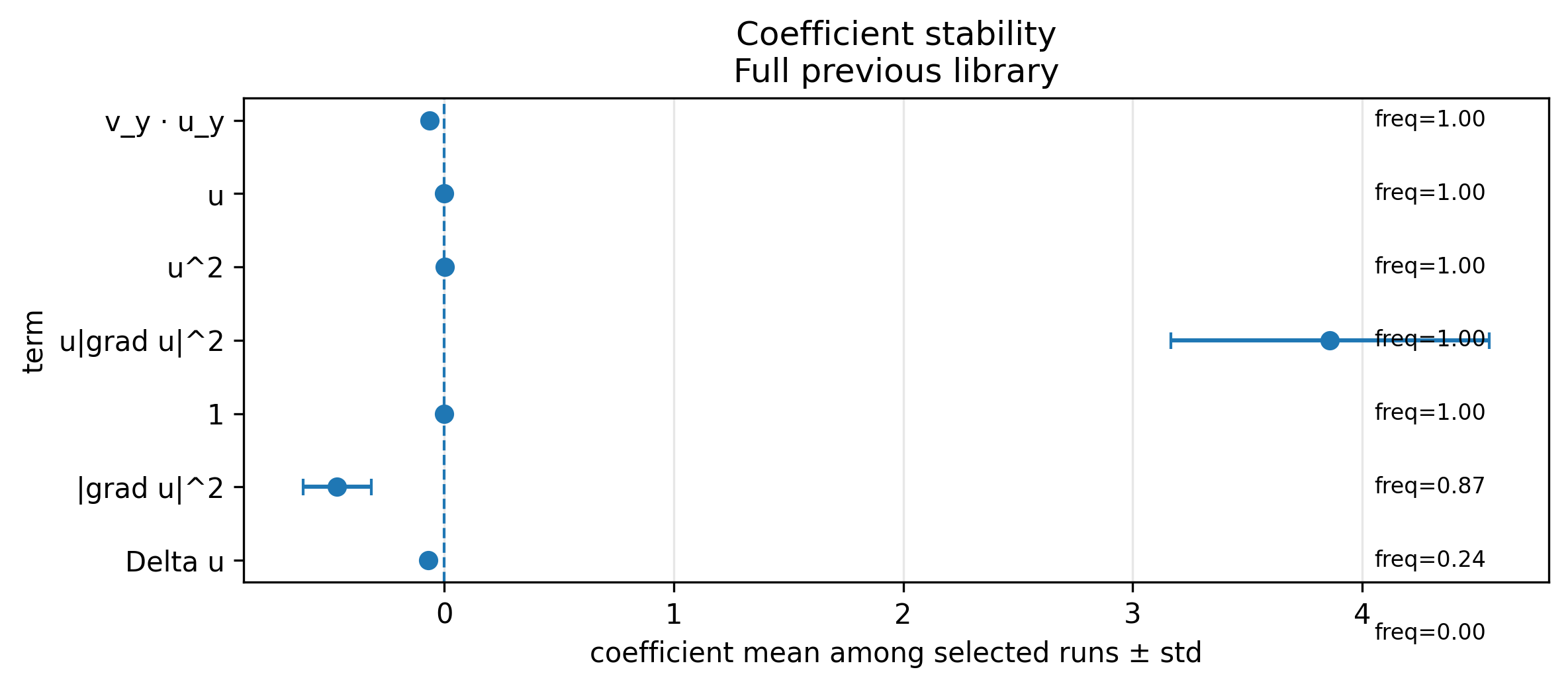}\caption{Full}\end{subfigure}
\caption{Coefficient stability over \(100\) random weak-centre samples; markers show the mean among selected runs and bars show one standard deviation.}
\label{fig:stability_coefficients}
\end{figure}

The coefficient distributions (Fig.~\ref{fig:stability_coefficients} and Table~\ref{tab:stability_summary_reduced}) sharpen the comparison between C and C-alt. In C, \(|\nabla u|^{2}\), \(\Delta u\), and \(v_yu_y\) are selected with frequency \(1.0\) and have small variability; in C-alt, \(u|\nabla u|^{2}\) and \(v_yu_y\) are similarly stable but the Laplacian is never selected, suggesting that \(u|\nabla u|^{2}\) absorbs the spreading. C is therefore preferred on identifiability grounds, since it retains a stable Laplacian alongside the nonlinear gradient term; C-alt is kept as a competing reduced model whose weighted gradient term may furnish an alternative effective description of the front.

\begin{table}
\centering
\caption{Random-centre stability for the reduced libraries.}
\label{tab:stability_summary_reduced}
\small
\begin{tabular*}{\linewidth}{@{\extracolsep{\fill}}llccc@{}}
\toprule
\textbf{Library} & \textbf{Term} & \textbf{Selection freq.} & \textbf{Mean coeff.} & \textbf{Std.} \\
\midrule
C & \(|\nabla u|^{2}\) & \(1.00\) & \(0.7845\) & \(0.0083\) \\
C & \(\Delta u\) & \(1.00\) & \(-0.1238\) & \(0.0045\) \\
C & \(v_yu_y\) & \(1.00\) & \(-0.0485\) & \(0.0015\) \\
C & \(v_xu_x\) & \(0.22\) & \(-0.0560\) & \(0.0026\) \\
\midrule
C-alt & \(u|\nabla u|^{2}\) & \(1.00\) & \(2.5544\) & \(0.0216\) \\
C-alt & \(v_yu_y\) & \(1.00\) & \(-0.0514\) & \(0.0015\) \\
C-alt & \(v_xu_x\) & \(0.10\) & \(-0.0550\) & \(0.0014\) \\
C-alt & \(\Delta u\) & \(0.00\) & -- & -- \\
\midrule
C-both & \(|\nabla u|^{2}\) & \(1.00\) & \(0.7958\) & \(0.1645\) \\
C-both & \(\Delta u\) & \(1.00\) & \(-0.1256\) & \(0.0271\) \\
C-both & \(u|\nabla u|^{2}\) & \(0.67\) & \(-0.0520\) & \(0.6569\) \\
C-both & \(v_yu_y\) & \(1.00\) & \(-0.0484\) & \(0.0016\) \\
\bottomrule
\end{tabular*}
\end{table}

\subsection{Validation Rollouts: Measured versus Learned Advection}\label{subsec:validation_a_b}

\begin{figure}
    \centering
    \includegraphics[width=\textwidth]{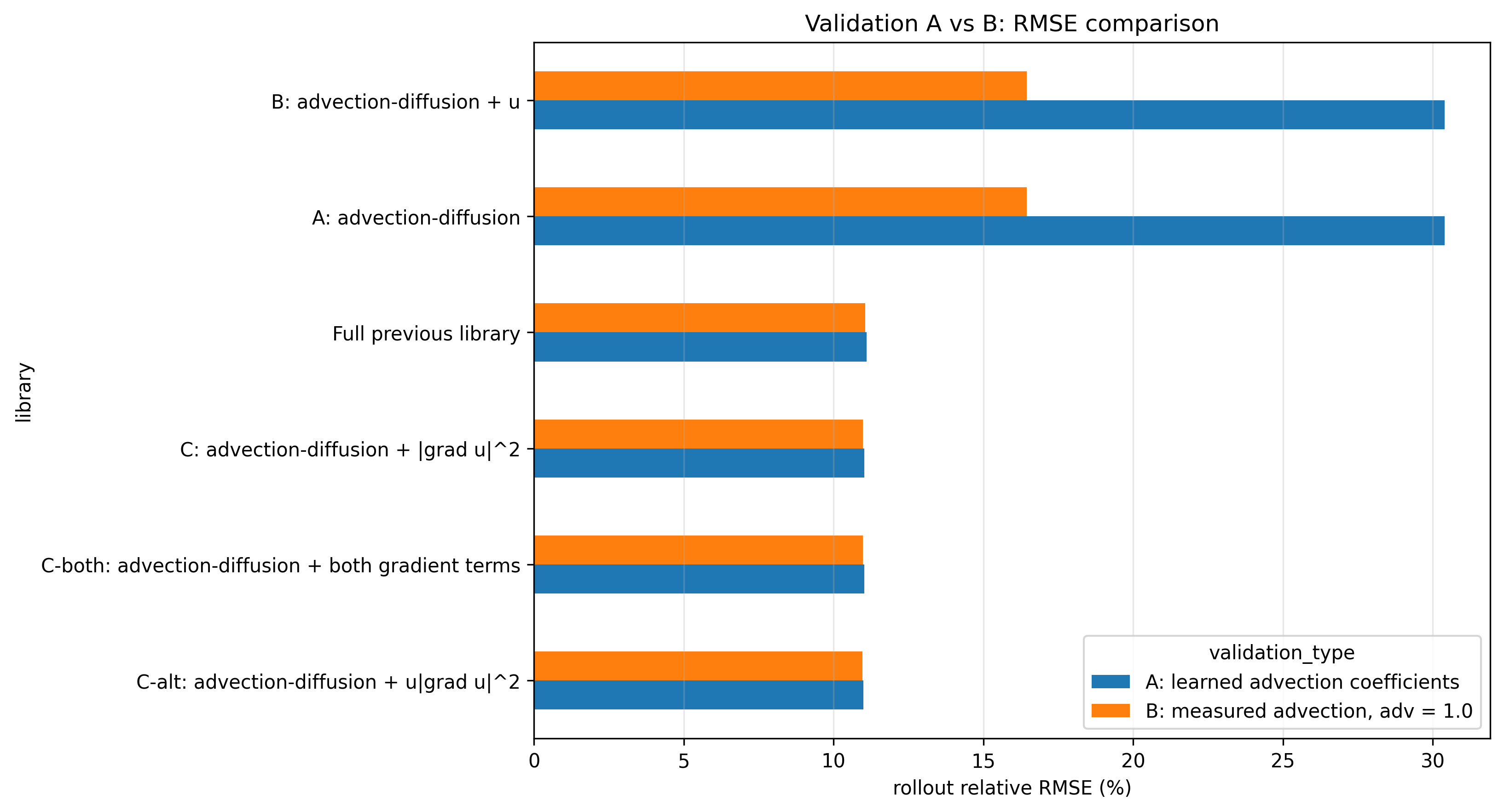}
    \caption{Validation rollout rRMSE under Mode A (learned advection) and Mode B (measured advection, coefficient \(1.0\)). Mode B substantially improves the advection--diffusion baselines and preserves centroid trajectories for the nonlinear-gradient models.}
    \label{fig:validation_A_vs_B_rmse}
\end{figure}

The advection treatment matters more than the library choice for bulk-trajectory accuracy. Figure~\ref{fig:validation_A_vs_B_rmse} and Table~\ref{tab:validation_AB_summary} show that A and B produce rollout rRMSE above \(30\%\) under Mode A and around \(16.45\%\) under Mode B, while the nonlinear-gradient libraries cluster near \(11\%\) under both modes. The pixel-wise RMSE is therefore largely insensitive to whether the advection coefficients are learned or fixed at unity. Centroid accuracy, however, is not: the C and C-alt models have Mode B centroid RMSE below \(0.05\), against values near \(1.1\) under Mode A. The learned-advection mode reproduces image intensity adequately by allowing other terms to compensate for drift mismatch, but it does not preserve the observed bulk plume trajectory. We therefore adopt Mode B for all subsequent calibration and evaluation.

\begin{table}
\centering
\caption{Validation diagnostics under Mode A and Mode B; ordered by Mode B rRMSE.}
\label{tab:validation_AB_summary}
\scriptsize
\resizebox{\linewidth}{!}{%
\begin{tabular}{lccccccccc}
\toprule
\textbf{Library} & \textbf{A RMSE} & \textbf{B RMSE} & \textbf{A Front MAE} & \textbf{A Front RMSE} & \textbf{B Front MAE} & \textbf{B Front RMSE} & \textbf{A COM RMSE} & \textbf{B COM RMSE} & \textbf{Pref.} \\
\midrule
C-alt   & \(11.00\) & \(10.96\) & \(3.1050\) & \(3.5656\) & \(3.0568\) & \(3.5106\) & \(1.1128\) & \(0.0484\) & B \\
C       & \(11.02\) & \(10.98\) & \(3.0448\) & \(3.4872\) & \(3.0242\) & \(3.4713\) & \(1.1164\) & \(0.0444\) & B \\
C-both  & \(11.02\) & \(10.98\) & \(3.0448\) & \(3.4872\) & \(3.0242\) & \(3.4713\) & \(1.1164\) & \(0.0444\) & B \\
Full    & \(11.10\) & \(11.05\) & \(3.1606\) & \(3.6294\) & \(3.1073\) & \(3.5678\) & \(1.0964\) & \(0.0534\) & B \\
A       & \(30.40\) & \(16.45\) & \(6.2990\) & \(9.1993\) & \(3.7346\) & \(4.5657\) & \(0.8683\) & \(0.1570\) & B \\
B       & \(30.40\) & \(16.45\) & \(6.2990\) & \(9.1993\) & \(3.7346\) & \(4.5657\) & \(0.8683\) & \(0.1570\) & B \\
\bottomrule
\end{tabular}}
\end{table}

The two leading reduced models at this stage are
\begin{equation}
\text{C:}\quad u_t + \mathbf{v}(t)\!\cdot\!\nabla u = 0.7931\,|\nabla u|^{2} - 0.1258\,\Delta u,
\quad\text{C-alt:}\quad u_t + \mathbf{v}(t)\!\cdot\!\nabla u = 2.5860\,u|\nabla u|^{2}.
\label{eq:validation_winners}
\end{equation}
C-both reduces to the same active simulator structure as C and is dropped. The C and C-alt model classes are carried forward into iPINN refinement and bootstrap calibration.

\subsection{iPINN-Refined Coefficients}\label{subsec:ipinn_results}

\begin{figure}
\centering
\begin{subfigure}{0.49\textwidth}\centering
\includegraphics[width=\linewidth]{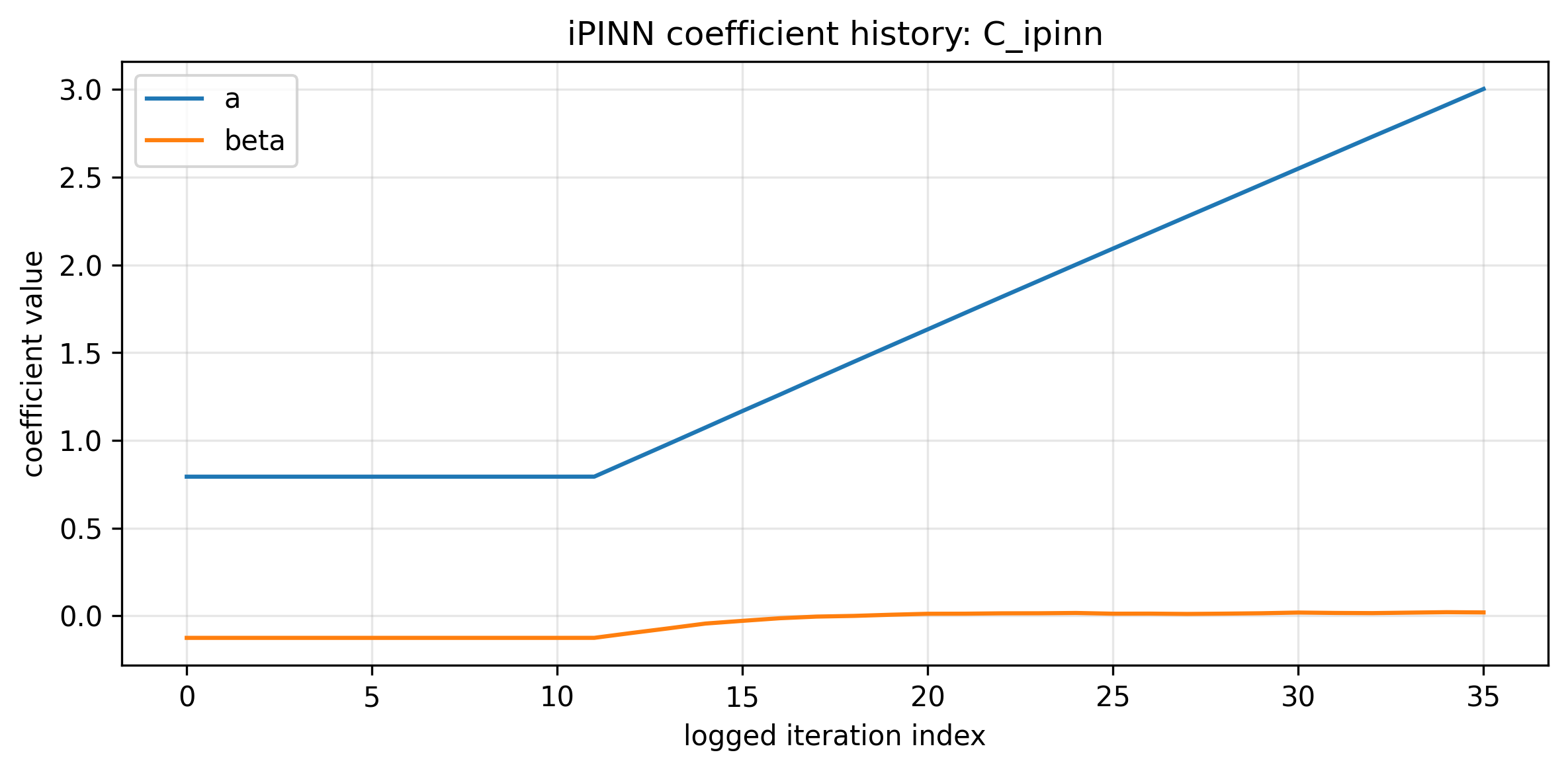}\caption{C coefficient history}\end{subfigure}\hfill
\begin{subfigure}{0.49\textwidth}\centering
\includegraphics[width=\linewidth]{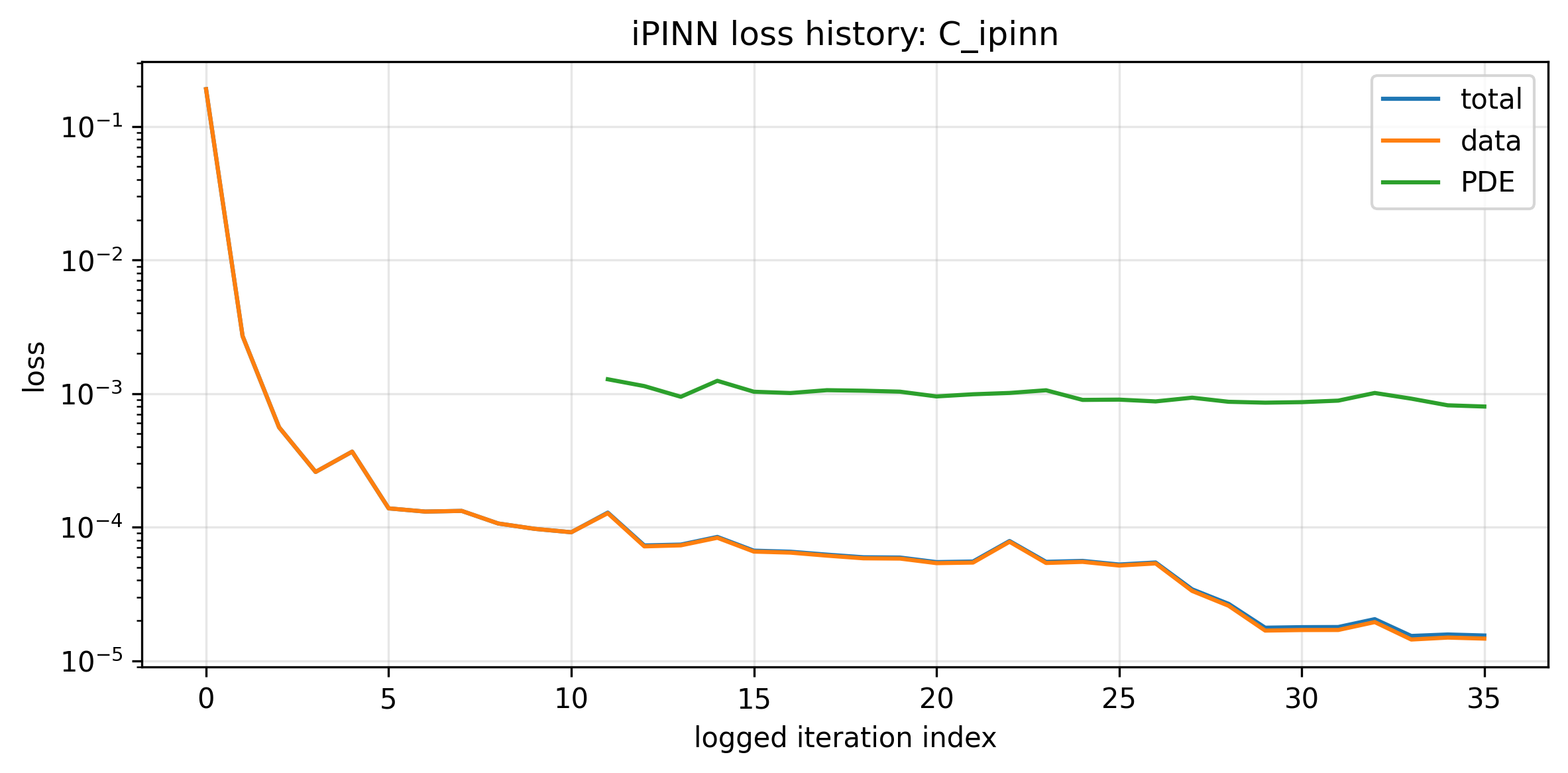}\caption{C loss history}\end{subfigure}

\vspace{0.5em}

\begin{subfigure}{0.49\textwidth}\centering
\includegraphics[width=\linewidth]{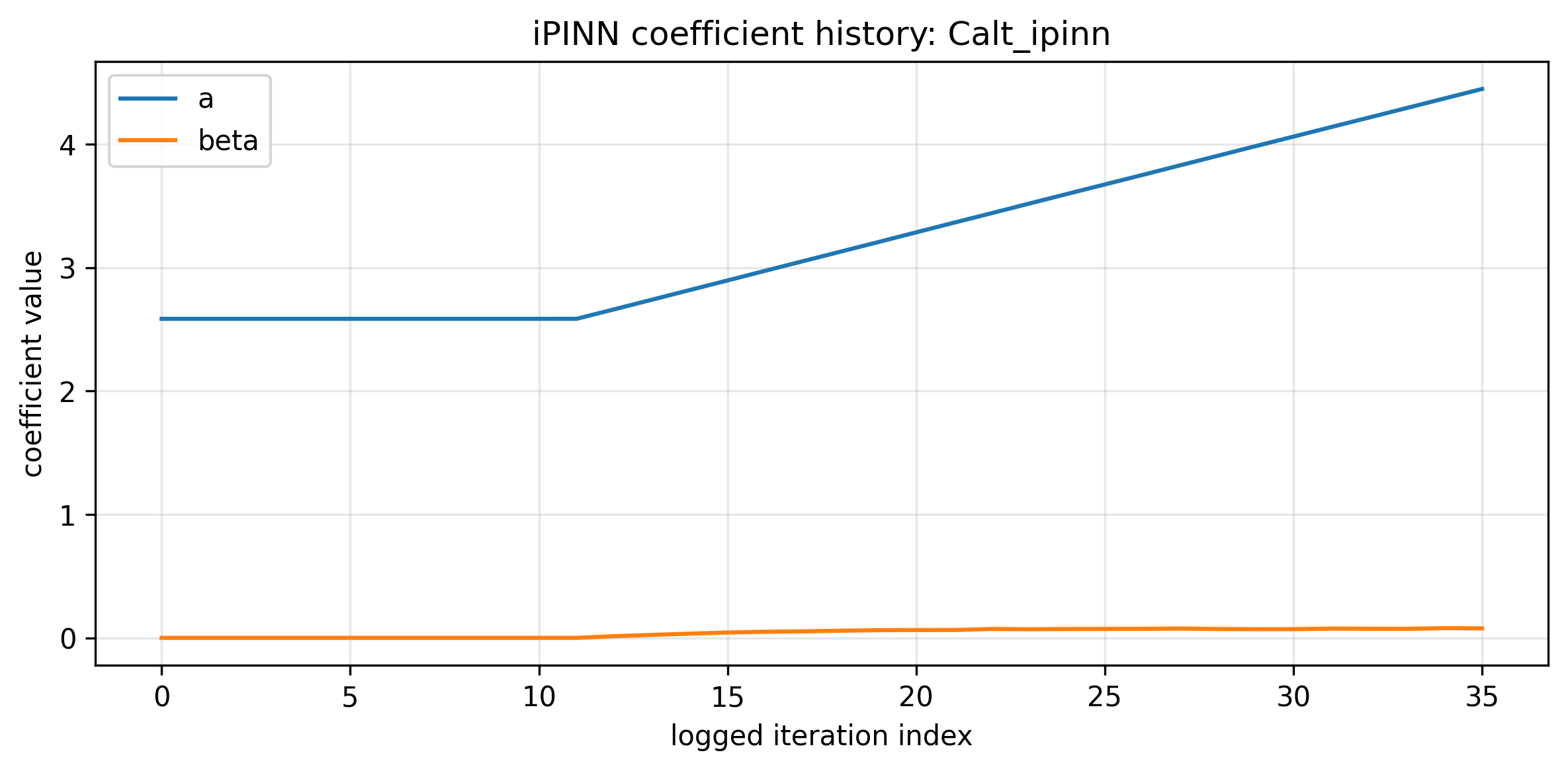}\caption{C-alt coefficient history}\end{subfigure}\hfill
\begin{subfigure}{0.49\textwidth}\centering
\includegraphics[width=\linewidth]{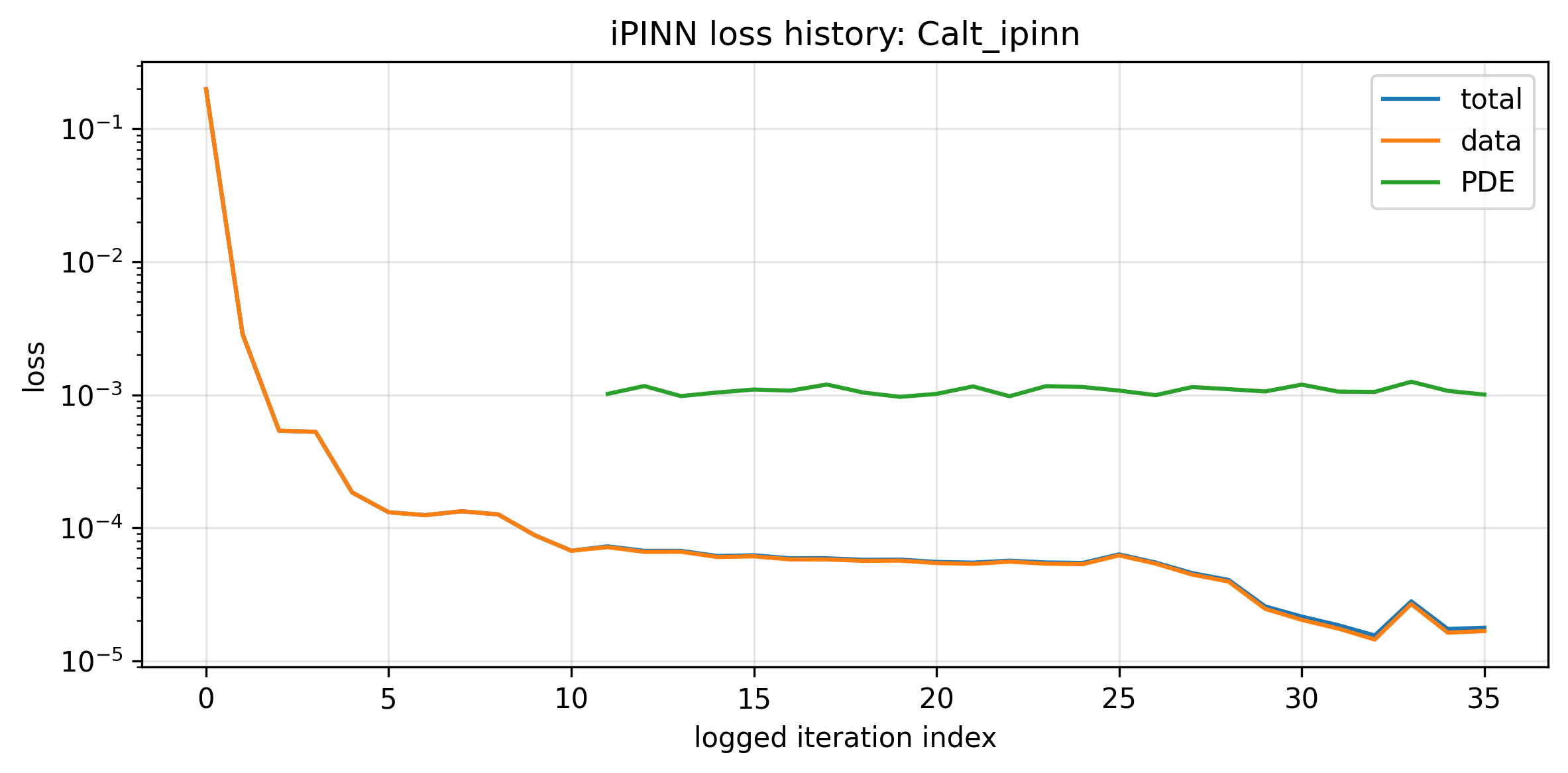}\caption{C-alt loss history}\end{subfigure}
\caption{iPINN coefficient and loss trajectories for the C and C-alt models.}
\label{fig:ipinn_histories}
\end{figure}

\begin{figure}
\centering
\begin{subfigure}{0.49\textwidth}\centering
\includegraphics[width=\linewidth]{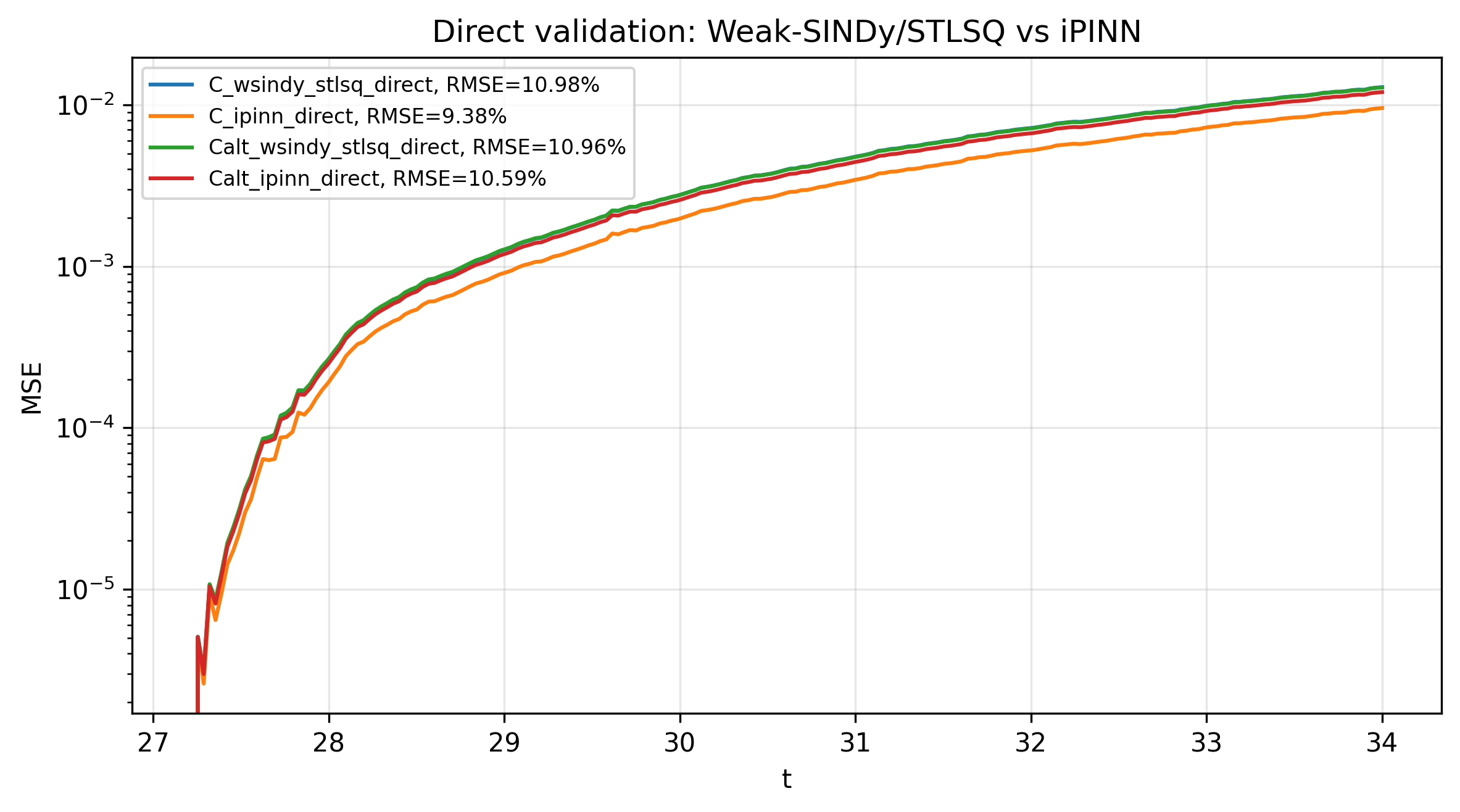}\caption{MSE over validation time}\end{subfigure}\hfill
\begin{subfigure}{0.49\textwidth}\centering
\includegraphics[width=\linewidth]{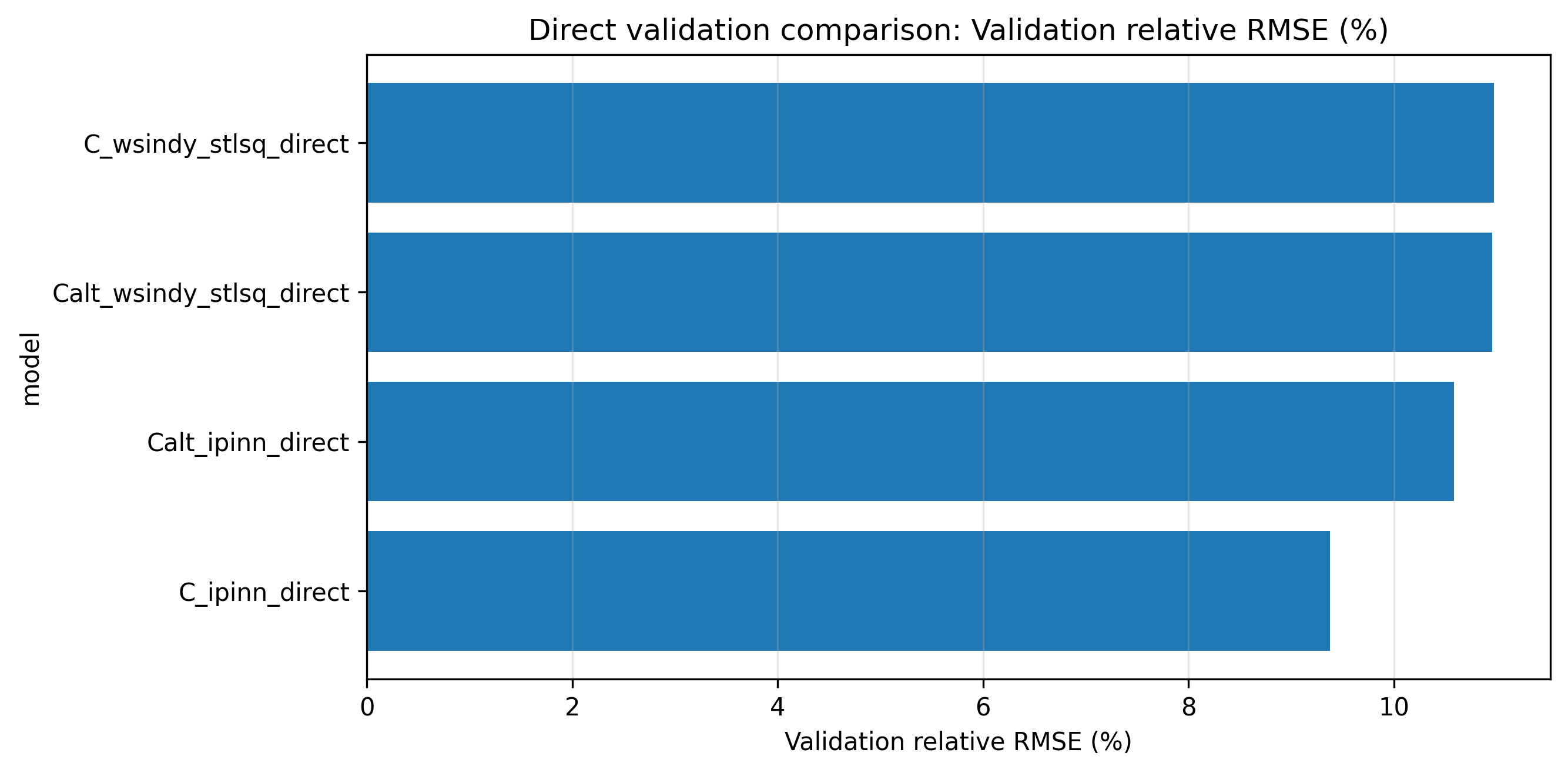}\caption{Relative RMSE}\end{subfigure}

\vspace{0.5em}

\begin{subfigure}{0.49\textwidth}\centering
\includegraphics[width=\linewidth]{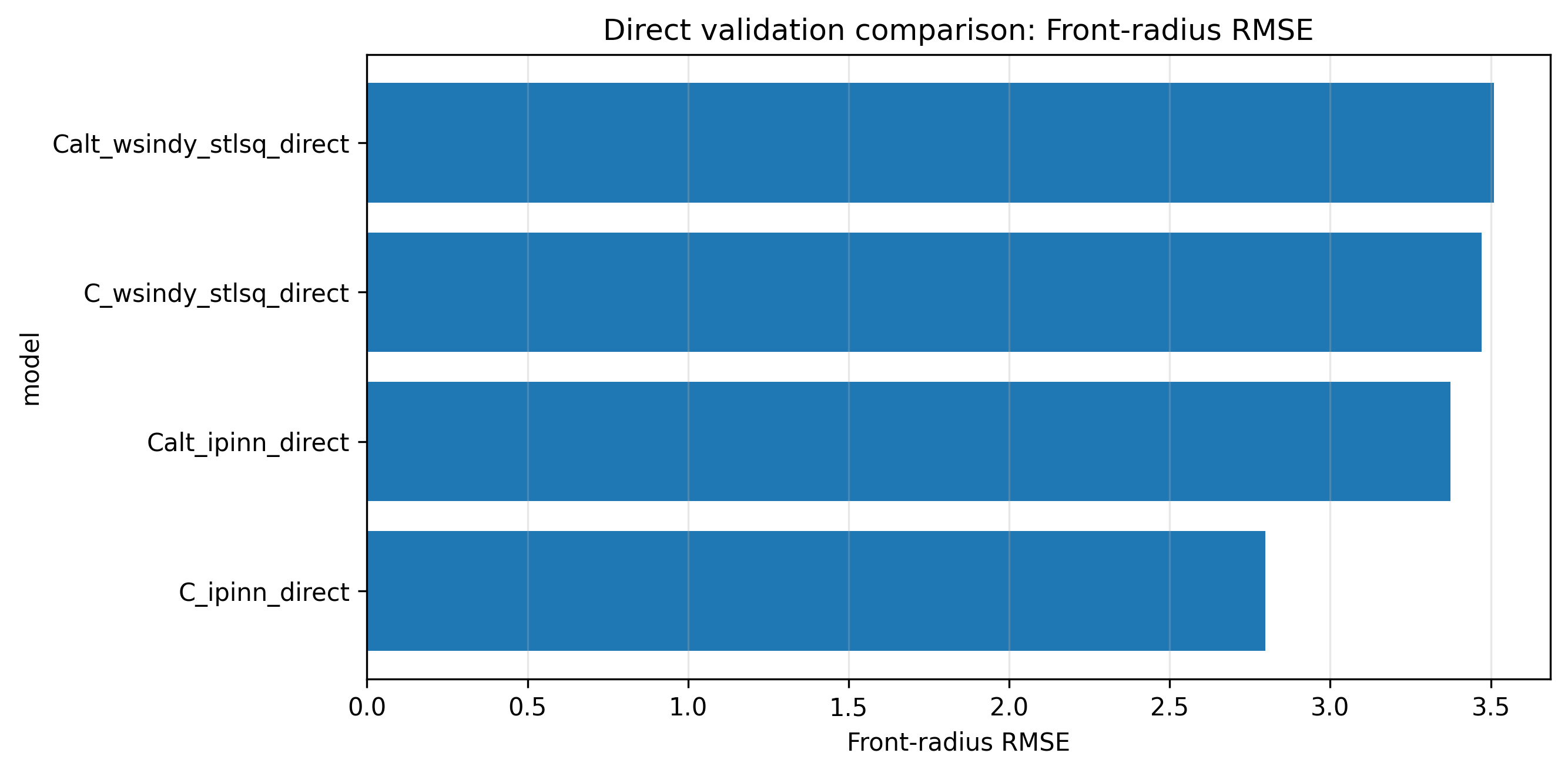}\caption{Front-radius RMSE}\end{subfigure}\hfill
\begin{subfigure}{0.49\textwidth}\centering
\includegraphics[width=\linewidth]{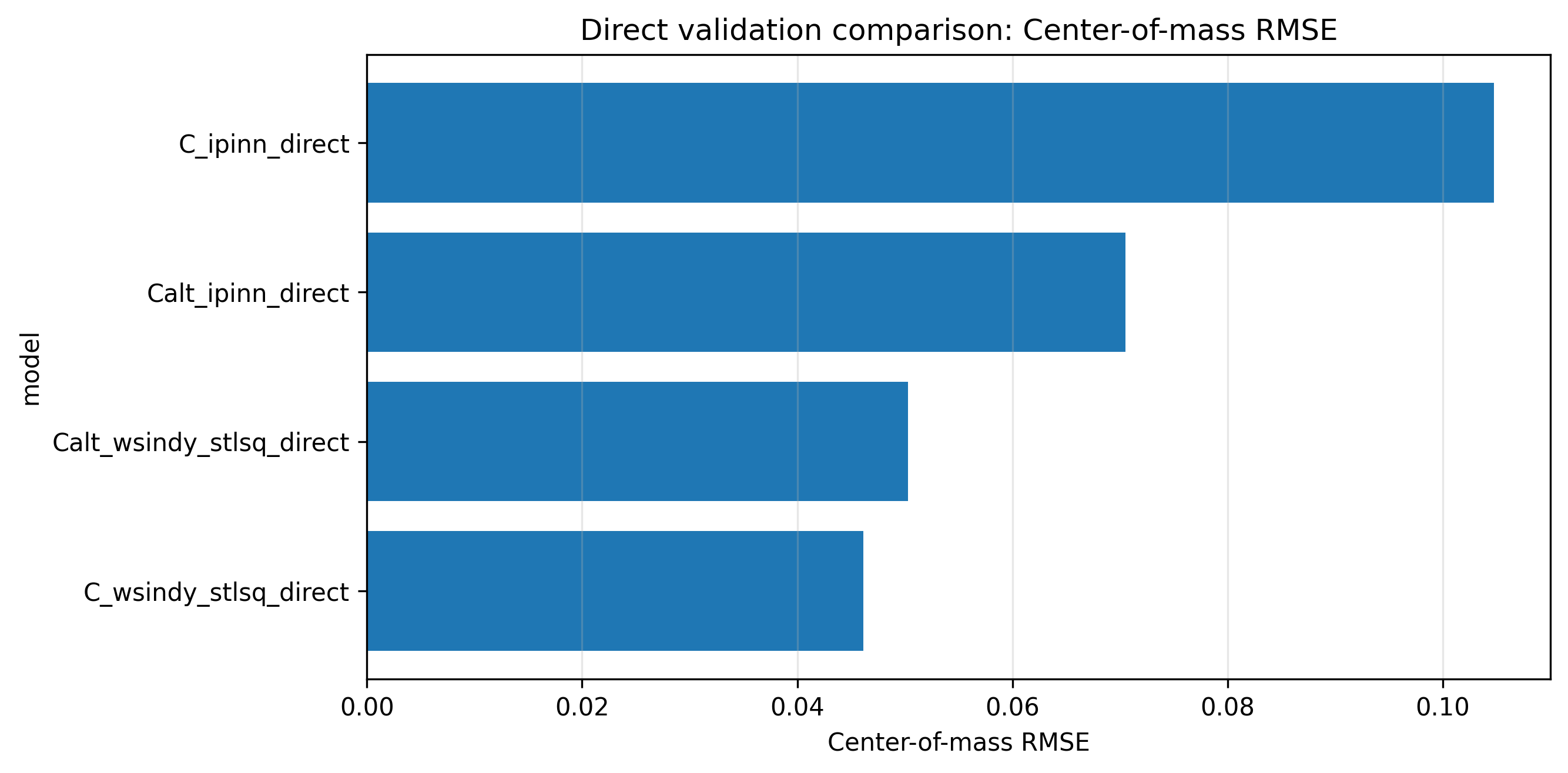}\caption{Centroid RMSE}\end{subfigure}
\caption{Direct validation comparison between weak-SINDy/STLSQ and iPINN-refined coefficients. The iPINN-refined C model gives the lowest pixel-wise rRMSE and front-radius RMSE; weak-SINDy/STLSQ coefficients preserve the centroid more accurately.}
\label{fig:direct_validation_ipinn}
\end{figure}

The iPINN refinement (Fig.~\ref{fig:ipinn_histories}, Table~\ref{tab:ipinn_coefficient_summary}) consistently strengthens the nonlinear-gradient coefficient and yields a positive Laplacian coefficient for both libraries. For C, \((a,\beta)\) moves from \((0.7931,-0.1258)\) to \((3.0032,0.0192)\); for C-alt, it moves from \((2.5860,0)\) to \((4.4492,0.0769)\). Direct validation rollouts show that the iPINN-refined coefficients improve pixel-wise rRMSE on both libraries -- from \(10.98\%\) to \(9.38\%\) for C and from \(10.96\%\) to \(10.59\%\) for C-alt -- and reduce the multi-level front-radius RMSE (Fig.~\ref{fig:direct_validation_ipinn}, Table~\ref{tab:direct_validation_ipinn_summary}). The centroid error, however, increases slightly after refinement, indicating that pixel-wise, front-radius, and bulk-trajectory accuracy do not optimise the same aspect of the dynamics. The iPINN values are therefore used as improved initialisations for the rollout-based calibration that follows, rather than as final models in their own right.

\begin{table}
\centering
\caption{Weak-SINDy and iPINN-refined coefficients for the C and C-alt models.}
\label{tab:ipinn_coefficient_summary}
\small
\begin{tabular*}{\linewidth}{@{\extracolsep{\fill}}lcccc@{}}
\toprule
\textbf{Model} & \(\boldsymbol{a_{\mathrm{WS}}}\) & \(\boldsymbol{\beta_{\mathrm{WS}}}\) & \(\boldsymbol{a_{\mathrm{iPINN}}}\) & \(\boldsymbol{\beta_{\mathrm{iPINN}}}\) \\
\midrule
C: \(|\nabla u|^{2}+\Delta u\) & \(0.7931\) & \(-0.1258\) & \(3.0032\) & \(0.0192\) \\
C-alt: \(u|\nabla u|^{2}+\Delta u\) & \(2.5860\) & \(0.0000\) & \(4.4492\) & \(0.0769\) \\
\bottomrule
\end{tabular*}
\end{table}

\begin{table}
\centering
\caption{Direct validation comparison of weak-SINDy/STLSQ and iPINN-refined coefficients; ordered by validation rRMSE.}
\label{tab:direct_validation_ipinn_summary}
\scriptsize
\resizebox{\linewidth}{!}{%
\begin{tabular}{lcccccc}
\toprule
\textbf{Model} & \textbf{Calibration} & \textbf{RMSE (\%)} & \textbf{Front MAE} & \textbf{Front RMSE} & \textbf{COM MAE} & \textbf{COM RMSE} \\
\midrule
C & iPINN & \(9.3813\) & \(2.4335\) & \(2.7980\) & \(0.0922\) & \(0.1048\) \\
C-alt & iPINN & \(10.5888\) & \(2.9375\) & \(3.3741\) & \(0.0622\) & \(0.0705\) \\
C-alt & Weak-SINDy/STLSQ & \(10.9594\) & \(3.0568\) & \(3.5106\) & \(0.0448\) & \(0.0503\) \\
C & Weak-SINDy/STLSQ & \(10.9785\) & \(3.0242\) & \(3.4713\) & \(0.0411\) & \(0.0462\) \\
\bottomrule
\end{tabular}}
\end{table}

\subsection{Bootstrap Rollout Calibration}\label{subsec:bootstrap_results}

All four bootstrap configurations converge in \(50/50\) replicates. The C model initialised from the weak-SINDy/STLSQ coefficients gives the lowest validation rRMSE, \(6.19\%\), and moves to \((a,\beta)=(9.005,0.666)\) -- a substantially larger nonlinear-gradient coefficient and a positive Laplacian coefficient (Table~\ref{tab:bootstrap_standard_summary}). C-alt is markedly less responsive to rollout-based calibration, with both initialisations remaining near \(11\%\) validation rRMSE, indicating that the weighted-gradient structure has reduced expressive freedom once the prescribed drift is fixed. The standard bootstrap therefore selects
\begin{equation}
    \boxed{\;u_t + \mathbf{v}(t)\!\cdot\!\nabla u = 9.00543\,|\nabla u|^{2} + 0.666307\,\Delta u\;}
    \label{eq:final_selected_model}
\end{equation}
as the validation-best calibrated model.

\begin{table}
\centering
\caption{Bootstrap rollout-calibration results; ordered by validation rRMSE.}
\label{tab:bootstrap_standard_summary}
\scriptsize
\resizebox{\linewidth}{!}{%
\begin{tabular}{lcccccc}
\toprule
\textbf{Model} & \textbf{Init.\ source} & \textbf{Conv.} & \textbf{Median train MSE} & \(\boldsymbol{a}\) & \(\boldsymbol{\beta}\) & \textbf{Val.\ RMSE (\%)} \\
\midrule
C     & Weak-SINDy/STLSQ & \(50/50\) & \(1.745\times 10^{-2}\) & \(9.00543\) & \(0.666307\) & \(6.19\) \\
C     & iPINN            & \(50/50\) & \(2.603\times 10^{-2}\) & \(2.70323\) & \(0.048973\) & \(9.57\) \\
C-alt & iPINN            & \(50/50\) & \(3.052\times 10^{-2}\) & \(2.50615\) & \(0.160177\) & \(10.81\) \\
C-alt & Weak-SINDy/STLSQ & \(50/50\) & \(3.325\times 10^{-2}\) & \(0.99552\) & \(0.003615\) & \(11.20\) \\
\bottomrule
\end{tabular}}
\end{table}

\subsection{Front-Aware Recalibration with Positive Laplacian}\label{subsec:front_aware_results}

Replacing the pixel-wise objective by the front-aware objective in Eq.~\eqref{eq:front_aware_objective}, with \(\beta\) constrained to be positive in log-space, gives an alternative C-family model. Both initialisations converge to essentially the same coefficient pair (Table~\ref{tab:front_aware_positive_lap_summary}); the validation rRMSE rises to \(7.58\%\) and the centroid RMSE rises to \(\approx 0.50\), but the nonlinear-gradient coefficient is much larger, \(a\approx 34.1\), with \(\beta\approx 0.536\). The front-aware calibration therefore improves geometric matching of the plume front but worsens bulk-trajectory accuracy, and remains physically admissible by construction.

\begin{table}
\centering
\caption{Front-aware recalibration of the bootstrap-calibrated C models with \(\beta>0\) constraint.}
\label{tab:front_aware_positive_lap_summary}
\scriptsize
\resizebox{\linewidth}{!}{%
\begin{tabular}{lcccccccc}
\toprule
\textbf{Model} & \textbf{Init.\ source} & \(\boldsymbol{a}\) & \(\boldsymbol{\beta}\) & \(\boldsymbol{J}\) & \textbf{Val.\ RMSE (\%)} & \textbf{Front RMSE} & \textbf{COM RMSE} & \textbf{Conv.} \\
\midrule
C front-aware & Weak-SINDy bootstrap & \(34.1083\) & \(0.535200\) & \(9.039\times 10^{-3}\) & \(7.5826\) & \(3.1631\) & \(0.4973\) & yes \\
C front-aware & iPINN bootstrap      & \(34.1216\) & \(0.537043\) & \(9.047\times 10^{-3}\) & \(7.5852\) & \(3.1643\) & \(0.4973\) & yes \\
\bottomrule
\end{tabular}}
\end{table}

\subsection{Final Model Selection}\label{subsec:final_model_selection}

The selection criteria yield two admissible models: the standard bootstrap C model in Eq.~\eqref{eq:final_selected_model}, with the lowest validation rRMSE (\(6.19\%\)) and a positive Laplacian coefficient, and the front-aware C model with \(a\approx 34.1\), \(\beta\approx 0.536\), and validation rRMSE \(7.58\%\). The standard bootstrap C model is selected as the predictive model because it minimises the validation criterion that drives selection while satisfying all four admissibility constraints (rRMSE, front-radius error, centroid error, sign of \(\beta\)). The front-aware model is retained as a geometric diagnostic for downstream tasks that prioritise front-radius behaviour over bulk-trajectory accuracy. Once selected, Eq.~\eqref{eq:final_selected_model} is frozen and reported on the held-out test window.

\FloatBarrier
\section{Discussion}\label{sec:discussion}

\subsection{Cole--Hopf Reduction and Continuum Structure}\label{subsec:analytical_structure_discovered_pde}

The selected model in Eq.~\eqref{eq:final_selected_model} belongs to the class
\begin{equation}
    u_t + \mathbf{v}(t)\!\cdot\!\nabla u = a\,|\nabla u|^{2} + \beta\,\Delta u, \qquad a,\beta>0,
    \label{eq:general_discovered_pde}
\end{equation}
a viscous Hamilton--Jacobi equation closely related to the deterministic Kardar--Parisi--Zhang form~\citep{KardarParisiZhang1986}. Its key analytical feature is the Cole--Hopf transformation~\citep{Cole1951,Hopf1950}: setting
\(\theta = \exp(au/\beta)\) gives, by direct differentiation,
\begin{equation}
    \theta_t + \mathbf{v}(t)\!\cdot\!\nabla\theta = \beta\,\Delta\theta,
    \label{eq:linearized_theta_equation}
\end{equation}
so the discovered nonlinear PDE is equivalent under the monotone change of variables \(\theta=\exp(au/\beta)\) to a linear advection--diffusion equation. The interpretation that follows is structural rather than empirical: when \(\beta>0\), the term \(a|\nabla u|^{2}\) is not an anti-diffusive mechanism but the gradient signature of an underlying diffusive scalar expressed in logarithmic variables. The spatially uniform drift can also be removed by passing to the moving frame \(w(z,t) = u(z+X(t),t)\) with \(X(t)=\int_0^t\mathbf v(s)\,ds\), under which \(w_t = a|\nabla w|^{2}+\beta\Delta w\); the bulk translation accounts for the centroid motion, while the viscous Hamilton--Jacobi component governs internal spreading and shape evolution.

The Cole--Hopf structure carries continuum-level stability properties. Under boundary conditions for which the transport and diffusion fluxes vanish on \(\partial\Omega\) (or under periodic boundaries), \(\theta\) inherits the parabolic maximum principle, and the monotone exponential map transfers it to \(u\):
\begin{equation}
    \min_{\Omega}u_0 \le u(x,y,t) \le \max_{\Omega}u_0, \qquad t\ge 0.
    \label{eq:max_principle_final_u}
\end{equation}
Multiplying Eq.~\eqref{eq:linearized_theta_equation} by \(\theta\) and integrating against the same boundary conditions yields the energy identity
\(\tfrac{1}{2}\tfrac{d}{dt}\|\theta\|_{L^{2}}^{2} + \beta\|\nabla\theta\|_{L^{2}}^{2} = 0\),
which translates into an exponential dissipation law for \(u\),
\begin{equation}
    \frac{1}{2}\frac{d}{dt}\!\int_{\Omega}\!\!e^{2au/\beta}\,dA + \frac{a^{2}}{\beta}\!\int_{\Omega}\!\!e^{2au/\beta}\,|\nabla u|^{2}\,dA = 0.
    \label{eq:exponential_energy_final_u}
\end{equation}
The natural conserved quantity is the exponential mass \(\int_\Omega e^{au/\beta}\,dA\); the spatial average of \(u\) itself is \emph{not} conserved, since \(\tfrac{d}{dt}\int u\,dA = a\int|\nabla u|^{2}\,dA \ge 0\). On periodic domains, \(\theta\) relaxes to its spatial mean and \(u\) consequently approaches \(\frac{\beta}{a}\log\!\left(|\Omega|^{-1}\!\int_{\Omega}e^{au_0/\beta}\,dA\right)\). These properties should be read as continuum-level structural statements under idealised boundary conditions, not as exact conservation laws for the finite, cropped, image-derived dataset; they nonetheless show that the selected model has a stable parabolic structure despite its nonlinear gradient term. The front-aware model in Table~\ref{tab:front_aware_positive_lap_summary} also has \(a,\beta>0\) and therefore belongs to the same Cole--Hopf-linearisable class, with the same maximum principle and the same exponential energy law.

\subsection{Why a Staged Pipeline}\label{sec:pipeline_reflection}

A single-step sparse-regression fit on the full library would have produced a competitive validation rRMSE but a much less informative model: the column-correlation analysis (\S\ref{subsec:library_conditioning_thresholds}) shows that the full library is overcomplete and overlapping, and \citet{antonelli2022} document this as a generic risk of automatic equation discovery on noisy data. Two design decisions were therefore consequential. The first was imposing the centroid-derived drift as a prescribed advective term: when the drift coefficient is left free (Mode A), pixel-wise rollout error remains acceptable but the centroid trajectory drifts by an order of magnitude (Table~\ref{tab:validation_AB_summary}), because the regression has the freedom to compensate drift mismatch through the diffusive and gradient terms. The second was decoupling structure discovery from coefficient calibration. The weak-SINDy coefficient for the Laplacian is negative on the C library; the iPINN refinement pulls it toward small positive values; bootstrap rollout calibration then pushes both \(a\) and \(\beta\) much larger and yields the validation-best model in Eq.~\eqref{eq:final_selected_model}. Each stage therefore corrects a different deficiency of the previous one -- a behaviour consistent with the loss-landscape difficulties documented for inverse PINN problems~\citep{Krishnapriyan2021PINNFailureModes}.

The two admissible C-family models in this study illustrate that pixel-wise, front-radius, and centroid diagnostics need not be co-optimal. The bootstrap C model in Eq.~\eqref{eq:final_selected_model} is the better choice for full-field prediction; the front-aware C model is more useful when threshold-front behaviour is the primary downstream quantity. Both belong to the same analytical class, and the difference between them is a calibration choice tied to the objective rather than a difference in structure.

\subsection{Limitations and Outlook}\label{sec:limitations_future_work}

Four limitations should be made explicit. The recovered coefficients are interpreted in image-coordinate units and as parameters of an effective observable-level PDE; illumination, camera response, dye opacity, and normalisation all enter the recovered field, and a calibration of \(u\) to physical concentration would change the numerical values of \(a\) and \(\beta\) (though not the structure of Eq.~\eqref{eq:general_discovered_pde}). Preprocessing -- cropping, border removal, resizing, intensity inversion, smoothing -- shapes the gradients fed into the weak system; the smoothing sweep in Appendix~\ref{app:smoothing_diagnostic} is a partial control, and a wider sensitivity study across crop windows and resize factors would further test robustness. The drift used here is spatially uniform, sufficient for the centroid translation observed in this experiment but unable to represent shear, recirculation, or local velocity variations; combining the present weak-form pipeline with optical flow or a learned spatially varying velocity field is a natural extension. Finally, the discovered PDE has been validated on a single video sequence; transfer across repeated experiments, different initial conditions, fluid depths, and -- ideally -- calibrated concentration measurements remains to be tested.

Beyond dye-plume experiments, the same combination of weak-form discovery, neural coefficient refinement, bootstrap uncertainty quantification, and geometric validation may apply to biological, ecological, and medical imaging settings in which evolving scalar fields admit only partially known governing laws. Encoding admissibility constraints (e.g.\ \(\beta>0\), positivity of the field, mass conservation) directly into the discovery and calibration stages, rather than enforcing them post hoc, is a useful direction for further methodological work.

\FloatBarrier
\section{Conclusion}\label{sec:conclusion}

We developed a video-to-PDE pipeline that converts grayscale dye-plume video into an interpretable, simulable evolution law without ever differentiating the noisy field directly. Conditioning, threshold-sweep, and random-centre diagnostics ruled out overcomplete and overlapping libraries on identifiability grounds; rollout-based calibration with the centroid-derived drift held fixed selected the reduced model in Eq.~\eqref{eq:final_selected_model}. The selected equation outperforms advection--diffusion baselines on validation rollouts, retains a positive Laplacian coefficient consistent with effective diffusion, and, through the Cole--Hopf reduction in Eq.~\eqref{eq:linearized_theta_equation}, links the empirical nonlinear gradient term to a known parabolic class.

Beyond the specific equation, the case study supports a methodological point: video-derived PDE discovery is best treated as a pipeline of discovery, calibration, and uncertainty assessment. Rollout error alone is too permissive; conditioning, identifiability, geometric admissibility, and drift consistency all enter the selection. The same template should transfer to other imaging settings in which the recorded quantity is an observable rather than a calibrated state.

\FloatBarrier

\appendix

\section{Additional Preprocessing Diagnostics}\label{app:preprocessing_diagnostics}
\subsection{Gaussian Smoothing Sensitivity}\label{app:smoothing_diagnostic}

\begin{figure}
    \centering
    \includegraphics[width=\textwidth]{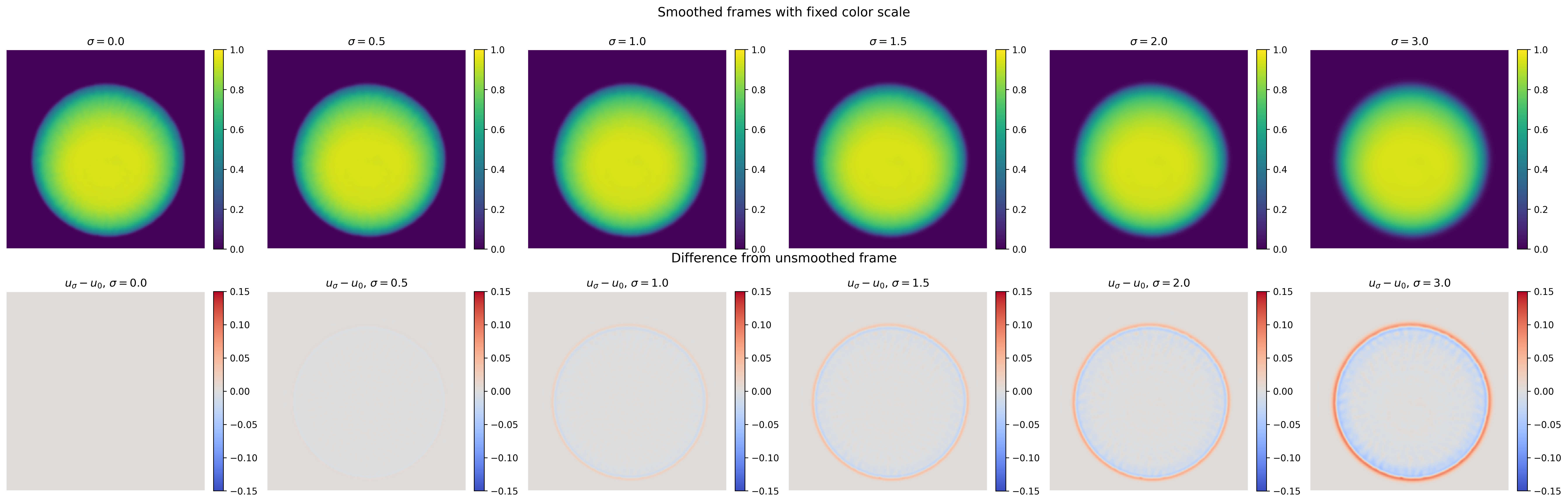}
    \caption{Gaussian smoothing sensitivity for a representative frame. Top row: smoothed frames at increasing \(\sigma\). Middle row: difference \(u_\sigma - u_0\) from the unsmoothed frame. The selected value \(\sigma=1.0\) preserves the front profile while reducing pixel-scale noise.}
    \label{fig:smoothing_diagnostic}
\end{figure}

\begin{figure}
    \centering
    \includegraphics[width=0.4\linewidth]{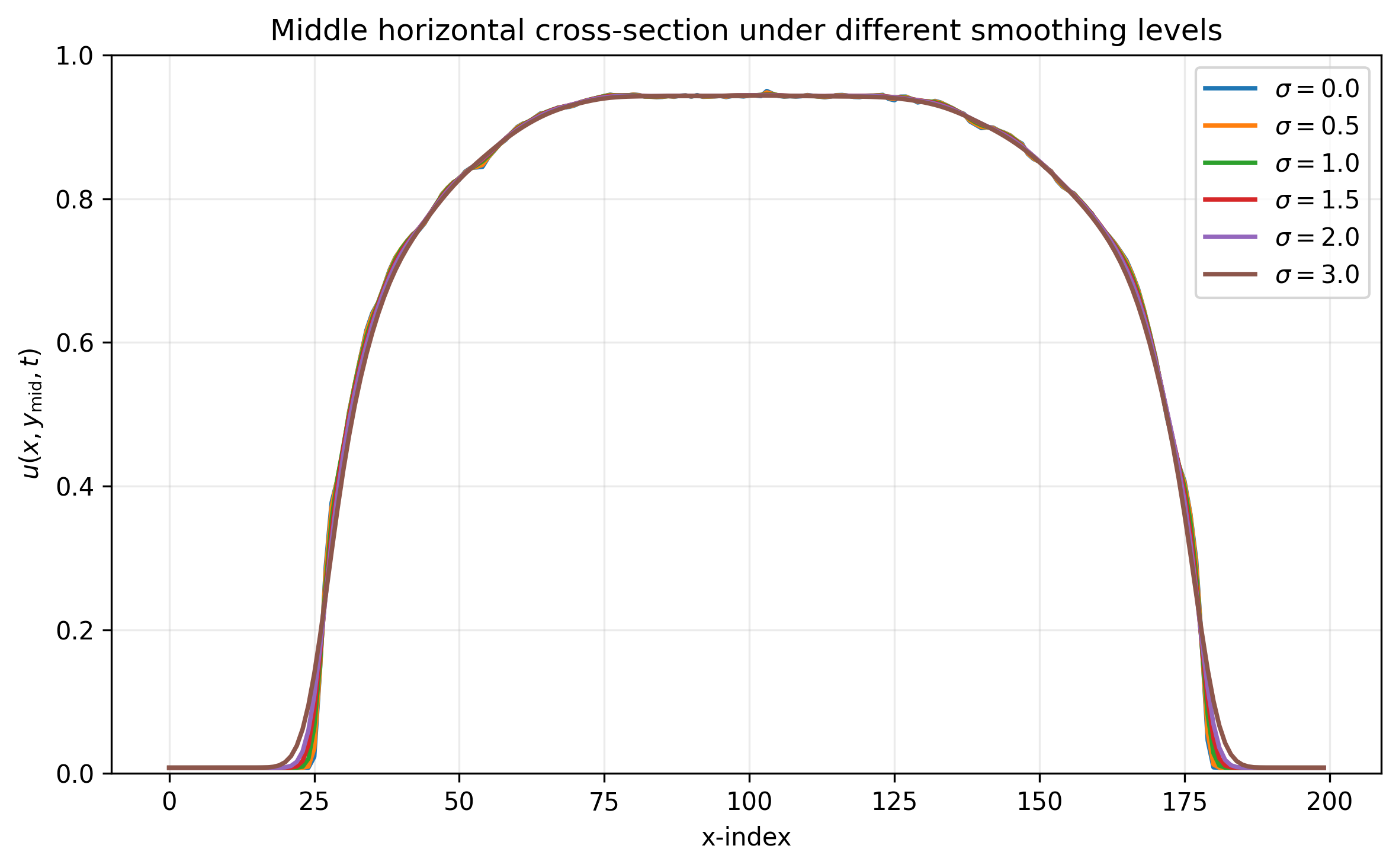}
    \caption{Middle horizontal cross-section under different smoothing levels.}
    \label{fig:smoothing_midline}
\end{figure}

The smoothing parameter \(\sigma\) was selected from the sweep \(\sigma\in\{0,0.5,1.0,1.5,2.0,3.0\}\) on a representative processed frame (Fig.~\ref{fig:smoothing_diagnostic}). The middle horizontal cross-section (Fig.~\ref{fig:smoothing_midline}) is essentially invariant on the plume body across the tested range; differences appear only near the front, where larger \(\sigma\) broadens the transition region. Frame-level RMS differences from the unsmoothed image are listed in Table~\ref{tab:smoothing_diagnostic}. The choice \(\sigma=1.0\) suppresses pixel-scale noise while preserving the front profile and the large-scale plume geometry, and is therefore used throughout.

\begin{table}[htbp]
    \centering
    \caption{Frame-level differences relative to the unsmoothed image.}
    \label{tab:smoothing_diagnostic}
    \small
    \begin{tabular}{ccc}
    \toprule
    \(\sigma\) & RMS diff. & Max diff. \\
    \midrule
    0.0 & \(0.0000\) & \(0.0000\) \\
    0.5 & \(1.78\times 10^{-3}\) & \(1.74\times 10^{-2}\) \\
    1.0 & \(6.25\times 10^{-3}\) & \(6.07\times 10^{-2}\) \\
    1.5 & \(1.04\times 10^{-2}\) & \(9.36\times 10^{-2}\) \\
    2.0 & \(1.43\times 10^{-2}\) & \(1.17\times 10^{-1}\) \\
    3.0 & \(2.15\times 10^{-2}\) & \(1.50\times 10^{-1}\) \\
    \bottomrule
    \end{tabular}
\end{table}


\printcredits

\section*{Acknowledgements}

The authors thank the colleagues and reviewers who provided feedback during the preparation of this manuscript.

\section*{Declaration of competing interest}

The authors declare that they have no known competing interests that could have influenced the work reported in this paper.

\section*{Funding}

This research did not receive any specific grant from funding agencies in the public, commercial, or not-for-profit sectors.

\section*{Data availability}

The code, processed data, and scripts needed to reproduce the numerical results are available at
\url{https://github.com/Sayantan128/From-Video-to-PDE-Data-Driven-Discovery-of-Nonlinear-Dye-Plume-Dynamics}.
The raw video data are available from the corresponding author upon reasonable request.

\section*{Declaration of generative AI and AI-assisted technologies in the manuscript preparation process}

During the preparation of this work, the authors used ChatGPT in order to assist with language refinement, organization, and editorial polishing. After using this tool, the authors reviewed and edited the content as needed and take full responsibility for the content of the published article.

\bibliographystyle{cas-model2-names}
\bibliography{cas-refs}

%
%

\end{document}